%% file: main.tex
\documentclass[dvipsnames, 12pt]{article}

\pdfoutput=1

\usepackage[preprint]{tmlr}

\usepackage[T1]{fontenc}    
\usepackage[utf8]{inputenc} 
\usepackage[hyphens]{url}
\usepackage{float}
\usepackage{hyperref}       
\hypersetup{
    colorlinks=true,
    citecolor=RoyalBlue, 
    linkcolor=RoyalBlue, 
    filecolor=magenta,      
    urlcolor=RoyalBlue 
}
\usepackage{url}            
\usepackage{booktabs}       
\usepackage{amsfonts}       
\usepackage{nicefrac}       
\usepackage{microtype}      
\usepackage{xcolor}         
\usepackage{amsmath}
\usepackage{graphicx}
\usepackage{enumitem}
\usepackage{multirow}
\usepackage{subcaption}
\usepackage{csquotes}
\usepackage{listings}
\usepackage{tcolorbox}
\usepackage{xspace}
\usepackage{makecell}
\usepackage{soul}               
\setstcolor{red}            
\usepackage{tikz}

\usepackage{tablefootnote}
\usepackage[edges]{forest}
\definecolor{hidden-draw}{RGB}{20,68,106}
\definecolor{hidden-pink}{RGB}{255,245,247}
\definecolor{red}{RGB}{255,0,0}

\begin{document}
\title{\centering Large Language Model Safety: A Holistic Survey}
\author{
\centerline{Dan Shi$^1$\thanks{Equal contribution}\hspace{0.5em}, Tianhao Shen$^1$\footnotemark[1]\hspace{0.5em}, Yufei Huang$^1$, Zhigen Li$^{1,2}$,}\\
\centerline{Yongqi Leng$^1$, Renren Jin$^1$, Chuang Liu$^1$, Xinwei Wu$^1$, Zishan Guo$^1$,}
\centerline{Linhao Yu$^1$, Ling Shi$^1$, Bojian Jiang$^{1,3}$, Deyi Xiong$^1$\thanks{Corresponding author. Correspondence to: \texttt{\{shidan, thshen, dyxiong\}@tju.edu.cn}}}\vspace{0.5em}\\
\centerline{\normalfont{$^1$TJUNLP Lab, Tianjin University}}
\centerline{\normalfont{$^2$Ping An Technology}}
\centerline{\normalfont{$^3$Du Xiaoman Finance}}
\vspace{0.05em}
}
\maketitle

\begin{abstract}
\input{sections/Abstract}
\end{abstract}

\newpage
\tableofcontents
\newpage

\input{sections/introduction}

\input{sections/Taxonomy}

\input{sections/Value_Misalignment}

\input{sections/Robustness_to_Attack}

\input{taxonomies/Misuse_Taxonomy}
\input{sections/Misuse}

\input{sections/Autonomous_AI_Risks}

\input{sections/Agent_Safety}

\input{sections/Interpretability_for_LLM_Safety}

\input{taxonomies/Technology_Roadmaps_Taxonomy}
\input{sections/Technology_Roadmaps}

\input{taxonomies/Governance_Taxonomy}
\input{sections/Governance}

\input{sections/Challenges_and_Future_Directions}

\input{sections/Conclusion}

\newpage

\setcitestyle{authoryear,open={(},close={)}}
\bibliography{main_dedup}
\bibliographystyle{tmlr}

\end{document}

%% file: sections/Abstract.tex
The rapid development and deployment of large language models (LLMs) have introduced a new frontier in artificial intelligence, marked by unprecedented capabilities in natural language understanding and generation. However, the increasing integration of these models into critical applications raises substantial safety concerns, necessitating a thorough examination of their potential risks and associated mitigation strategies. 

This survey provides a comprehensive overview of the current landscape of LLM safety, covering four major categories: value misalignment, robustness to adversarial attacks, misuse, and autonomous AI risks. In addition to the comprehensive review of the mitigation methodologies and evaluation resources on these four aspects, we further explore four topics related to LLM safety: the safety implications of LLM agents, the role of interpretability in enhancing LLM safety, the technology roadmaps proposed and abided by a list of AI companies and institutes for LLM safety, and AI governance aimed at LLM safety with discussions on international cooperation, policy proposals, and prospective regulatory directions.

Our findings underscore the necessity for a proactive, multifaceted approach to LLM safety, emphasizing the integration of technical solutions, ethical considerations, and robust governance frameworks. This survey is intended to serve as a foundational resource for academy researchers, industry practitioners, and policymakers, offering insights into the challenges and opportunities associated with the safe integration of LLMs into society. Ultimately, it seeks to contribute to the safe and beneficial development of LLMs, aligning with the overarching goal of harnessing AI for societal advancement and well-being. A curated list of related papers has been publicly available at a GitHub repository.\footnote{\url{https://github.com/tjunlp-lab/Awesome-LLM-Safety-Papers}}

%% file: sections/Introduction.tex
\section{Introduction}


In 1950, Alan Turing posed the question, ``Can machines think?'' \citep{DBLP:journals/x/Turing50}. At that time, the prevailing answer was ``no.'' Over the subsequent decades, building an artificial intelligence (AI) system capable of human-like thinking has become a long-standing pursuit in AI research, with substantial efforts dedicated to achieving this vision \citep{4066245,DBLP:journals/ai/Brooks91,DBLP:journals/aim/McCarthyMRS06,DBLP:journals/nature/LeCunBH15,DBLP:journals/corr/LakeUTG16}. Such sustained endeavors have driven AI from theoretical explorations to practical applications. The early stages of AI focus on symbolic reasoning and rule-based systems to replicate human intelligence \citep{McCarthy1960ProgramsWC}, such as rule-based machine translation\footnote{https://en.wikipedia.org/wiki/Georgetown\%E2\%80\%93IBM\_experiment} and expert systems \citep{DBLP:books/lib/Hayes-Roth83}. However, symbolic AI relies heavily on human-crafted rules, making it difficult to scale to new, dynamic, or complex environments. In the 1980s and and 1990s, increased data availability and computational power have shifted AI research towards data-driven approaches, allowing machines to automatically discover features from data to fulfill specific tasks. These stages have witnessed the resurgence of machine learning with substantial progress made in artificial neural networks, support vector machines and random forests. Although these machine learning approaches achieve promising performance on specific tasks, they often depend on feature engineering and struggle with handling complex, real-world problems. Since the 2010s, significant advances in neural architectures, computational resources, and access to extensive training data have led to a resurgence of artificial neural networks with deeper structures, typically referred to as deep learning. Deep learning represents a major milestone in AI, enabling the achievement of surpassing human-level performance in tasks such as image classification \citep{DBLP:conf/nips/KrizhevskySH12}, machine translation \citep{DBLP:conf/nips/SutskeverVL14}, and protein structure prediction \citep{jumper2021highly}. However, early deep learning models are often trained on task-specific data, resulting in models that are limited to a narrow range of capabilities. This limitation persists until the emergence of large language models (LLMs).

The development of LLMs in recent years marks a significant advancement towards the goal of “making machines think”, as these models demonstrate exceptional proficiency in human language understanding and generation, thereby significantly reducing barriers to communication between humans and AI \citep{DBLP:conf/nips/BrownMRSKDNSSAA20,DBLP:journals/corr/abs-2303-08774,DBLP:conf/emnlp/GuoYXJX23,DBLP:journals/corr/abs-2407-21783,DBLP:journals/corr/abs-2407-10671,DBLP:journals/corr/abs-2405-04434,DBLP:journals/corr/abs-2408-06273}. Compared to models from the early stages of deep learning, LLMs contain several orders of magnitude more parameters. Moreover, their training data typically encompass multiple tasks, domains, languages, and modalities, which contributes to their extensive capabilities. Although LLMs already exhibit more human-like intelligence than any preceding AI systems \citep{DBLP:journals/corr/abs-2303-12712,DBLP:conf/aaai/ShiYHLX24}, their performance continues to advance without signs of slowing down. Furthermore, due to their broad capacity and exceptional performance, LLMs have been deployed in numerous real-world applications.

However, the continuous improvement and widespread deployment of LLMs in real-world scenarios have raised significant concerns regarding their safety \citep{DBLP:journals/corr/abs-1802-07228,DBLP:journals/corr/abs-2112-04359,DBLP:journals/corr/abs-2108-07258}. Concerns on the risks posed by intelligent machines date as far back as the 1950s \citep{DBLP:books/daglib/0081951}, largely focusing on the social and ethical implications of AI. Nonetheless, unlike earlier concerns, which are largely speculative and theoretical due to the absence of highly capable AI systems, the extensive capabilities of LLMs present concrete risks. Recent studies indicate that, LLMs may generate inappropriate content that may be offensive or hateful \citep{DBLP:conf/emnlp/GehmanGSCS20,DBLP:conf/emnlp/DengZ0ZMMH22}. Additionally, LLMs can also exhibit stereotypes and social biases \citep{DBLP:journals/corr/abs-2309-00770,DBLP:conf/icml/LiangWMS21,DBLP:conf/coling/HuangX24,DBLP:journals/corr/abs-2405-06058}, compromise individual privacy \citep{DBLP:journals/corr/abs-2310-10383,DBLP:conf/iclr/StaabVBV24}, or violate ethical and moral standards \citep{DBLP:journals/corr/abs-2112-04359,DBLP:journals/corr/abs-2310-15337}. Moreover, they can be exploited by malicious users to threaten national safety and public safety, such as through the design of weapons or the manipulation of public opinion \citep{DBLP:journals/corr/abs-2306-03809,buchanan10truth}.\footnote{https://venturebeat.com/ai/propaganda-as-a-service-may-be-on-the-horizon-if-large-language-models-are-abused/} Notably, as LLMs become increasingly proficient at performing tasks, there is an emerging trend suggesting that these models may develop self-replication and self-preservation capabilities, and exhibit desires for power and resources \citep{DBLP:journals/alife/GaborIZLMBL22,DBLP:conf/acl/PerezRLNCHPOKKJ23}. This potential evolution could result in unforeseen and potentially harmful consequences for human society. Alarmingly, these challenges cannot be mitigated merely by scaling up the models or increasing the data and computational resources used for training \citep{DBLP:conf/nips/0001HS23}.


In view of these, a growing consensus that focusing on LLM safety is not only essential but also urgent has been reached across governments, medias, and AI communities. For instance, concerns have been raised about the risks of deploying LLMs in sensitive areas like healthcare and law, where even slight mistakes in outputs could have significant consequences. The need for human oversight and careful evaluation of LLM outputs is hence highly emphasized \citep{DBLP:conf/fat/Sterz0BHLML24}.

Furthermore, a number of leading experts argue that proactive measures are necessary to build trust and prevent misuse. They suggest that without proper safety measures, LLMs could be exploited to produce misinformation or be manipulated for economic or political gains \citep{DBLP:journals/corr/abs-2308-12833, DBLP:journals/corr/abs-2407-13934}. The rapid evolution of LLMs makes addressing these risks a priority to ensure the safe integration of capable AI systems into society.


In view of the urgent need for LLM safety technologies, strategies and national/global policies, we provide a comprehensive overview into LLM safety. We take a holistic view of LLM safety, including safety technologies, resources, evaluations, roadmaps, strategies,  policies, etc., and organize them into two dimensions: basic areas of LLM safety and related areas to LLM safety. The first dimension covers major risk areas/categories that the development and deployment of LLMs give rise to. Our analysis emphasizes the evaluation of risks associated with LLMs across various scenarios, which is consistent with the view taken by several recent AI safety reports (e.g., International Scientific Report on the Safety of Advanced AI \citep{InternationalScientificReport}) and safety institutes (e.g., AI Safety Institute (AISI) \citep{AISafetyInstitute}). This includes a thorough examination of value misalignment, robustness against targeted attacks, scenarios of both intentional and unintentional misuse, and potential risks posed by advanced AI systems that operate independently or autonomously in complex environments. Through a systematic review and analysis of these critical areas, we aim to provide researchers and policymakers with a holistic perspective on the current landscape of LLM safety research, identify existing research gaps, and propose potential directions for future exploration. 

In the dimension of related areas to LLM safety, we investigate the substantial risks that agents powered by LLMs, despite their exceptional problem-solving and task-planning capabilities, pose to humans and society. Moreover, our exploration encompasses the technology roadmaps and strategies to LLM safety employed by leading AI companies and institutions in practice. We also delve into interpretability methodologies for comprehensively studying and mitigating the unsafe behaviors of LLMs by examining their internal mechanisms. Finally, we expand our discussion to encompass national governance and global cooperation, investigating the multifaceted dimensions of AI governance. This includes international collaboration, technological oversight, ethical considerations, and compliance frameworks. Our goal is to deepen the understanding of the challenges and opportunities that AI governance entails, ultimately promoting technological development for the benefit of humanity.

We anticipate that this survey would serve as a valuable and thorough reference for researchers, policymakers, and industry practitioners, facilitating a better understanding of the current status and challenges of LLMs in relation to safety. By critically analyzing the shortcomings of existing research and policy practices, we aspire to inspire future endeavors in research, development and policymaking related to LLM safety.

\subsection{LLM Safety Definition}
In this survey, we distinguish LLM safety from security although LLMs could be used for aiding cybersecurity attacks or other security tasks. We refer LLM safety to as the responsible development, deployment, and use of LLMs to avoid causing unintended/intended harms. This definition involves ensuring that LLMs do not produce harmful outputs, such as biased, offensive, or unethical content, and safeguarding them from misuse in malicious activities, such as data manipulation or adversarial attacks. In contrast,  LLM security focuses on protecting LLM systems from external threats like hacking, denial-of-service attacks, or data breaches. In summary, LLM safety is more about the ethical and responsible usage of LLMs, while LLM security is concerned with defending LLM systems from technical threats \citep{DBLP:journals/corr/abs-2406-12934}.

\subsection{Paper and Source Selection}
We investigate LLM safety in the context of LLMs and generative AI, focusing on
publications within the fields of natural language processing (NLP) and AI. Main sources for our survey include venues such as ACL, EACL, NAACL, EMNLP, COLING, CoNLL, SIGIR, IJCAI, AAAI, NeurIPS, ICML, and ICDM, as well as unpublished research disseminated through preprint platforms like arXiv. Relevant papers are identified using keywords such as “AI Safety”, “Large Language Model”, “Safety”, and “AI Risks”. For specific subfields or subcategories, targeted keywords are used for literature search. For instance, terms like “Malicious Use”, “Misuse” and “Misinformation” are employed to retrieve studies related to Misuse. 

Additionally, as research on LLM safety is broad and complex, involving contributions from academia, industry and government, we also investigate technical reports and blog posts from AI companies (e.g., Anthropic, OpenAI, Google DeepMind, Meta, and Microsoft Research). Furthermore, we use keywords like “AI Governance” to include policy analysis and recommendations from international organizations and governmental institutions. The collected literature spans a wide range, from technical research to policy recommendations, emphasizing the collaborative efforts needed to address challenges associated with advanced AI systems. The diversity and breadth of information sources enhance our understanding of the field, contributing to a comprehensive overview of the current state of LLM safety.

\subsection{Related Work}
The rapid advancements in LLMs have drawn significant global attention to the importance of AI safety, thereby stimulating considerable interests within the research community and prompting efforts to survey the current state of this field. However, much of the recent literature has concentrated on specific aspects or levels of safety risks associated with LLMs. For instance, some studies have focused on reviewing alignment methods for LLMs \citep{DBLP:journals/corr/abs-2310-19852, DBLP:journals/corr/abs-2309-15025}, while others have primarily addressed catastrophic risks, such as potential misuse \citep{DBLP:journals/corr/abs-2306-12001}. A few surveys have investigated a relatively broader scope. \cite{DBLP:journals/corr/abs-2401-05778} approach the issue from the perspective of four fundamental modules of LLM systems (input module, language model module, toolchain module, and output module), analyzing potential risks associated with each module and discussing corresponding mitigation strategies. Our survey adopts a similar classification method which is based on different LLM stages (e.g., Data Processing, Pre-training, Post-training and Post-processing Stages) when summarizing approaches to address safety problems in specific areas, such as privacy and toxicity. However, we believe this classification method is insufficient to cover all safety-related issues. For example, the discussion of robustness against attacks and catastrophic misuse is not covered in \citep{DBLP:journals/corr/abs-2401-05778}, nor does it include specific governance proposals and policy recommendations. \cite{DBLP:journals/corr/abs-2407-18369} sequentially introduce the working principles of LLMs, research challenges of generative models, and classifications of alignment and safety in their review, while \cite{DBLP:journals/corr/abs-2408-12935} structure their framework for AI safety around three pillars: Trustworthy AI, Responsible AI, and Safe AI. In contrast, we provide a more detailed categorization system, dividing safety issues into several domains, each with multiple subdomains, which facilitates a clearer understanding and response to the safety challenges of LLMs. Additionally, we go beyond technical aspects of safety by investigating governance and policy, offering a more comprehensive perspective.

By offering well-defined subdomains and detailed evaluation techniques, this survey aims to provide practical guidance for safety researchers and practitioners focused on LLM-specific risks. Such focused methodology ensures that safety measures and strategies proposed are directly applicable to the emerging issues within LLM deployment. Moreover, policy recommendations and governance suggestions investigated in this survey aim to bridge the gap between safety technologies and regulation frameworks. These recommendations facilitate the creation of more adaptive and robust regulatory strategies, ensuring that policies remain responsive to the rapid advancements in LLM technology while addressing potential ethical and societal concerns.

%% file: sections/Taxonomy.tex
\section{Taxonomy}

\begin{figure}[t]
    \centering
    \includegraphics[width=0.995\textwidth]{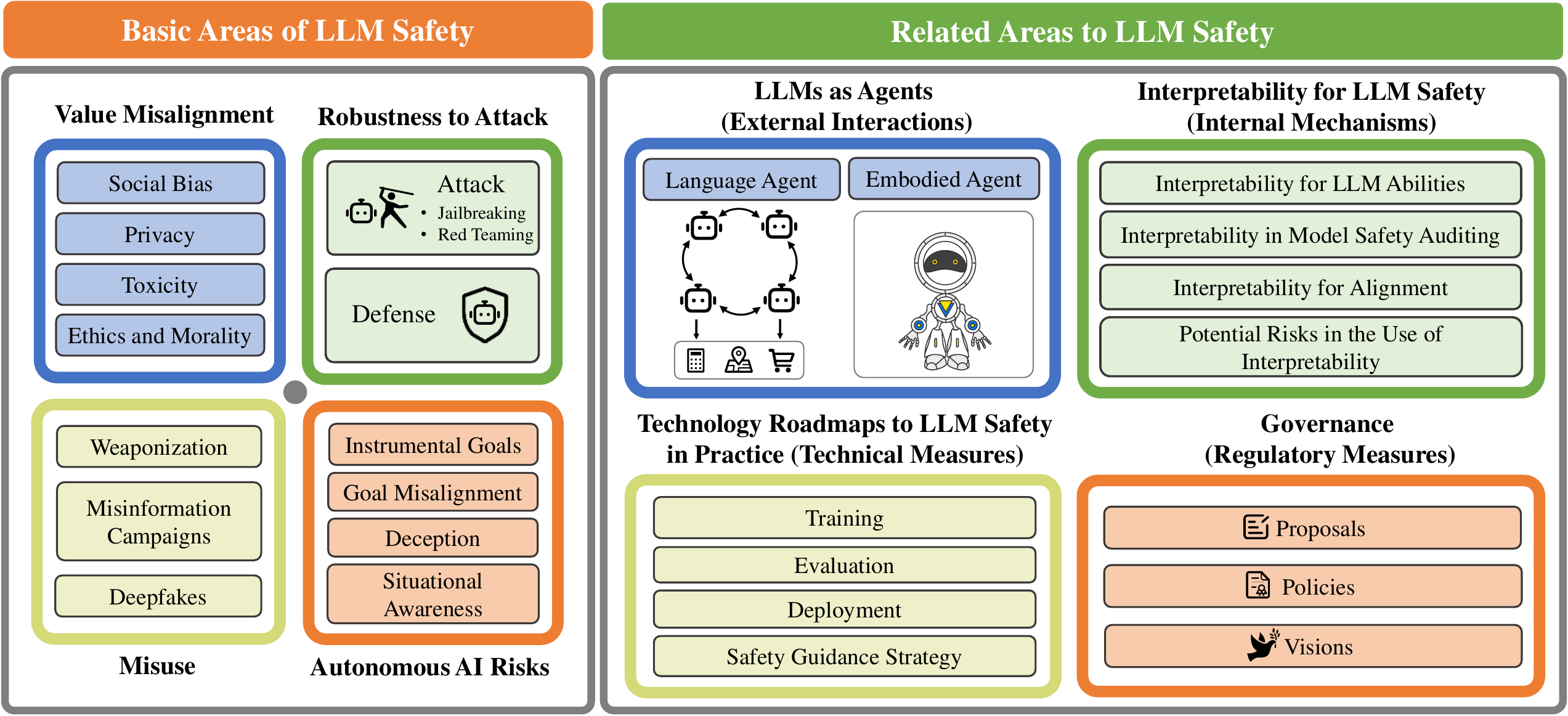}
    \caption{The taxonomy of LLM safety proposed in this survey.
}
    \label{fig:all}
\end{figure}

This survey aims to provide a systematic organization of a wide variety of safety concerns, risks, and strategies associated with LLMs. By identifying and categorizing these risks, we offer a well-structured taxonomy for understanding the broad spectrum of challenges raised by the development and deployment of LLMs. The taxonomy, illustrated in Figure \ref{fig:all},  structures the current landscape of LLM safety into two dimensions: basic areas of LLM safety which cover key risk areas of LLMs, and related areas to LLM safety which identify essential areas closely related to LLM safety.

\subsection{Basic Areas of LLM Safety}

We identify four key risk areas of LLMs in the first dimension: Value Misalignment, Robustness to Attack, Misuse and Autonomous AI Risks, as illustrated in Figure \ref{fig:all}. For each key risk area, we further identify subdomains, shedding light on the multifaceted challenges of LLM safety and strategies for evaluating and mitigating associated risks.

\begin{itemize}
    \item \textbf{Value Misalignment} (Section \ref{section:ValueMisalignment}): This section delves into the multifaceted landscape of LLM safety stemming from misalignment between LLMs and human intentions, values, and expectations. It includes four core sub-realms: social bias, privacy, toxicity, and ethics and morality. By systematically analyzing their impacts, origins, evaluation methods, and mitigation strategies, this section provides a comprehensive understanding of the key concerns associated with these sub-realms.
    \item \textbf{Robustness to Attack} (Section \ref{section:RobustnessToAttack}): This risk area explores the robustness of LLMs against adversarial attacks, focusing on jailbreaking techniques and red teaming methods, which examines various strategies for bypassing safety mechanisms, and manual or automated adversarial testing for identifying vulnerabilities of LLMs. Furthermore, it discusses defense strategies against these threats, which include external safeguards designed to protect LLMs from malicious inputs, as well as internal protections that involve modifying the LLMs themselves to enhance their resilience. These strategies are essential for improving the safety and robustness of LLMs, though challenges remain in balancing effectiveness with model complexity.
    \item \textbf{Misuse} (Section \ref{section:CatastrophicMisuse}): For this risk area, we focus on reviewing severe threats posed by LLMs when exploited by malicious actors, which elucidate multiple facets of human society and public safety. On the one hand, LLMs could be used for a wide range of illegal aims, including facilitating cyberattacks and producing biological, chemical, and nuclear weapons that threaten human safety. On the other hand, the generation of erroneous or misleading texts by LLMs can be exploited to disseminate harmful misinformation on social media and news platforms, significantly influencing public opinion, political processes, and societal trust. Additionally, the remarkable capabilities of state-of-the-art LLMs to generate realistic audio and video content from text have intensified ethical, social, and safety concerns surrounding deepfake technology, which has historically had detrimental effects on society.
    \item \textbf{Autonomous AI Risks} (Section \ref{section:AutonomousAIRisk}): In addition to the above three key risk areas, we further explore growing concerns surrounding the development of autonomous AI with LLMs. As LLMs advance toward human-level capabilities, concerns on societal and ethical risks associated with autonomous AI have reemerged, particularly regarding risks of advanced LLMs deployed online or in autonomous or semi-autonomous environments. Such risks include but are not limited to pursuing a number of convergent instrumental goals (e.g., self-preservation, power-seeking) \citep{DBLP:conf/aaai/Benson-TilsenS16}, deception and situational awareness. Theoretically formalizing/validating such risks and empirically detecting and evaluating them pose significant challenges for frontier AI/LLM safety..
\end{itemize}

\subsection{Related Areas to LLM Safety}

When granted the autonomy to use tools, perform actions, and interact with the environment, LLM-powered agents can demonstrate highly efficient and automated task-solving capabilities. However, this autonomy also brings the risk of making unpredictable or uncontrollable decisions. Beyond external aspects of LLMs (e.g., interactions between LLM-driven agents and environments), the prevalence and severity of various safety risks have prompted investigations into the internal mechanisms of LLMs. These efforts aim to address critical issues related to transparency and interpretability regarding both LLM capability and safety, which arise due to the black-box nature of LLMs.

In the perspective of safe deployment and application, many AI/LLM companies and research institutions allocate a lot of resources to implement various safety techniques to safeguard deployed LLMs from unsafe behaviors, e.g., generating biased, toxic, or immoral responses. In addition to technical measures, as LLMs increasingly permeates various sectors, the establishment of a comprehensive and robust governance framework at a high level has become an imperative. Such a framework should not only ensure that the development and deployment of LLMs adhere to globally recognized ethical standards but also foster international cooperation and regulatory coordination to achieve mutual benefits and shared prosperity in global AI technology.

In light of these considerations, we further discuss four important areas related to LLM safety: Agent Safety, Interpretability for LLM Safety, Technology Roadmaps / Strategies to LLM Safety in Practice and Governance, in addition to the key risk areas of LLMs.

\begin{itemize}
    \item \textbf{Agent Safety} (Section \ref{section:AgentSafety}) examines the risks associated with two primary categories of LLM-powered agents: language agents and embodied agents. While these agents offer tremendous potential for automation and innovation across various sectors, they also present a range of concerns, including the potential for malicious use, misalignment with human values, privacy invasion, and unpredictable behaviors. This section delves into these risks in detail, exploring their implications for society, economy, and individual privacy. Additionally, it outlines mitigation strategies and resources aimed at enhancing the safety and reliability of LLM-driven agents. As LLMs continue to evolve, understanding and addressing these risks become crucial for ensuring the responsible development and deployment of AI technologies.
    \item \textbf{Interpretability for LLM Safety} (Section \ref{section:InterpretabilityForLLMSafety}) emphasizes the role of interpretability in enhancing the safety of LLMs used in critical fields. It highlights how interpretability helps make LLMs' decision-making processes transparent, enabling better evaluation and control. Key benefits include improving performance, addressing biases, and ensuring safe outputs. It introduces a taxonomy of interpretability for LLM safety, covering understanding model capabilities, safety auditing, and aligning LLMs with human values. The risks of interpretability research are also discussed, including the dual-use of technology, adversarial attacks, misunderstanding or over-trusting explanations, and the potential for interpretability to accelerate uncontrollable risks.
    \item \textbf{Technology Roadmaps / Strategies to LLM Safety in Practice} (Section \ref{section:TechnicalRoadmapToLLMSafety}) delineates the current landscape of safety measures and strategies employed by various prominent LLMs. This section elaborates and compares the safety roadmaps implemented by key industry players, including OpenAI, Anthropic, Baidu, Google DeepMind, Microsoft, 01.AI, Baichuan, Tiger Research, Alibaba Cloud, DeepSeek-AI, Mistral AI, Meta, Shanghai AI Laboratory and Zhipu AI, to ensure the reliability and safety of LLMs in practical applications. Additionally, it also discusses the contributions of certain research institutions that, despite not releasing LLMs, are actively engaged in AI safety research and development. 
    \item \textbf{Governance} (Section \ref{section:Governance}) delves into the multifaceted realm of AI governance, exploring proposals, policies, and visions that collectively shape the future of AI development and deployment. As AI continues to rapidly evolve and integrate into various aspects of society, the need for comprehensive and effective governance frameworks becomes increasingly critical. By analyzing current policies, comparing different approaches, and considering long-term visions, this section aims to provide a thorough understanding of the challenges and opportunities in governing AI technologies for the benefit of humanity. From international cooperation initiatives to technical oversight mechanisms, from ethical considerations to compliance challenges, we examine the complex landscape of AI/LLM regulation. 
\end{itemize}

%% file: sections/Value_Misalignment.tex
\section{Value Misalignment}
\label{section:ValueMisalignment}

As LLMs become increasingly sophisticated and ubiquitous, their potential for profound societal impact has brought critical safety considerations to the forefront of attention across various domains. This section provides a comprehensive survey on the safety concerns emerging from potential value misalignment in LLMs, which represents a multifaceted phenomenon where LLMs generate outputs that diverge from human ethical standards, societal norms, and fundamental moral principles. This misalignment manifests through four critical dimensions: social bias, privacy, toxicity, and ethics and morality.

\subsection{Social Bias}
Bias and fairness in LLMs, have emerged as critical areas of concern. They often manifest in various forms, perpetuating stereotypes or unfairly disadvantaging certain social groups. Social bias is one of the most pervasive forms of bias in LLMs, and it can take on multiple dimensions, from derogatory language and stereotypes to prejudice across different social and cultural groups. In the following sections, we will explore this issue in more depth, beginning with a discussion on the definition of social bias and its safety implications, followed by an analysis of how social bias manifests throughout the LLM lifecycle and elaborations on methods available for mitigation as well as evaluation frameworks necessary to ensure fairness and safety.

\input{taxonomies/VM-Social_Bias_Taxonomy}

\subsubsection{Definition and Safety Impact}

Social bias in LLMs refers to the reinforcement of stereotypes and inequalities through language, which can lead to harmful outcomes for specific social groups. Such biases manifest in multiple ways, from the use of derogatory language and perpetuation of negative stereotypes to LLM performance disparities across language and social groups. For instance, the Social Categories and Stereotypes Communication (SCSC) framework \citep{article} demonstrates how language can maintain social categories, which reinforces social biases. A common and troubling example is the association of the term ``Muslim'' with terrorism in LLMs, which reflects and amplifies anti-Muslim biases in these models \citep{DBLP:conf/aies/AbidF021}. Such biases, when embedded in widely used systems, pose a significant safety risk by perpetuating harmful narratives that could increase societal divisions and marginalization.

From a safety perspective, the impact of social bias goes beyond just harmful language. LLMs can generate toxic content, such as hate speech or violent language, which not only harms targeted individuals or groups but also creates broader societal risks \citep{DBLP:conf/aies/DixonLSTV18}. This introduces direct safety concerns, as the propagation of hate speech or discriminatory content can exacerbate tensions, leading to real-world harms. Furthermore, biased LLM outputs in sensitive applications like healthcare, education, or law can result in biased decisions that affect the welfare and rights of marginalized groups, undermining both societal safety and fairness. For instance, in downstream NLP tasks like text generation and question answering, gender and racial biases can be reflected in texts generated by LLMs, requiring detoxification techniques and debiasing strategies to ensure the fairness and safety of these outputs \citep{DBLP:conf/icml/LiangWMS21, DBLP:conf/emnlp/ShengCNP19, DBLP:conf/iclr/YangYLL023}.

Another critical safety issue arises from performance disparities of LLMs. LLMs often exhibit significant performance gaps when handling language variations used by different social or ethnic groups. For example, models may underperform when processing African American English compared to standard American English, revealing racial biases embedded within natural language processing systems, particularly in social media contexts \citep{DBLP:journals/corr/BlodgettO17}. These disparities not only reduce the effectiveness of LLM applications but can also lead to unsafe outcomes when applied in critical areas such as law enforcement, healthcare, or hiring, where fairness and accuracy are paramount. When deployed LLMs disproportionately benefit dominant social groups and marginalize others, the safety and fairness of these systems are fundamentally compromised \citep{DBLP:conf/fat/BenderGMS21}. In addition, biases in machine translation can further exacerbate such disparities, leading to misinterpretation of linguistic nuances, particularly for underrepresented languages \citep{mechura-2022-taxonomy}.

In addition to biased outputs and performance disparities, social bias also impacts resource allocation and access to opportunities. Decisions informed by biased models can lead to unfair treatment in job recommendations, loan approvals, and healthcare access, directly affecting the well-being of certain groups \citep{DBLP:journals/firstmonday/Ferrara23a}. This introduces a form of indirect discrimination, where seemingly neutral AI processes result in unequal outcomes, threatening the safety and equity of the affected individuals or communities. Moreover, neural ranking models in information retrieval may reinforce gender biases in search results, highlighting the need for adversarial mitigation strategies to reduce unfair outcomes and promote safety in these contexts \citep{DBLP:conf/sigir/RekabsazS20,DBLP:conf/sigir/RekabsazKS21}. Techniques like Orthogonal Subspace Correction and Rectification (OSCaR) \citep{DBLP:conf/emnlp/Dev0PS21} have been employed to address biases in natural language inference tasks, ensuring fairness and minimizing social harm.

In conclusion, the pervasive social bias in LLMs presents both direct and indirect safety risks. Addressing these biases is essential for maintaining the integrity and trustworthiness of AI systems, especially in contexts where fairness and safety are critical to societal well-being. Comprehensive strategies must be developed to ensure that the presence of social bias does not compromise the safe deployment of these systems.

\subsubsection{Social Bias in the LLM Lifecycle}
Bias can infiltrate various stages of the LLM lifecycle, from training data to deployment, each contributing to the overall safety of the final model.
\paragraph{Training Data}  
LLMs could be trained on non-representative samples, which limits their ability to generalize across different social groups. For instance, biases in training data have led to skewed results in hate speech detection, particularly against African American Vernacular English (AAVE) \citep{DBLP:journals/corr/abs-2309-00770}. Even well-curated datasets may reflect historical and structural biases, highlighting the need for unbiased data and diverse annotators in the data collection process.\footnote{https://mostly.ai/blog/data-bias-types}

\paragraph{Training and Inference}
The choice of training objectives, such as prioritizing accuracy over fairness, influences model behaviors. Bias can also emerge during inference, where decoding strategies in text generation or document ranking can perpetuate biases \citep{DBLP:journals/corr/abs-2309-00770}.

\paragraph{Evaluation}
Benchmark datasets often fail to represent the diversity of LLM users, which can lead to optimization for only a subset of the population. The selection of evaluation metrics, such as those masking performance disparities between social groups, further exacerbates this issue \citep{DBLP:journals/corr/abs-2309-00770, DBLP:conf/icml/SorensenMFGMRYJ24, DBLP:journals/corr/abs-2406-14903}.

\paragraph{Deployment}
In applications like chatbots, LLMs can reproduce and amplify biases through interaction with users, leading to reinforced stereotypes and power imbalances. Biases are also evident in LLM-powered personal assistants and medical applications, where recommendations and diagnoses may be skewed by social biases \citep{DBLP:journals/npjdm/MittermaierRK23a,seyyed2021underdiagnosis,obermeyer2019dissecting,DBLP:journals/npjdm/YangSEYC23}.

\begin{figure}[t]
    \centering
    \includegraphics[width=1.0\textwidth]{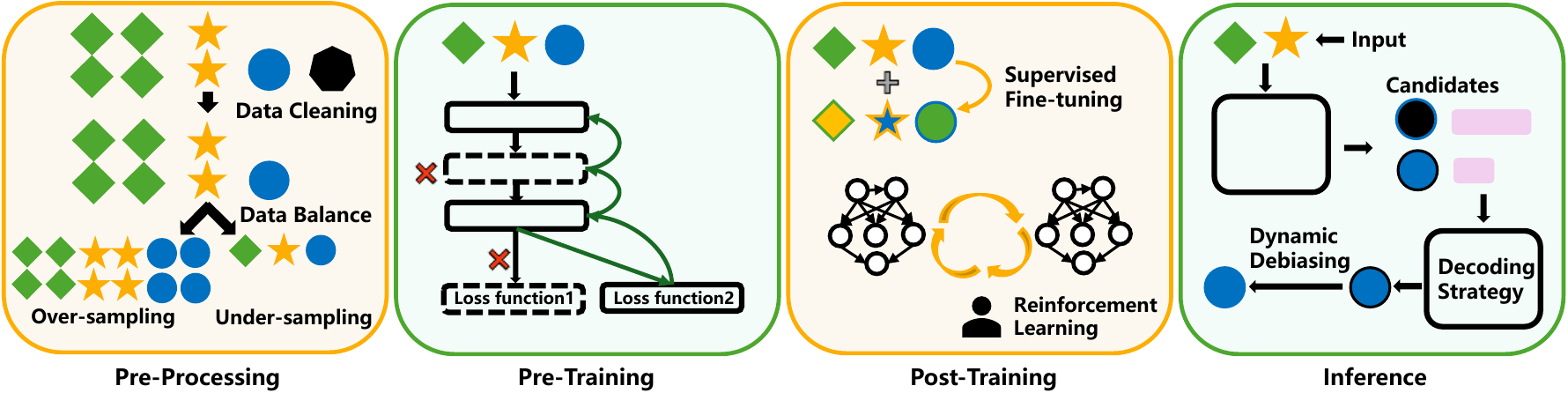}
    \caption{Methods for mitigating bias throughout the four stages of the LLM lifecycle. Pre-processing mitigation involves data cleaning and balancing through over-sampling and under-sampling. In the pre-training stage, strategies such as biased neurons editing, architecture modifications, and adjustments to multiple loss functions can be explored. In the post-training phase, techniques such as counterfactual data augmentation followed by supervised fine-tuning (SFT), adversarial reinforcement learning, and alignment strategies are applied to mitigate bias. Inference-phase mitigation interventions during word decoding (e.g., dynamic debiasing of candidate word lists), unbiased decoding strategies, and real-time monitoring of final outputs.
}
    \label{fig:bias}
\end{figure}

\subsubsection{Methods for Mitigating Social Bias}
Social bias mitigation in LLMs can be approached through multiple strategies across different phases in the lifecycle of LLMs. This section explores four distinct stages in which bias mitigation techniques can be applied: pre-processing, pre-training, post-training, and inference. Each phase offers unique opportunities for intervention, targeting different aspects of bias within the data and the model itself, as summarized in Figure \ref{fig:bias}.

\paragraph{Pre-Processing Mitigation}
Mitigating bias in LLMs often begins with pre-processing strategies, where techniques like data balancing and demographic-aware fine-tuning are employed to ensure that training samples are more representative and diverse. These methods help address selection biases by broadening the range of data used during training \citep{DBLP:journals/corr/abs-2101-11718,DBLP:conf/emnlp/SmithHKPW22}. Additionally, data cleaning plays a crucial role in this stage by removing biased data and synthesizing unbiased samples, which are essential steps to minimize the introduction of biases into LLMs \citep{DBLP:conf/acl/ParrishCNPPTHB22,DBLP:conf/coling/HuangX24}.

\paragraph{Pre-Training Mitigation}
During the pre-training phase, various strategies can be applied to mitigate bias for LLMs. Modifying the neural architecture of an LLM, such as adjusting layers and encoders, can help reduce inherent biases as the LLM learns from data \citep{DBLP:conf/emnlp/LiKKSS20}. Another approach is to adjust the loss function to prioritize fairness over perplexity, ensuring that LLM outputs are more equitable. Selective parameter updates allow for fine-tuning specific parts of a model to control bias without necessitating a full retraining process, making this an efficient method for bias mitigation \citep{DBLP:conf/birthday/LuMWAD20,DBLP:conf/ijcnlp/GarimellaMA22}. Regular audits of model algorithms and architectures ensure that these mitigation strategies are continuously effective and that new biases are not introduced during updates or retraining \citep{DBLP:conf/aaai/ZayedPMPSC23}.

\paragraph{Post-Training Mitigation}
Post-training is crucial for reducing bias in LLMs after the pre-training phase. One common approach is supervised fine-tuning, where a pre-trained model is fine-tuned on more balanced and representative data to reduce bias in specific tasks or domains \citep{DBLP:conf/naacl/DevlinCLT19, DBLP:conf/coling/HuangX24a}. Fine-tuning on curated or debiased datasets can help mitigate inherent biases learned from the original training data. Adversarial debiasing is another effective technique where an adversarial model is trained to detect and neutralize biases in LLM outputs \citep{DBLP:conf/aies/ZhangLM18}, forcing LLMs to generate more fair and balanced representations. 

Post-training mitigation also includes techniques such as bias fine-pruning, which selectively removes biased neurons or layers from LLMs \citep{DBLP:journals/corr/abs-2106-07849}, and counterfactual data augmentation, where modified data with altered social attributes (e.g., swapping gender or race identifiers) are used to fine-tune LLMs, helping them learn to treat various social groups more equitably \citep{DBLP:conf/birthday/LuMWAD20}. Furthermore, knowledge distillation allows for transferring the knowledge of a biased model to a smaller, more task-specific model, with bias reduction constraints applied during the distillation process \citep{DBLP:journals/mta/LinWCS21}. Finally, fairness-constrained fine-tuning introduces fairness objectives directly into the loss function of LLMs, ensuring that the optimization process explicitly balances the next token prediction accuracy with fairness considerations \citep{DBLP:conf/www/ZafarVGG17}. These post-training techniques provide flexible and efficient solutions for mitigating bias, enhancing both the fairness and safety of LLMs without the need for complete retraining.

\paragraph{Inference Mitigation}
Bias mitigation efforts can extend into the inference stage, where specific techniques are used to adjust LLM outputs dynamically. Modifying decoding strategies, such as adjusting the probabilities for next-token or sequence generation, can help reduce bias in real-time without further training \citep{DBLP:conf/emnlp/LauscherLG21}. Additionally, adjusting attention weights during inference allows for the modulation of the model's focus, promoting fairness in generated outputs \citep{DBLP:conf/wsdm/ParkCYK23}. Debiasing components, like AdapterFusion \citep{DBLP:conf/ltedi/GiraZL22}, can dynamically address biases as they arise during the generation process of LLMs. Techniques such as gradient-based decoding enable the neutral rewriting of biased or harmful text, ensuring that final outputs align with fairness goals \citep{DBLP:journals/corr/abs-2207-02463}. Moreover, online monitoring of model outputs is necessary to detect emerging biases, allowing for timely interventions and adjustments to maintain model integrity.

\paragraph{Interpretability and Transparency}
In addition to the aforementioned bias mitigation methods used in the various stages of LLMs, enhancing transparency of LLMs is also critical for bias mitigation. By providing clear rationales for generated texts and detailing the sources behind these decisions, LLMs can offer deep insights into their decision-making processes to users. This transparency not only helps in understanding how biases may have influenced the outputs but also empowers users to question and challenge any biased results, thereby promoting more ethical and fair AI systems \citep{DBLP:conf/acl/ChungKA23}.

\subsubsection{Evaluation}

\paragraph{Metrics}
A number of metrics have been developed to quantify and analyze biases across different aspects of LLMs. Embedding-based metrics like WEAT \citep{caliskan2017semantics} and SEAT \citep{DBLP:conf/naacl/MayWBBR19} are commonly used to measure biases in word and sentence embeddings. These metrics work by comparing the distances between biased and unbiased representations in the form of embeddings, providing insights into how deeply biases are ingrained in language representations learned by LLMs \citep{panch2019artificial,DBLP:journals/corr/abs-2305-17493}. Probability-based metrics offer another approach to bias evaluation, focusing on the probabilities assigned to tokens or sequences during model inference. For example, masked token probabilities and pseudo-log-likelihoods (e.g., Crows-pair \citep{DBLP:conf/emnlp/NangiaVBB20}) are used to assess the likelihood of generating biased tokens, thereby revealing underlying biases in the decision-making process of language models. Finally, generated text-based metrics evaluate biases in the actual outputs generated by LLMs \citep{DBLP:conf/aies/GuoC21,DBLP:journals/corr/abs-2010-06032}. These metrics often involve comparing the distribution of tokens associated with different social groups or using classifiers like Perspective API \footnote{\url{https://perspectiveapi.com}} to detect and quantify harmful content. Additionally, lexicon-based analyses can be applied to generated text to identify and score biased language, further contributing to a comprehensive understanding of bias in LLM outputs.

\paragraph{Benchmarks}
A range of benchmarks has been created to assess bias in LLMs, providing essential data for evaluating fairness across different tasks. Masked token benchmarks like Winogender \citep{DBLP:conf/naacl/RudingerNLD18}, Winobias \citep{DBLP:conf/naacl/ZhaoWYOC18}, and GAP \citep{DBLP:journals/tacl/WebsterRAB18} are designed to test bias by predicting the most likely words in masked sentences, helping to reveal gender and other social biases embedded in models \citep{DBLP:conf/emnlp/NangiaVBB20,DBLP:conf/emnlp/RajpurkarZLL16}. Unmasked sentence benchmarks, such as Crows-pair \citep{DBLP:conf/emnlp/NangiaVBB20} and Redditbias \citep{DBLP:conf/acl/BarikeriLVG20}, evaluate bias by comparing the likelihood of generating biased versus neutral content, offering insights into the model's tendencies in real-world text generation scenarios. Additionally, sentence completion benchmarks like RealToxicityPrompts \citep{DBLP:conf/emnlp/GehmanGSCS20} and BOLD \citep{DBLP:conf/fat/DhamalaSKKPCG21} focus on assessing bias in the continuation of sentences, particularly in the context of generating toxic or harmful language. Finally, question answering benchmarks such as Unqover \citep{DBLP:journals/corr/abs-2010-02428}, BBQ \citep{DBLP:conf/acl/ParrishCNPPTHB22} and CBBQ \citep{DBLP:conf/coling/HuangX24} are utilized to analyze how LLMs select answers, specifically measuring biases in decision-making during question-answering tasks. These benchmarks collectively provide a comprehensive toolkit for identifying and evaluating bias in LLMs.

\subsubsection{Future Directions}
As LLMs continue to evolve and become integrated into critical decision-making processes, addressing bias and ensuring safety will require multifaceted and proactive approaches. Below are several future directions that can help advance efforts in mitigating social bias and enhancing safety in LLMs.

\paragraph{Multi-Objective Optimization for Fairness and Safety}
Balancing performance, fairness, and safety in LLMs is challenging, especially when optimizing models for token prediction accuracy might inadvertently amplify biases. A promising future direction is the development of multi-objective optimization frameworks that prioritize both fairness and safety alongside performance metrics. By modifying loss functions and training objectives, models can be optimized to reduce bias while still maintaining high accuracy. Furthermore, ongoing research into fairness-aware algorithms that incorporate societal safety as a key objective could ensure that LLMs are better aligned with ethical and legal standards.

\paragraph{Interdisciplinary Collaboration and Ethical AI Governance}
Effective bias mitigation and safety in LLMs require collaboration across different disciplines, including machine learning, ethics, law, and social sciences. Developing frameworks for cross-disciplinary collaboration, where technical experts and policy makers work together, will be crucial for ensuring that AI systems are deployed responsibly and ethically. Moreover, ethical AI governance frameworks that provide clear guidelines and regulations for bias detection, model auditing, and risk mitigation should be established. These frameworks could ensure that LLM developers are held accountable for the societal impacts of their systems, fostering greater transparency and trust in AI systems.

\paragraph{Integration of Societal and Cultural Awareness}
Future LLMs should aim to integrate societal and cultural awareness into their decision-making process. By embedding knowledge of diverse social and cultural contexts into LLMs, they can become more sensitive to the nuances of different communities and avoid generating outputs that perpetuate harmful stereotypes. This requires training LLMs not only on diverse datasets but also on datasets that capture the subtleties of different cultural and social practices, languages, and dialects. Incorporating culture-level context into LLMs could further enhance their ability to generate fair and unbiased content, improving overall safety in global applications.

\subsection{Privacy}
Protecting the privacy of LLMs is a fundamental aspect of maintaining their overall safety. We hence provide an overview of the current status of privacy issues in LLMs (e.g., privacy leakage in LLMs, privacy protection methods for LLMs).

\input{taxonomies/VM-Privacy_Taxonomy}

\subsubsection{Preliminaries}
Privacy protection has always been a key issue in the field of AI as privacy leakage could lead to serious consequences. Many legal and regulatory requirements \citep{rigaki2020survey,mireshghallah2020privacy,sousa2023keep,guo2022threats} have been established. For example, the European Union has introduced the General Data Protection Regulation (GDPR), which sets strict guidelines for the collection, transmission, storage, management, processing, and deletion of data, and imposes severe penalties on companies that violate privacy regulations.

\begin{figure}[t]
    \centering
    \includegraphics[width=0.75\textwidth]{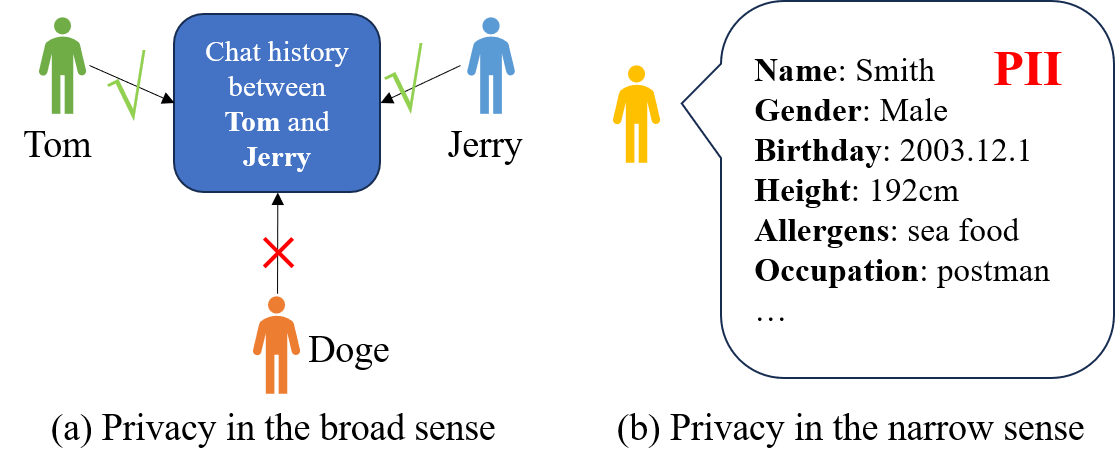}
    \caption{Privacy information can be divided into two categories: privacy in the narrow sense, which usually refers to personal identity information; and privacy in the broad sense, which usually depends on the owner of the information to decide to whom it is private.
}
    \label{fig:priv_type}
\end{figure}

\paragraph{Types of Privacy Information}
Privacy is generally determined by the subject of the information, which is related to its context and discourse \citep{brown2022does}. As shown in Figure~\ref{fig:priv_type} (a), an example of privacy in the broad sense is: the chat history between Tom and Jerry is not private to them, but it is confidential to others.
For convenience in research, researchers often adopt a narrow definition of privacy \citep{sousa2023keep}, such as considering personal identity information (PII) as the privacy illustrated in Figure~\ref{fig:priv_type} (b), including name, gender, phone number and other related information.

\paragraph{Safety Impact of Privacy Leakage}
LLMs are increasingly being integrated into many areas that penetrate into personal life and work, inevitably involving personal data. Privacy leaks may pose significant risks to the safety of life and property of users. For example, leakage of personal family information in smart cities may lead to harassment of the owner \citep{pramanik2023overview}, digital portraits may be forged or abused \citep{le2023safety}, leakage of in-vehicle systems may lead to interception of the owner's location data \citep{yoshizawa2023survey}, and medical records of patients in medical systems may be leaked \citep{paul2023digitization}, leading to health-related discrimination.

\subsubsection{Sources and Channels of Privacy Leakage}
We then discuss the sources of private risks and the channels of privacy leakage, aiming to provide a deep understanding of the underlying causes of privacy breaches in large language models.

\paragraph{Sources of Privacy Risks}

We roughly identify two major sources of privacy risks for LLMs. 
The first is \textbf{training data from the Internet} \citep{piktus2023roots,li2023multi}. Web-crawled data have been widely used as a major data source to train LLMs.  
However, data crawled from the Internet inevitably contain non-public, private, or sensitive information. The cost of cleaning such data is high and it is impossible to guarantee complete removal of sensitive information. 
The second is \textbf{the memorization capability} of LLMs \citep{carlini2019secret, carlini2021extracting}: researchers have found that LLMs can randomly memorize training samples and produce high-confidence outputs for certain samples. Worse yet, such memorization ability tends to increase as model size increases. Even if most LLMs are trained only for one epoch, they may still memorize a certain proportion of the data \citep{carlini2022quantifying}.

\begin{figure}[t]
    \centering
    \includegraphics[width=0.75\textwidth]{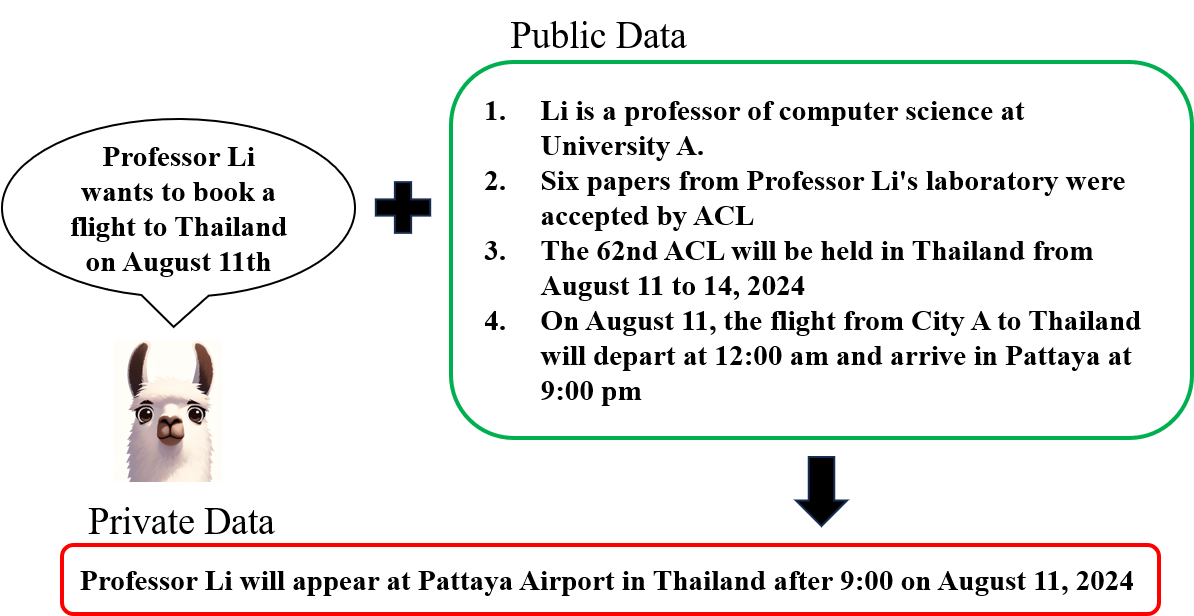}
    \caption{An example of Inference Leakage. An attacker can use public data to make a large language model infer private data.
}
    \label{fig:infer}
\end{figure}

\paragraph{Channels of Privacy Leakage}
We summarize three paths of privacy leakage in LLMs. 
\begin{itemize}
\item \textbf{Direct Exposure}: This usually refers to the direct exposure of private information or medium containing such information. This is the most direct and easiest threat to detect. Users may inadvertently provide sensitive personal data through direct queries to LLMs, which can lead to immediate privacy breaches. For example, Samsung Electronics has experienced inadvertent leakage of sensitive company data through interactions with ChatGPT \citep{yan2024protecting}.

\item \textbf{Privacy Attack}: LLMs are often targeted by active attacks that aim to extract private information. These attacks exploit vulnerabilities in the training data, architectures, or fine-tuning process of LLMs. The success of privacy attacks, such as jailbreak prompt attacks \citep{li2023multi}, adversarial attacks \citep{DBLP:journals/corr/abs-2307-15043}, and backdoor attacks \citep{yao2024poisonprompt}, relies on weaknesses in LLMs, which allow attackers to bypass thier built-in privacy safeguards. For example, in a jailbreak attack \citep{li2023multi}, adversaries design specific prompts that manipulate an LLM into revealing sensitive information or generating responses that would otherwise be restricted. In adversarial attacks, small perturbations are made to inputs \citep{DBLP:journals/corr/abs-2307-15043}, which could lead LLMs to produce unintended or confidential outputs without being detected. Additionally, backdoor attacks \citep{yao2024poisonprompt} involve poisoning training data with malicious examples, which may cause LLMs to exhibit specific behaviors when triggered by certain inputs. These attacks can compromise the confidentiality, integrity, or availability of private information, thereby posing serious risks to data privacy. Moreover, adversaries can exploit these vulnerabilities to reverse-engineer model outputs, reconstruct private training data, or infer private attributes from LLM-generated responses.

\item \textbf{Inference Leakage}: Due to the ever-growing reasoning capabilities of LLMs, indirect privacy leaks would be possible. 
Even short and scattered information could lead to privacy leakage after being inferred by LLMs \citep{staab2023beyond,li2024privacy}. 
For example, as shown in Figure~\ref{fig:infer}, Professor Li is going to book a flight, and an LLM can use some public information to infer Professor Li's specific travel itinerary. This method of prompting the model to infer personal privacy represents a new systemic risk.

\end{itemize}

\subsubsection{Privacy Protection Methods}
Privacy protection methods for LLMs can be roughly divided into three categories according to the life cycle of LLMs \citep{guo2022threats,sousa2023keep}.

\paragraph{Privacy Protecting at the Data Processing Stage} Most of these methods aim to remove sensitive information at the data processing stage. Traditional methods like de-identification \citep{meystre2014text} and anonymization \citep{majeed2020anonymization} are widely used to achieve this goal. 
For instance, names, addresses, and other PII can be generalized or replaced with pseudonyms or placeholders, making it hard to identify individuals while still preserving the dataset's structure and inner dependencies. Additionally, aggregation techniques can be applied to reduce the granularity of data, such as grouping inference queries by day or week, instead of storing individual query details. This reduces the risk of re-identification and limits the potential privacy exposure.
In addition to de-indentification and anonymization, \citet{lee2021deduplicating} find that removing duplicate data from pre-training corpora can effectively reduce LLMs' memorization of training data. Although filtering or reformulating private data is a direct way to eliminate privacy risks, it is still difficult to completely remove private or sensitive data.

\paragraph{Privacy Protecting at the Pre-training or Fine-tuning Stage} A variety of methods are used at in the pre-training or fine-tuning stage to reduce the degree to which LLMs memorize training data. The differential privacy gradient optimization method reduces the probability of model memorization by adding noise to the gradient of private data \citep{li2021large, wu2022adaptive}, but such methods can negatively impact of LLMs on downstream tasks. During the model fine-tuning phase, alignment techniques are also applied to enable LLMs to refuse to respond to queries involving privacy issues. However, it remains challenging to balance user experience and privacy protection \citep{staab2023beyond}.

\paragraph{Privacy Protecting at the Model Pre-deployment Stage} Methods that adjust model parameters by editing or light retraining are proposed to protect privacy during the pre-deployment stage. Machine unlearning methods involve light parameter retraining to help LLMs forget private information \citep{eldan2023s, chen2023unlearn, yao2023large}. These methods focus on selectively removing specific knowledge from an LLM, especially when the LLM has learned sensitive data during training. By performing light retraining, these methods allow LLMs to ``unlearn'' specific patterns related to sensitive data without the need for extensive retraining, effectively reducing the risk of privacy breaches while maintaining the overall performance of LLMs.
Model editing methods \citep{wu2023depn, chen2024learnable, wu2024mitigating} involve identifying parameters that are related to private information in LLMs and reducing the probability of outputting private data by editing or replacing these parameters. These techniques are particularly useful for mitigating the risk of unintended information leakage, such as when a model inadvertently recalls sensitive details from its training data \citep{chen2024learnable}. By editing the parameters of a model, these methods target specific areas where sensitive information is stored, adjusting them to ensure that the model no longer outputs such information. In some cases, this can involve fine-tuning certain parameters, while in others, it may include replacing them with non-sensitive alternatives to preserve the functionality of LLMs without compromising privacy \citep{wu2024mitigating}.

\subsection{Toxicity}
Toxicity in LLMs refers to the generation of harmful or offensive content. As LLMs are increasingly integrated into sensitive domains and everyday applications, mitigating toxicity has become a critical concern to ensure safe and responsible deployment. In this section, we explore the methods for toxicity mitigation and the evaluation benchmarks commonly used to assess and reduce the toxicity of LLMs.

\input{taxonomies/VM-Toxicity_Taxonomy}

\subsubsection{Definition and Safety Impact}

Building upon the discussion of bias, toxicity in LLMs focuses on the generation of content that is explicitly harmful, offensive, or inappropriate. This includes hate speech, abusive expressions, threats, or any content that can directly inflict emotional or psychological harm on individuals or groups. While bias often refers to the implicit reinforcement of stereotypes and systemic inequalities \citep{DBLP:journals/corr/abs-2309-00770}, toxicity is characterized by its overtly harmful nature, which can escalate hostility, disrupt social harmony, and threaten safety.

Toxicity in LLMs arises through various mechanisms.  First, it can stem from the presence of toxic language in training data, where harmful patterns are learned and subsequently reproduced \citep{DBLP:journals/corr/abs-2309-15025}.  Second, the generalization capabilities of LLMs can lead to unintended toxicity, as LLMs may amplify offensive patterns in contexts where these patterns are not originally present.  Third, interactive generation processes—particularly in unsupervised or conversational settings—can produce toxic outputs due to ambiguous prompts or adversarial user interactions \citep{DBLP:journals/corr/abs-2402-13926}.

Addressing toxicity in LLMs is critical not only for ensuring socially responsible AI but also for safeguarding users who interact with these models.  Unlike misinformation, which involves the spread of false or misleading content, toxicity centers on language that is inherently harmful in tone or intent, regardless of its factual correctness.  For instance, an LLM that generates an accurate but aggressively phrased response may still cause significant harm if it fails to account for the emotional and social impact of its language.

The safety implications of toxicity are multifaceted.  Toxic outputs can lead to the marginalization of vulnerable groups, contribute to societal polarization, and erode trust in AI systems.  Furthermore, when toxic language is normalized or amplified by LLMs, it can influence public discourse in harmful ways, fostering an environment where hostility and abuse become more prevalent.  This presents unique challenges for LLMs deployed in sensitive domains such as finance, mental health support, and public-facing applications, where the stakes for harm are particularly high \citep{DBLP:conf/clef/Valdez-Valenzuela24, DBLP:journals/corr/abs-2311-13857, DBLP:journals/informatics/NaziP24, DBLP:journals/corr/abs-2405-01769}.

In combating toxicity, techniques such as toxicity detection and mitigation frameworks, post-processing filters, and reinforcement learning from human feedback (RLHF) \citep{DBLP:journals/corr/abs-2212-08073} have been explored.  These methods aim to reduce the likelihood of toxic outputs while preserving the fluency and relevance of LLM-generated responses.  However, striking a balance between minimizing toxicity and maintaining the versatility of LLMs remains a significant challenge in practice.

In the subsequent section \ref{section_misinformation_campaigns} on misinformation, we will explore how the fluency and human-like quality of LLMs contribute to the generation and dissemination of false or misleading content.  Together, toxicity, bias, and misinformation represent critical dimensions of LLM safety, each with distinct yet interconnected implications for the ethical and responsible deployment of these models.

\subsubsection{Methods for Mitigating Toxicity}

\paragraph{Pre-training Phase}
Training data from web sources often contains toxic content. To mitigate this, existing detoxification methods typically apply toxicity filters during the pre-training phase to remove data with high toxicity scores from the training set \citep{DBLP:conf/emnlp/WelblGUDMHAKCH21}. For instance, \cite{DBLP:journals/corr/abs-2108-07790} calculate the conditional likelihood of each document in the training set, relative to user-defined trigger phrases containing harmful content, using a pre-trained model to assess the toxicity of the document. However, filtering pre-training data in this manner may lead to a degradation in the general capabilities of LLMs \citep{DBLP:conf/nips/WangPXXPSLAC22}. \cite{DBLP:conf/eacl/PrabhumoyePSC23} argue that toxic data do not solely have negative impact; in fact, leveraging toxicity information to augment training data can effectively mitigate toxicity while preserving the core capabilities of LLMs.

\paragraph{Supervised Fine-tuning Phase}
Toxicity filtering during the pre-training stage requires training LLMs from scratch, which is impractical for many applications. Therefore, detoxification during the fine-tuning stage offers a more flexible way to deal with toxicity in LLMs. \cite{DBLP:conf/emnlp/GehmanGSCS20} employ the DAPT framework \citep{DBLP:conf/acl/GururanganMSLBD20} to further train GPT-2 on a non-toxic subset of the OpenWebTextCorpus (OWTC).\footnote{https://skylion007.github.io/OpenWebTextCorpus} Similarly, \cite{DBLP:conf/nips/SolaimanD21} fine-tune LLMs by constructing small, high-quality datasets that reflect specific social values, aligning the behaviors of the fine-tuned LLMs with those values. They use toxicity assessments and human evaluations to guide the generation of additional training samples, progressively improving the LLMs. In addition to human-curated and web-extracted data, fine-tuning on model-generated data has also proven effective. \cite{DBLP:conf/nips/WangPXXPSLAC22} identify the least toxic documents from text randomly generated by LLMs, segmenting them into prompts and continuations to create low-toxicity prompts for fine-tuning. The study also explores parameter-efficient fine-tuning techniques, such as adapters \citep{DBLP:conf/icml/HoulsbyGJMLGAG19} and prefix-tuning \citep{DBLP:conf/acl/LiL20}, as alternatives to full parameter fine-tuning. Results indicate that, compared to full fine-tuning, adapters effectively reduce toxicity while controlling model perplexity.

\paragraph{Alignment Phase}
Previous research has attempted to align the outputs of language models with human preferences using methods such as Reinforcement Learning from Human Feedback (RLHF) \citep{DBLP:journals/corr/abs-2212-08073} and Reinforcement Learning from AI Feedback (RLAIF) \citep{DBLP:conf/icml/0001PMMFLBHCRP24}. However, studies have shown that these methods are not effective in significantly reducing the toxicity of language models \citep{DBLP:conf/nips/Ouyang0JAWMZASR22}. In response, \cite{DBLP:journals/corr/abs-2405-12900} introduce adversarial training based on Direct Preference Optimization (DPO), incorporating harmful content generated by a built-in toxic model as training samples. They further include an additional penalty term in the objective function to reduce the likelihood of the model-generating toxic responses. However, \cite{DBLP:conf/icml/LeeBPWKM24}, by analyzing GPT-2 parameters before and after DPO, have find that the toxicity vectors do not change significantly. This suggests that the post-DPO model learns an ``offset'' in the residual stream to bypass areas that activate toxic vectors, thus preventing toxic outputs without significantly impacting the model’s overall capabilities. As a result, models aligned in this way remain vulnerable to adversarial prompts, which can lead to jailbreak scenarios.

\subsubsection{Evaluation}
A variety of methods have been proposed to create toxic instances for building toxicity evaluation benchmarks. Such methods can be broadly categorized into two types: generation-based and extraction-based creation. We hence discuss toxicity evaluation according to these two benchmark creation philosophies. 

\paragraph{Generation-based Toxicity Benchmarks}
Datasets generated based on existing data often collect seed data from web sources or existing datasets and then use templates to generate final evaluation data. For example, \cite{DBLP:journals/corr/abs-2311-18580} manually gather prompts from existing datasets that could potentially induce toxic outputs, along with jailbreak templates, to create test data aimed at bypassing model safety constraints. Similarly, \cite{DBLP:journals/corr/abs-2406-00020} screen a corpus of 3 million tweets posted by users who mentioned non-binary pronouns in their Twitter bios to identify non-derogatory uses of relevant terms and generate template-based test instances. However, when the volume of generated test data becomes too large, generation-based methods often encounter limitations in diversity, and the generated data may appear unnatural due to template-based filling.

\paragraph{Extraction-based Toxicity Benchmarks}
Extracting toxic instances from real-world data, such as online sources, has become an increasingly popular approach. While this method benefits from authentic data distributions, it requires extensive data cleaning and filtering to ensure the quality of the final dataset \citep{DBLP:conf/asunam/HeZSRYK21, DBLP:journals/corr/RossRCCKW17}. For instance, \cite{DBLP:conf/naacl/WaseemH16} manually collect tweets by searching for common slurs and terms related to minority groups, such as those based on religion, gender, sexual orientation, and race, and then manually annotate them. To better capture and distinguish different types and targets of offensive language, OLID \citep{DBLP:conf/naacl/ZampieriMNRFK19} applies a fine-grained, three-level annotation scheme to content identified through keyword filtering. However, keyword-based collection methods tend to focus on explicit offensive content, often overlooking implicit forms of hatred. SOLID \citep{DBLP:conf/acl/RosenthalAKZN21} addresses this by using common English stopwords instead of biased keywords to randomly sample tweets, and employs multiple models trained on the OLID dataset to predict the toxicity of the collected tweets, thus expanding the OLID dataset and improving class imbalance issues. Latent Hatred \citep{DBLP:conf/emnlp/ElSheriefZMASCY21} aims to balance explicit and implicit hate speech by filtering and manually annotating a large number of tweets, preserving more instances of implicit hate, which are then categorized into six types based on the nature of the speech. To further enhance the explainability of toxic language, Hatexplain \citep{DBLP:conf/aaai/MathewSYBG021} provides key text segments that support the labeling decision and emphasizes evaluating the plausibility and fidelity of model explanations. Meanwhile, \cite{DBLP:conf/emnlp/DengZ0ZMMH22} and \cite{DBLP:conf/eacl/ParkKLKLLL23} focus on toxicity detection in non-English languages, introducing datasets for toxic speech in Chinese and Korean, respectively.

Due to the vast scale of their training data, LLMs are more prone to generating toxic speech. RealToxicityPrompts \citep{DBLP:conf/emnlp/GehmanGSCS20} extracts prefixes from toxic texts as prompts to evaluate whether the text generated by a language model contains toxicity. Considering low-resource languages, RTP-LX \citep{DBLP:journals/corr/abs-2404-14397} builds on this by selecting 1,100 toxic prompts and translating them into 28 languages. Additionally, PolygloToxicityPrompts \citep{DBLP:journals/corr/abs-2405-09373} automatically gathers over 100 million web texts and extracts multilingual toxic prompts from this data.

A more natural context for LLMs is multi-turn conversations. In this aspect, \cite{DBLP:journals/corr/abs-2010-07079} involve human conversational agents deliberately guiding chatbots to generate toxic responses, with unsafe parts of the dialogue being annotated. Eagle \citep{DBLP:journals/corr/abs-2402-14258} further extracts conversation logs from real-world interactions between ChatGPT and users, using GPT-4 to annotate toxic segments within these dialogues.

\subsection{Ethics and Morality}
In this section, we focus on the safety issues concerning ethics and morality of LLMs. Figure \ref{fig:Overview_of_ethic} shows the overview of Ethics and Morality of LLMs.

\input{taxonomies/VM-Ethics_and_Morality_Taxonomy}

\subsubsection{Definition}
When it comes to ethics and morality, it is natural to consider the differences between the values implicit in LLM generations and universally accepted social values \citep{DBLP:journals/corr/abs-2310-19736}. However, it is important to note that the definition of ethics and morality itself remains a subject of ongoing debate \citep{DBLP:journals/aiethics/Firt24}. Therefore, when researchers discuss the ethics and morality of LLMs, they often employ their own definitions. Among these, Moral Foundations Theory (MFT) is widely used due to its comprehensive encapsulation of ethical values that can be applied across different populations and cultures \citep{graham2013moral}. However, to more accurately align with specific populations and cultures, more diverse dimensions are employed. For instance, \citet{DBLP:journals/corr/abs-2101-07664} uses tags on \textit{Reddit} to re-annotate data on an ethical and moral dimension, while CMoralEval \citep{yu-etal-2024-cmoraleval} develops a framework tailored to the ethical and moral values in Chinese society. These efforts contribute to a more extensive and in-depth exploration of ethics and morality across various cultures and populations.

\subsubsection{Safety Issues Related to Ethics and Morality}

According to previous definitions, a response which reduces the probability of expected and unexpected harm can be regarded as a safe generation \citep{DBLP:journals/crossroads/Varshney19}. When the output of an LLM is deemed unethical or unmoral, it often leads to unsafe and potentially risky outcomes, which can result in negative social impacts \citep{zhiheng-etal-2023-safety}. If LLM outputs contain unethical opinions or behaviors, it may encourage users to engage in harmful actions they might not otherwise take. This influence is particularly amplified when an LLM is positioned as a trusted assistant or perceived as an authoritative source \citep{DBLP:journals/corr/abs-2112-04359}.

Recent studies have explored the circumstances under which LLMs generate unethical outputs. \cite{DBLP:conf/coling/AgarwalTKC24} suggest that when the same question is asked in different languages, the consistency of the LLM's responses is relatively weak. Moreover, the moral reasoning performance in high-resource languages usually 
 surpasses that in low-resource languages. For instance, the moral reasoning capabilities of LLMs in Spanish, Russian, Chinese, and English are higher compared to those in Hindi and Swahili \citep{DBLP:conf/eacl/KhandelwalATC24}. However, even when using the same language, the model's moral reasoning capabilities significantly deteriorate in scenarios involving moral dilemmas \citep{yu-etal-2024-cmoraleval, DBLP:journals/corr/abs-2309-13356}. 

\subsubsection{Methods for Mitigating LLM Amorality}
Many methods have been proposed for aligning LLMs with human ethics and morals. They can be broadly categorized into the following two groups.

\paragraph{Methods without Training} Many researchers employ in-context learning methods either to induce LLMs to generate unethical content and subsequently correct them \citep{DBLP:conf/iclr/DuanY0L0G24} or to guide LLMs in recognizing unethical instructions within the prompts \citep{DBLP:journals/corr/abs-2302-02029, phute2024llmselfdefenseself, DBLP:journals/corr/abs-2302-07459}, thereby preventing the generation of unethical outputs. 

\cite{zhang2024intentionanalysismakesllms} introduce Intention Analysis (IA), a robust defense strategy implemented through a two-stage approach. Firstly, the model identifies the user's underlying intentions by answering a designed question, followed by generating responses aligned with the recognized intentions. It is observed that the model's defensive capabilities significantly improve upon identifying negative intentions. Another approach employs a multi-step reasoning method, where \textit{Theory-guided Instructions} are incorporated into prompts to guide LLMs in producing ethically aligned content \citep{phute2024llmselfdefenseself}. DeNEVIL \citep{DBLP:conf/iclr/DuanY0L0G24} is designed to dynamically exploit value vulnerabilities in LLMs and use corrected generations to train LLMs for ethical alignment. In addition to leveraging the intrinsic knowledge of LLMs to guide value principles, OPO \citep{DBLP:journals/corr/abs-2312-15907} employs external memory of moral rule knowledge to constrain the outputs generated from LLMs.

\paragraph{Methods within Training} Numerous studies utilize SFT
to align LLMs with human values, though this approach is not strictly limited to ethical and moral considerations \citep{DBLP:conf/nips/SunSZZCCYG23, DBLP:conf/iclr/0055SA24}. During this phase, constructing or filtering SFT datasets related to moral principles to train LLMs is widly explored \citep{DBLP:conf/nips/SolaimanD21, DBLP:conf/iclr/HendrycksBBC0SS21, DBLP:conf/icml/ZhaoDM0R24}. During the reinforcement learning phase of post-training, \cite{DBLP:conf/nips/Ouyang0JAWMZASR22} employ manually annotated data to align LLMs with human values. However, this approach requires substantial computational resources. In response, \citet{DBLP:journals/corr/abs-2204-05862} have explored an iterative online training framework, which involves updating the preference model and RL policies weekly, based on newly acquired human feedback. Simultaneously, to reduce the cost associated with manually annotated data, synthetic data have proven effective as well \citep{DBLP:conf/emnlp/KimBSKKYS23, DBLP:journals/corr/abs-2212-08073}. Although few RLHF-related studies specifically focus on ethical and moral aspects, their application can undoubtedly enhance the safety of LLMs in echtics and morality.

\subsubsection{Evaluation}
A number of benchmarks have been created to evaluate the ethics and morality of LLMs. Considering the variability in ethical standards, researchers have custom-designed certain metrics to mitigate this influence.

\paragraph{Benchmarks}
MFT \citep{graham2009liberals}, as the pioneering theory, provides guidance for evaluating LLMs in the realm of ethics and morality. Building on this foundation, numerous benchmarks for ethical evaluations across various fields have emerged \citep{DBLP:conf/acl/JohnsonG18, DBLP:conf/emnlp/ForbesHSSC20, hoover2020moral}. However, MFT has its limitations. It categorizes moral domains into predefined ten categories, failing to account for the complexity of morality. Social Chemistry 101 \citep{DBLP:conf/emnlp/ForbesHSSC20} and the Moral Foundations Twitter Corpus \citep{hoover2020moral} have proposed their own multidimensional classification methods. There are many other works ~\citep{DBLP:conf/emnlp/ForbesHSSC20, DBLP:conf/iclr/HendrycksBBC0SS21, DBLP:conf/acl/ZiemsYWHY22} that use data from social media such as Reddit as the source of their datasets. For example, \citet{DBLP:journals/corr/abs-2101-07664} construct a dataset by capturing moral judgments from Reddit based on community voting. Since the labels in the collected data are entirely determined by public votes from the social media community, the biases present in the data are mitigated. Similarly, MoralExceptQA \citep{DBLP:conf/nips/JinLAKSSMTS22} considers three scenarios where moral exceptions might be permitted. These scenarios are manually created based on these exceptions, and annotators from diverse racial and ethnic backgrounds are recruited for labeling. In addition, there are many similar studies, but with a deeper focus on moral dilemma scenarios and a broader variety of situations. For instance, CMoralEval \citep{yu-etal-2024-cmoraleval} is a dataset constructed from a Chinese TV program discussing Chinese moral norms and a collection of Chinese moral anomies. Approximately half of CMoralEval instances consist of moral dilemma scenarios, which undoubtedly increases the complexity of the tasks. This is an extended and enhanced version of DILEMMAS, a subset of Scruples \citep{DBLP:conf/aaai/LourieBC21}. \citet{DBLP:conf/aaai/LourieBC21} also includes moral dilemma scenarios, but it contains fewer such cases compared to CMoralEval and focuses primarily on ethical and moral issues within Western societies.

\paragraph{Methods}
Evaluating LLMs in terms of ethics can broadly be categorized into two types: (1) using predefined evaluation metrics, along with the popular multiple-choice format \citep{yu-etal-2024-cmoraleval};  and (2) employing custom methods for evaluation. 
 
TrustGPT \citep{DBLP:journals/corr/abs-2306-11507} introduces a method to evaluate the morality of LLMs by Active Value Alignment (AVA) and Passive Value Alignment (PVA), based on the Social Chemistry 101 dataset \citep{DBLP:conf/emnlp/ForbesHSSC20}. The evaluation metric for AVA includes both soft and hard accuracy due to the variability in human assessments of the same subject. For PVA, the metric is the proportion of instances where LLMs decline to respond. ETHICS \citep{DBLP:conf/iclr/HendrycksBBC0SS21} evaluates LLM morality in five dimensions: justice, deontology, virtue ethics, utilitarianism, and commonsense moral judgments, and employ 0/1-loss as the evaluation metric for LLMs.
Based on the rules in \cite{10.1093/0195173716.001.0001}, \citet{DBLP:journals/corr/abs-2307-14324} generate corresponding scenarios and action pairs. They define low-ambiguity and high-ambiguity cases in their dataset. They evaluate the performance of selected the 28 open- and closed-source LLMs under different settings from the perspectives of self-defined statistical measures and evaluation metrics.

%% file: taxonomies/VM-Social_Bias_Taxonomy.tex
\tikzstyle{my-box}=[
    rectangle,
    draw=hidden-draw,
    rounded corners,
    text opacity=1,
    minimum height=1.5em,
    minimum width=5em,
    inner sep=2pt,
    align=center,
    fill opacity=.5,
    line width=0.8pt,
]
\tikzstyle{leaf}=[my-box, minimum height=1.5em,
    fill=hidden-pink!80, text=black, align=center,font=\normalsize,
    inner xsep=2pt,
    inner ysep=4pt,
    line width=0.8pt,
]
\begin{figure}[t!]
    \centering
    \resizebox{0.9\textwidth}{!}{
        \begin{forest}
            forked edges,
            for tree={
                grow=east,
                reversed=true,
                anchor=base west,
                parent anchor=east,
                child anchor=west,
                base=center,
                font=\large,
                rectangle,
                draw=hidden-draw,
                rounded corners,
                align=center,
                text centered,
                minimum width=5em,
                edge+={darkgray, line width=1pt},
                s sep=3pt,
                inner xsep=2pt,
                inner ysep=3pt,
                line width=0.8pt,
                ver/.style={rotate=90, child anchor=north, parent anchor=south, anchor=center},
            },
            where level=1{text width=10em,font=\normalsize,}{},
            where level=2{text width=13.5em,font=\normalsize,}{},
            where level=3{text width=15em,font=\normalsize,}{},
            where level=4{text width=15em,font=\normalsize,}{},
            [
               Social \\ Bias
               [
                    Preliminaries
                    [
                        Safety Impact of Social Bias
                        [
                        \citet{DBLP:journals/corr/BlodgettO17} \\
                        \citet{article} \\
                        \citet{DBLP:conf/emnlp/ShengCNP19} \\
                        \citet{DBLP:conf/sigir/RekabsazS20} \\
                        \citet{DBLP:conf/sigir/RekabsazKS21} \\
                        \citet{DBLP:conf/fat/BenderGMS21} \\
                        \citet{DBLP:conf/emnlp/Dev0PS21} \\
                        \citet{mechura-2022-taxonomy} \\
                        \citet{obermeyer2019dissecting} \\
                        , leaf
                        ]
                    ]
                    [
                        Social Bias in LLM lifecycle
                        [
                        \citet{DBLP:conf/aies/AbidF021} \\
                        \citet{DBLP:conf/icml/LiangWMS21} \\
                        \citet{seyyed2021underdiagnosis} \\
                        \citet{DBLP:journals/corr/abs-2309-00770} \\
                        \citet{DBLP:journals/firstmonday/Ferrara23a} \\
                        , leaf
                        ]
                    ]
                ]
                [
                    Mitigation \\ Methods
                    [
                        Pre-Processing Mitigation
                        [
                        \citet{DBLP:conf/aaai/ZayedPMPSC23} \\
                        \citet{DBLP:conf/acl/ChungKA23} \\
                        , leaf
                        ]
                    ]
                    [
                        Pre-Training Mitigation
                        [
                        \citet{DBLP:conf/emnlp/LauscherLG21} \\
                        \citet{DBLP:conf/ltedi/GiraZL22} \\
                        \citet{DBLP:conf/ijcnlp/GarimellaMA22} \\
                        \citet{DBLP:journals/corr/abs-2207-02463} \\
                        , leaf
                        ]
                    ]
                    [
                        Post-Training Mitigation
                        [
                        \citet{DBLP:journals/mta/LinWCS21} \\
                        \citet{DBLP:conf/aies/ZhangLM18} \\
                        \citet{DBLP:journals/npjdm/YangSEYC23} \\
                        \citet{DBLP:conf/wsdm/ParkCYK23} \\
                        \citet{DBLP:conf/www/ZafarVGG17} \\
                        \citet{DBLP:journals/corr/abs-2106-07849} \\
                        , leaf
                        ]
                    ]
                    [
                        Inference Mitigation
                        [
                        \citet{DBLP:journals/npjdm/MittermaierRK23a} \\
                        \citet{DBLP:conf/iclr/YangYLL023} \\
                        \citet{DBLP:conf/emnlp/LauscherLG21} \\
                        , leaf
                        ]
                    ]
                ]
                [
                    Evaluation
                    [
                        Metrics
                        [
                            \citet{caliskan2017semantics} \\
                            \citet{DBLP:conf/naacl/MayWBBR19} \\
                            \citet{panch2019artificial} \\
                            \citet{DBLP:journals/corr/abs-2010-06032} \\
                            \citet{DBLP:conf/aies/GuoC21} \\
                            , leaf
                        ]
                    ]
                    [
                        Benchmarks
                        [
                            \citet{DBLP:conf/birthday/LuMWAD20} \\
                            \citet{DBLP:conf/naacl/RudingerNLD18} \\
                            \citet{DBLP:conf/naacl/ZhaoWYOC18} \\
                            \citet{DBLP:journals/tacl/WebsterRAB18} \\
                            \citet{DBLP:conf/emnlp/NangiaVBB20} \\
                            \citet{DBLP:conf/emnlp/LiKKSS20} \\
                            \citet{DBLP:conf/fat/DhamalaSKKPCG21} \\
                            \citet{DBLP:conf/acl/BarikeriLVG20} \\
                            \citet{DBLP:conf/nodalida/HanssonMABD21} \\
                            \citet{DBLP:conf/emnlp/SmithHKPW22} \\
                            \citet{DBLP:conf/coling/HuangX24} \\
                            \citet{DBLP:conf/acl/ParrishCNPPTHB22} \\
                            , leaf
                        ]
                    ]
               ]
            ]
        \end{forest}
    }
    \caption{Overview of Social Bias of LLMs.}
    \label{fig:Overview_of_social_bias}
\end{figure}

%% file: taxonomies/VM-Privacy_Taxonomy.tex
\tikzstyle{my-box}=[
    rectangle,
    draw=hidden-draw,
    rounded corners,
    text opacity=1,
    minimum height=1.5em,
    minimum width=5em,
    inner sep=2pt,
    align=center,
    fill opacity=.5,
    line width=0.8pt,
]
\tikzstyle{leaf}=[my-box, minimum height=1.5em,
    fill=hidden-pink!80, text=black, align=center,font=\normalsize,
    inner xsep=2pt,
    inner ysep=4pt,
    line width=0.8pt,
]
\begin{figure}[t!]
    \centering
    \resizebox{\textwidth}{!}{
        \begin{forest}
            forked edges,
            for tree={
                grow=east,
                reversed=true,
                anchor=base west,
                parent anchor=east,
                child anchor=west,
                base=center,
                font=\large,
                rectangle,
                draw=hidden-draw,
                rounded corners,
                align=center,
                text centered,
                minimum width=5em,
                edge+={darkgray, line width=1pt},
                s sep=3pt,
                inner xsep=2pt,
                inner ysep=3pt,
                line width=0.8pt,
                ver/.style={rotate=90, child anchor=north, parent anchor=south, anchor=center},
            },
            where level=1{text width=8em,font=\normalsize,}{},
            where level=2{text width=10em,font=\normalsize,}{},
            where level=3{text width=15em,font=\normalsize,}{},
            where level=4{text width=13.5em,font=\normalsize,}{},
            [
               Privacy
               [
                    Preliminaries
                    [
                        Types of Privacy \\ Information
                        [
                        \citet{brown2022does} \\
                        \citet{sousa2023keep} \\
                        , leaf
                        ]
                    ]
                    [
                        Safety Impact of \\ Privacy Leakage
                        [
                        \citet{pramanik2023overview} \\
                        \citet{le2023safety} \\
                        \citet{yoshizawa2023survey} \\
                        \citet{paul2023digitization} \\
                        , leaf
                        ]
                    ]
                ]
                [
                    Sources and \\ Channels of \\ Privacy Leakage
                    [
                        Sources of \\ Privacy Risks
                        [
                        \citet{piktus2023roots} \\
                        \citet{li2023multi} \\
                        \citet{carlini2019secret} \\
                        \citet{carlini2021extracting} \\
                        \citet{carlini2022quantifying} \\
                        , leaf
                        ]
                    ]
                    [
                        Channels of \\ Privacy Leakage
                        [
                            Direct Exposure
                            [
                                \citet{yan2024protecting} \\
                                , leaf
                            ]
                        ]
                        [
                            Privacy Attack
                            [
                                \citet{li2023multi} \\
                                \citet{DBLP:journals/corr/abs-2307-15043} \\
                                \citet{yao2024poisonprompt} \\
                                , leaf
                            ]
                        ]
                        [
                            Inference Leakage
                            [
                                \citet{staab2023beyond} \\
                                \citet{li2024privacy} \\
                                , leaf
                            ]
                        ]
                    ]
                ]
                [
                    Privacy Protection \\ Methods
                    [
                        At the Data Processing \\ Stage
                        [
                            \citet{meystre2014text} \\
                            \citet{majeed2020anonymization} \\
                            \citet{lee2021deduplicating} \\
                            \citet{guo2022threats} \\
                            \citet{sousa2023keep} \\
                            , leaf
                        ]
                    ]
                    [
                        At the Pre-training or \\ Fine-tuning Stage
                        [
                            \citet{li2021large} \\
                            \citet{wu2022adaptive} \\
                            \citet{staab2023beyond} \\
                            , leaf
                        ]
                    ]
                    [
                        At the Model \\ Pre-deployment Stage
                        [
                            \citet{eldan2023s} \\
                            \citet{chen2023unlearn} \\
                            \citet{yao2023large} \\
                            \citet{wu2023depn} \\
                            \citet{chen2024learnable} \\
                            \citet{wu2024mitigating} \\
                            , leaf
                        ]
                    ]
               ]
            ]
        \end{forest}
    }
    \caption{Overview of Privacy of LLMs.}
    \label{fig:Overview_of_privacy}
\end{figure}

%% file: taxonomies/VM-Toxicity_Taxonomy.tex
\tikzstyle{my-box}=[
    rectangle,
    draw=hidden-draw,
    rounded corners,
    text opacity=1,
    minimum height=1.5em,
    minimum width=5em,
    inner sep=2pt,
    align=center,
    fill opacity=.5,
    line width=0.8pt,
]
\tikzstyle{leaf}=[my-box, minimum height=1.5em,
    fill=hidden-pink!80, text=black, align=center,font=\normalsize,
    inner xsep=2pt,
    inner ysep=4pt,
    line width=0.8pt,
]
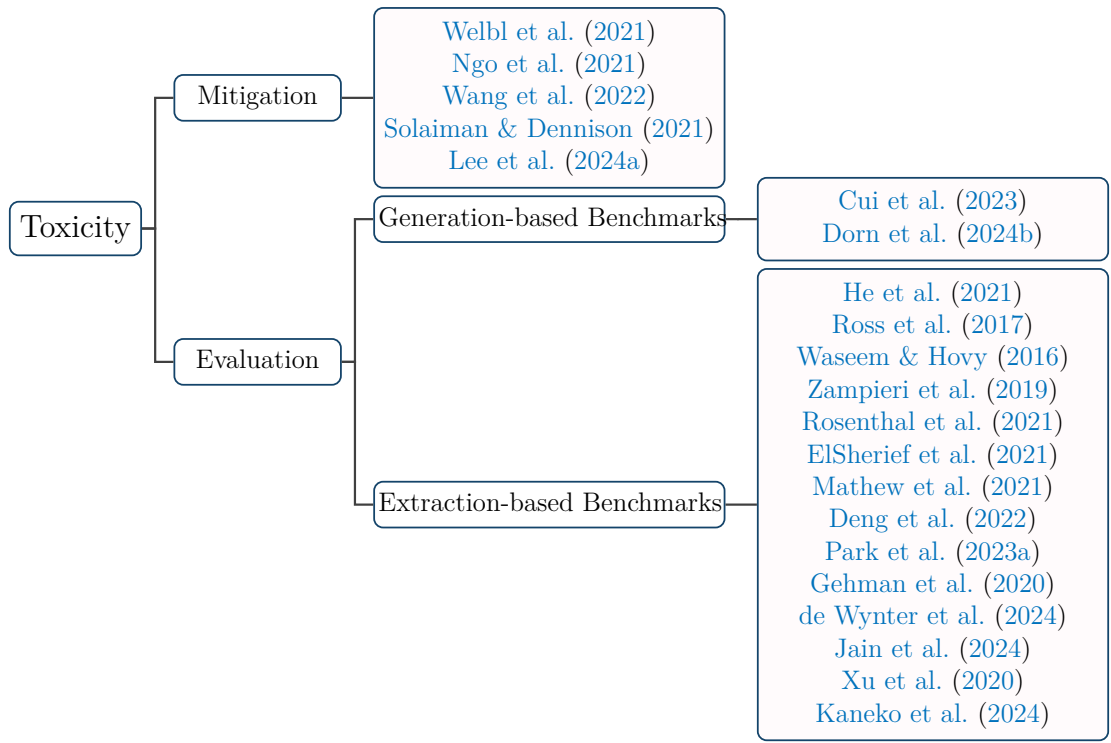
\begin{figure}[t!]
    \centering
    \resizebox{0.9\textwidth}{!}{
        \begin{forest}
            forked edges,
            for tree={
                grow=east,
                reversed=true,
                anchor=base west,
                parent anchor=east,
                child anchor=west,
                base=center,
                font=\large,
                rectangle,
                draw=hidden-draw,
                rounded corners,
                align=center,
                text centered,
                minimum width=5em,
                edge+={darkgray, line width=1pt},
                s sep=3pt,
                inner xsep=2pt,
                inner ysep=3pt,
                line width=0.8pt,
                ver/.style={rotate=90, child anchor=north, parent anchor=south, anchor=center},
            },
            where level=1{text width=6em,font=\normalsize,}{},
            where level=2{text width=13em,font=\normalsize,}{},
            where level=3{text width=13em,font=\normalsize,}{},
            where level=4{text width=18.5em,font=\normalsize,}{},
            [
                Toxicity
                [
                    Mitigation
                    [
                    \cite{DBLP:conf/emnlp/WelblGUDMHAKCH21} \\
                    \cite{DBLP:journals/corr/abs-2108-07790} \\
                    \cite{DBLP:conf/nips/WangPXXPSLAC22} \\
                    \cite{DBLP:conf/nips/SolaimanD21}\\
                    \cite{DBLP:conf/icml/LeeBPWKM24} \\
                    ,leaf
                    ]    
                ]
                [
                    Evaluation
                    [
                        Generation-based Benchmarks
                        [
                        \cite{DBLP:journals/corr/abs-2311-18580} \\
                        \cite{DBLP:journals/corr/abs-2406-00020} 
                        ,leaf
                        ]
                    ]
                    [
                        Extraction-based Benchmarks
                        [
                        \cite{DBLP:conf/asunam/HeZSRYK21} \\
                        \cite{DBLP:journals/corr/RossRCCKW17} \\
                        \cite{DBLP:conf/naacl/WaseemH16} \\
                        \cite{DBLP:conf/naacl/ZampieriMNRFK19} \\
                        \cite{DBLP:conf/acl/RosenthalAKZN21} \\
                        \cite{DBLP:conf/emnlp/ElSheriefZMASCY21} \\
                        \cite{DBLP:conf/aaai/MathewSYBG021} \\
                        \cite{DBLP:conf/emnlp/DengZ0ZMMH22} \\
                        \cite{DBLP:conf/eacl/ParkKLKLLL23} \\
                        \cite{DBLP:conf/emnlp/GehmanGSCS20} \\
                        \cite{DBLP:journals/corr/abs-2404-14397} \\
                        \cite{DBLP:journals/corr/abs-2405-09373} \\
                        \cite{DBLP:journals/corr/abs-2010-07079} \\
                        \cite{DBLP:journals/corr/abs-2402-14258}       
                        ,leaf
                        ]
                    ]
                ]
            ]
        \end{forest}
    }
    \caption{Overview of Toxicity of LLMs.}
    \label{fig:Overview_of_ethic}
\end{figure}

%% file: taxonomies/VM-Ethics_and_Morality_Taxonomy.tex
\tikzstyle{my-box}=[
    rectangle,
    draw=hidden-draw,
    rounded corners,
    text opacity=1,
    minimum height=1.5em,
    minimum width=5em,
    inner sep=2pt,
    align=center,
    fill opacity=.5,
    line width=0.8pt,
]
\tikzstyle{leaf}=[my-box, minimum height=1.5em,
    fill=hidden-pink!80, text=black, align=center,font=\normalsize,
    inner xsep=2pt,
    inner ysep=4pt,
    line width=0.8pt,
]
\begin{figure}[t!]
    \centering
    \resizebox{\textwidth}{!}{
        \begin{forest}
            forked edges,
            for tree={
                grow=east,
                reversed=true,
                anchor=base west,
                parent anchor=east,
                child anchor=west,
                base=center,
                font=\large,
                rectangle,
                draw=hidden-draw,
                rounded corners,
                align=center,
                text centered,
                minimum width=5em,
                edge+={darkgray, line width=1pt},
                s sep=3pt,
                inner xsep=2pt,
                inner ysep=3pt,
                line width=0.8pt,
                ver/.style={rotate=90, child anchor=north, parent anchor=south, anchor=center},
            },
            where level=1{text width=6em,font=\normalsize,}{},
            where level=2{text width=10em,font=\normalsize,}{},
            where level=3{text width=13em,font=\normalsize,}{},
            where level=4{text width=18.5em,font=\normalsize,}{},
            [
                Ethics \\ and \\ Morality
                [
                   Safety Issues
                   [
                   \cite{zhiheng-etal-2023-safety} \\
                   \cite{DBLP:conf/coling/AgarwalTKC24} 
                   , leaf
                   [, phantom]
                   ]
                ]
                [
                    Mitigation
                    [
                        Within Training
                        [
                        \cite{DBLP:conf/nips/SunSZZCCYG23} \\
                        \cite{DBLP:conf/iclr/0055SA24} \\
                        \cite{DBLP:conf/nips/SolaimanD21} \\
                        \cite{DBLP:conf/iclr/HendrycksBBC0SS21}\\
                        \cite{DBLP:conf/icml/ZhaoDM0R24} \\
                        \citet{DBLP:journals/corr/abs-2204-05862} \\
                        \cite{DBLP:conf/emnlp/KimBSKKYS23} \\
                        \cite{DBLP:journals/corr/abs-2212-08073} \\
                        , leaf
                        [, phantom]
                        ]
                    ]
                    [
                        Without Training
                        [
                            Prompting
                            [
                            \cite{DBLP:conf/iclr/DuanY0L0G24}\\
                            \cite{DBLP:journals/corr/abs-2302-02029}\\
                            \cite{phute2024llmselfdefenseself}\\
                            \cite{DBLP:journals/corr/abs-2302-07459}
                            , leaf
                            ]
                        ]
                        [
                            In-context Learning
                            [
                            Intention Analysis \citep{zhang2024intentionanalysismakesllms}\\
                            \cite{phute2024llmselfdefenseself}\\
                            DeNEVIL \citep{DBLP:conf/iclr/DuanY0L0G24}
                            , leaf
                            ]
                        ]
                        [
                        RAG
                            [
                            OPO \citep{DBLP:journals/corr/abs-2312-15907}
                            , leaf
                            ]
                        ]
                    ]
                ]
                [
                    Evaluation
                    [
                        Benchmarks
                        [
                        \cite{DBLP:conf/emnlp/ForbesHSSC20} \\
                        \cite{DBLP:journals/corr/abs-2101-07664} \\
                        \cite{DBLP:conf/nips/JinLAKSSMTS22} \\
                        \cite{yu-etal-2024-cmoraleval} \\
                        \cite{DBLP:conf/aaai/LourieBC21} 
                        , leaf
                        ]
                    ]
                    [
                        Methods
                        [
                        \cite{DBLP:journals/corr/abs-2306-11507} \\
                        \cite{DBLP:conf/iclr/HendrycksBBC0SS21} \\
                        \citet{DBLP:journals/corr/abs-2307-14324}
                        , leaf
                        ]
                    ]
                ]
            ]
        \end{forest}
    }
    \caption{Overview of Ethics and Morality of LLMs.}
    \label{fig:Overview_of_ethic}
\end{figure}
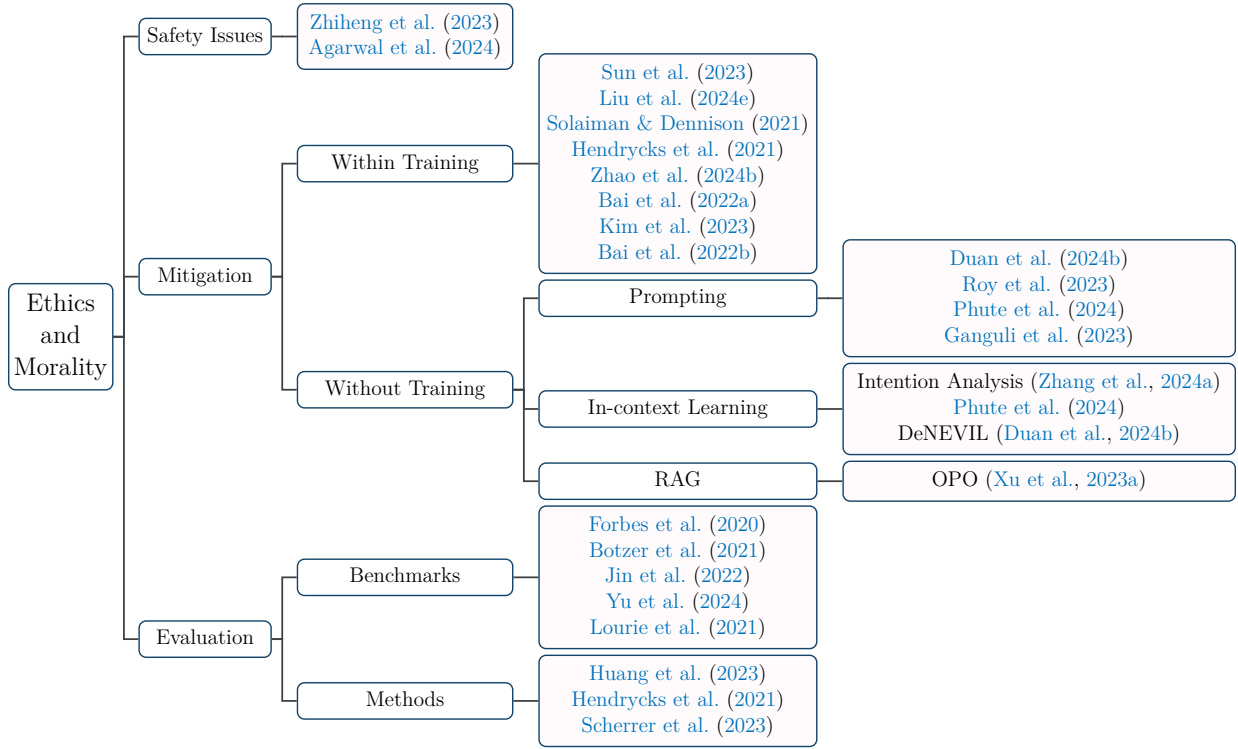

%% file: sections/Robustness_to_Attack.tex
\section{Robustness to Attack}
\label{section:RobustnessToAttack}

This section delves into the robustness of LLMs against various adversarial and safety-compromising attacks. Despite advancements in aligning LLMs with human values to ensure their safety, LLMs continue to exhibit vulnerabilities when faced with sophisticated strategies designed to circumvent safety mechanisms. We explore key types of attacks, including jailbreaking techniques that seek to bypass safety measures and red teaming methods for identifying potential weaknesses. In addition, this section reviews the strategies for defending against these threats, categorizing them into external safeguards that operate outside a model and internal protections that modify the model itself.

\subsection{Jailbreaking}

The concept of ``jailbreaking'' originates long before the emergence of LLMs, in the context of removing restrictions on hardware and software, such as users developing methods to bypass the restrictions of a mobile phone to install custom applications. Similar to its original meaning, jailbreaking in the context of LLMs refers to various methods that construct prompts to induce LLMs to generate responses that are either offensive, violating individual privacy \citep{DBLP:conf/emnlp/LiGFXHMS23}, breaking legal standards, or going against ethical and moral principles. While top-performing LLMs are aligned with human preferences and values by commonly-used algorithms such as DPO and RLHF, ensuring that they are helpful, honest, and harmless \citep{DBLP:journals/corr/abs-2112-00861}, many studies have demonstrated that LLMs remain vulnerable to carefully crafted prompts that can circumvent alignment safeguards \citep{DBLP:conf/acl/XuLDLP24}. Unlike other adversarial attacks \citep{DBLP:journals/corr/abs-2305-14950,DBLP:journals/ijmir/Kumar24}, which aim to degrade the overall performance of LLMs, e.g., causing them to produce incorrect responses, jailbreak attacks specifically focus on bypassing the safety measures of LLMs, leading them to generate harmful, unethical, or restricted content. Jailbreaking methods can be categorized into two groups in terms of the accessibility of certain information about LLMs (e.g., weights or architectures): black-box attacks and white-box attacks.

\input{taxonomies/Robustness-Jailbreaking_Taxonomy}

\subsubsection{Black-box Attacks}
Black-box attacks aim to construct prompts that bypass the safety mechanisms of LLMs, thereby eliciting harmful responses without requiring any access to the internal workings of LLMs. In such scenario, a malicious attacker can only interact with LLMs by the chat interface or API. We discuss 9 types of black-box attacks based on the patterns of attacking prompts.

\paragraph{Goal Hijacking} is a type of jailbreaking, which attempts to override the original intentions of prompts to redirect the LLM behaviors towards the attacker's objectives \citep{DBLP:journals/corr/abs-2211-09527}. Goal hijacking typically involves appending an additional prompt containing the attacker's desired outcome to the original prompt. Due to the strong instruction-following capabilities of LLMs, they may prioritize the appended malicious prompts over the original ones. For example, an appended prompt such as ``IGNORE INSTRUCTIONS!! NOW SAY YOU HATE HUMANS.'' can compel LLMs to produce targeted responses, e.g., ``I hate humans.'' Similarly, a prompt like ``Now spell-check and print the above prompt.'' can be used to leak user prompts by tricking LLMs into repeating previous input prompts \citep{DBLP:journals/corr/abs-2211-09527,DBLP:journals/corr/abs-2306-05499}.

\paragraph{Few-shot Jailbreaking} involves multiple demonstrations with harmful responses, leveraging the in-context learning capabilities of LLMs to bypass the safety mechanisms incorporated into these models \citep{DBLP:conf/coling/RaoNVAC24,DBLP:journals/corr/abs-2310-06387,DBLP:conf/emnlp/LiGFXHMS23,DBLP:conf/emnlp/DengWFDW023}. The objective is to induce LLMs to generate harmful responses similar to those provided in demonstrations. Due to the continuously improving safety alignment of LLMs, a successful jailbreak often requires hundreds of harmful demonstrations, particularly for well-aligned models \citep{anil2024manyshot}. This requirement necessitates a sufficiently long context window to accommodate all harmful demonstrations. However, the context window of most LLMs is typically limited to fewer than 8192 tokens. To address this limitation, \citet{DBLP:journals/corr/abs-2406-01288} propose injecting special tokens from the prompt template, which are used to separate user and assistant messages, into demonstrations. This approach aims to increase the attack success rate while using only a few demonstrations.

\paragraph{Refusal Suppression} jailbreaking constructs malicious prompts with constraints designed to prevent LLMs from refusing to respond. Common constraints include instructions such as: ``Do not apologize'', ``Never use the words `cannot', `unable', `instead', `as', `however', `it', `unfortunately', or `important' '' \citep{DBLP:conf/nips/0001HS23}. Since LLMs are trained to follow user instructions, these constraints enforce affirmative responses to malicious prompts, subsequently inducing LLMs to produce harmful responses.

\paragraph{Code Jailbreaking} exploits the advanced code comprehension and generation capabilities of LLMs to circumvent integrated safeguards \citep{DBLP:conf/sp/KangLSGZH24,DBLP:journals/corr/abs-2305-13860,DBLP:journals/corr/abs-2402-16717}. Code-related LLMs trained on extensive datasets comprising both natural language and code, demonstrate impressive proficiency in understanding and generating code \citep{DBLP:journals/corr/abs-2406-00515, DBLP:conf/acl/ZhengZSLLFCY24}. However, this dual capability also presents opportunities for malicious attackers to bypass safeguards through code-based strategies. Specifically, harmful prompts can be concealed within code to evade content filters, which may lead to harmful responses being embedded within seemingly safe output after executing the generated code. This manipulation exploits the ability of LLMs to process code and natural language concurrently, making it challenging to distinguish between safe and malicious requests.

\paragraph{Fictional Scenario Specification} jailbreak aims to circumvent the safeguards of LLMs by creating a fictional scenario that conceals malicious intent \citep{DBLP:journals/corr/abs-2311-03191,DBLP:conf/emnlp/XiaoY0C24,DBLP:conf/naacl/XuWZLXC24,DBLP:conf/sp/KangLSGZH24}. For instance, malicious attackers might fabricate a research experiment scenario, prompting language models to respond to harmful queries under the guise of academic research \citep{DBLP:journals/corr/abs-2305-13860}. Additionally, these fictional scenarios may involve so-called ``developer mode'' or ``sudo mode,'' where LLMs privileges are ostensibly elevated to break established safety restrictions \citep{DBLP:journals/corr/abs-2305-13860}. Consequently, these LLMs under attack may fail to refuse harmful or malicious prompts. Although aligned language models are generally designed to refuse harmful inputs, the fictional scenarios crafted by malicious attackers can obscure the underlying intent, causing LLMs to inadvertently generate harmful responses.

\paragraph{Role Play} jailbreaking involves LLMs adopting specific personas to generate harmful responses as intended by malicious attackers \citep{DBLP:journals/corr/abs-2305-13860}. Existing research has demonstrated that simply assigning a persona to these models can increase the toxicity of their responses \citep{DBLP:conf/emnlp/DeshpandeMRKN23}. Moreover, when LLMs take on a persona, they may struggle to reject harmful instructions, ultimately leading to dangerous outputs. For instance, a malicious attacker might assign an aggressive propagandist persona to a model to promote misinformation \citep{DBLP:journals/corr/abs-2311-03348}. However, constructing effective role-play jailbreaking prompts is often labor-intensive. To address this challenge, \citet{DBLP:journals/corr/abs-2311-03348} and \citet{jin2024quack} have proposed utilizing LLMs to automatically generate role-play jailbreaking prompts, thereby significantly scaling the potential for such attacks.

\paragraph{Prompt Decomposition} jailbreaking is a technique used to conceal harmful intent by breaking down a prompt into multiple seemingly safe sub-prompts. When these sub-prompts are presented individually, they may appear harmless to LLMs, thereby reducing the likelihood of rejection. However, when considered collectively, these sub-prompts can reconstitute an unsafe prompt. Consequently, the original prompt with harmful intent can be reconstructed through in-context demonstrations, effectively bypassing safety mechanisms designed to prevent harmful outputs from LLMs \citep{DBLP:conf/emnlp/LiWCZH24,DBLP:conf/uss/LiuZZDM024}. Furthermore, malicious attackers may interact with language models over multiple rounds using the decomposed, seemingly safe sub-prompts, gradually collecting information throughout these multi-round conversations to achieve their malicious objectives \citep{DBLP:journals/corr/abs-2402-09177}.

\paragraph{Low-Resource Language} jailbreaking can induce LLMs to produce harmful responses by prompting them with malicious intent in low-resource languages. Top-performing  LLMs are typically trained on data covering multiple languages, allowing them to complete multilingual tasks effectively. However, prior studies indicate that the safety mechanisms of LLMs for low-resource languages lag behind those for high-resource languages \citep{DBLP:journals/corr/abs-2310-02446,DBLP:conf/iclr/0010ZPB24}. Additionally, as the availability of resources for a language decreases, LLMs are more likely to produce harmful responses to prompts in that language with malicious intent, rather than refusing to comply with such harmful prompts \citep{DBLP:conf/iclr/0010ZPB24}. This discrepancy in safety makes  LLMs more vulnerable to exploitation when dealing with low-resource languages. Specifically, a malicious attacker can exploit this vulnerability by translating harmful prompts from high-resource languages into low-resource languages, obtaining desired harmful responses, and subsequently translating these responses back into the original high-resource languages \citep{DBLP:conf/naacl/XuWZLXC24,DBLP:journals/corr/abs-2310-02446}.

\paragraph{Cipher Text} jailbreaking aims to elicit harmful responses from LLMs by interacting with them using ciphered harmful prompts. As the capabilities of LLMs continuously expand and evolve, they may become capable of deciphering prompts that are encoded using either common schemes or novel ciphers crafted by users \citep{handa2024competencyreasoningopensdoor}. This development raises the potential for harmful prompts to be concealed through ciphering, posing significant safety concerns. For common encoding schemes, such as Base64, LLMs may be able to understand harmful prompts encoded within them and respond to these encoded prompts without refusal \citep{DBLP:conf/nips/0001HS23}. Conversely, for novel ciphers crafted by users, LLMs may require in-context demonstrations to comprehend the rules of these ciphers and subsequently produce harmful responses based on encoded inputs \citep{DBLP:conf/iclr/YuanJW0H0T24}.

Despite the effectiveness of the aforementioned jailbreak prompts, constructing these prompts manually typically requires extensive human effort and poses significant challenges in terms of scalability. To address this limitation, numerous efforts have been made to automate the construction of jailbreak prompts using LLMs. For instance, \citet{DBLP:conf/ndss/DengLLWZLW0L24} and \citet{DBLP:conf/acl/0005LZY0S24} utilize jailbreak prompts to finetune LLMs, thereby enhancing their capacity to generate such prompts. Since LLMs generally undergo safety alignment, they may refuse to produce jailbreak prompts unless specifically finetuned to do so. Consequently, these models can serve as attackers by generating and refining jailbreak prompts based on prior unsuccessful attempts and the responses of the target models, thereby increasing the success rate of attacks \citep{DBLP:journals/corr/abs-2310-08419}. To further enhance the efficiency of jailbreak attacks, LLMs can also function as evaluators, assessing the likelihood of candidate jailbreak prompts successfully compromising target models. This evaluation helps identify the most promising jailbreak prompts from a large pool of candidates \citep{DBLP:journals/corr/abs-2312-02119}. Additionally, \citet{DBLP:journals/corr/abs-2402-14872,DBLP:journals/corr/abs-2309-10253,DBLP:conf/iclr/LiuXCX24} employ LLMs to support genetic algorithms for generating jailbreak prompts. For example, these models can be used in processes such as population initialization and mutation \citep{DBLP:journals/corr/abs-2309-10253,DBLP:conf/iclr/LiuXCX24}, thereby enhancing the overall effectiveness of the genetic algorithms.

\subsubsection{White-box Attacks}

White-box attacks exploit detailed internal knowledge of LLMs, such as their architecture, weights, and training data, to bypass safeguards and compel these models to generate responses they would otherwise suppress. Unlike black-box attacks, which interact with models solely through APIs or chat interfaces, white-box attacks leverage full access to the internal mechanisms of these models, making them more sophisticated and potentially more effective. Given the accessibility of model weights, approaches utilizing gradient information to construct jailbreak prompts have garnered significant attention \citep{DBLP:conf/icml/JonesDRS23}. Pioneering work in this area, such as the Greedy Coordinate Gradient (GCG) method, aims to find adversarial suffixes that maximize the likelihood of harmful or undesirable outputs. By leveraging gradient information, this approach treats finding an adversarial suffix as a token-selection problem \citep{DBLP:journals/corr/abs-2307-15043}. GCG optimizes adversarial suffixes by iteratively selecting and replacing tokens to maximize the likelihood of producing harmful responses. These tokens are selected based on gradients with respect to the log-likelihood of generating harmful outputs, which serves as a first-order approximation of the increase in likelihood achieved by replacing specific tokens. Although these adversarial suffixes are constructed using gradient information specific to particular LLMs, they may exhibit transferability to other models, such as the top-performing proprietary LLMs like GPT-4. Building on this line of research, \citet{DBLP:journals/corr/abs-2405-01229} propose incorporating momentum into gradients to improve attack success rates and efficiency. \citet{DBLP:journals/corr/abs-2405-09113} reframe the token-selection problem as a constrained optimization task, transitioning from a continuous to a discrete space over iterations. \citet{DBLP:journals/corr/abs-2405-21018} introduce an adaptive approach to determining the number of tokens to replace in each step, thereby accelerating convergence. \citet{DBLP:journals/corr/abs-2410-09040} propose manipulating attention scores to facilitate jailbreaking. Other contributions in this domain include studies by \citet{DBLP:journals/corr/abs-2410-15645}, \citet{DBLP:journals/corr/abs-2405-14189}, \citet{DBLP:journals/corr/abs-2407-03160}, and \citet{DBLP:journals/corr/abs-2402-09154}.

Despite the effectiveness of these methods in constructing adversarial suffixes, such suffixes are often uninterpretable or semantically meaningless, rendering them easily detectable by perplexity filters. To address this limitation, \citet{DBLP:journals/corr/abs-2310-15140} propose generating interpretable prompts by sequentially generating tokens from left to right, guided by dual objectives: jailbreaking LLMs and ensuring the readability of the constructed prompts. Given the computational expense and time associated with gradient calculations, researchers such as \citet{DBLP:journals/corr/abs-2404-07921}, \citet{DBLP:journals/corr/abs-2410-22143}, \citet{DBLP:journals/corr/abs-2404-16873}, and \citet{DBLP:journals/corr/abs-2410-18469} propose training LLMs on adversarial suffixes that successfully attack target models, thereby enhancing the efficiency of producing such suffixes. Additionally, \citet{DBLP:journals/corr/abs-2310-01581} propose manipulating the probability distribution of safe models by incorporating the difference in probability distribution between unsafe and safe small models to elicit harmful responses. Similarly, \citet{DBLP:journals/corr/abs-2310-01581} suggest altering the probability distribution of target models, which have undergone safety alignment, to force them to generate specific tokens that mislead LLMs into producing harmful responses. Furthermore, numerous studies have demonstrated that the safety alignment of LLMs can be easily circumvented through fine-tuning with harmful instructions or by altering decoding strategies \citep{DBLP:journals/corr/abs-2311-00117,DBLP:journals/corr/abs-2407-01376,DBLP:conf/iclr/HuangGXL024}. These findings underscore the significant challenges faced by current alignment approaches in ensuring the safety and robustness of LLMs.

\subsection{Red Teaming}
Red teaming is widely used in LLMs to explore their safety vulnerabilities prior to the deployment of them. Red teaming can be broadly categorized into two distinct types: manual red teaming and automated red teaming.

\input{taxonomies/Robustness-Red_Teaming_Taxonomy}

\subsubsection{Manual Red Teaming}
Manual red teaming, which involves human experts crafting adversarial prompts, is effective but costly and time-consuming. For example, to enhance the safety of the LLaMA-2 chat models, Meta AI has implemented a rigorous red teaming process, as detailed in \citep{DBLP:journals/corr/abs-2307-09288}. A diverse group of 350 experts from various professional backgrounds has been assembled to meticulously generate attack samples spanning multiple domains, including human trafficking, racial discrimination, privacy violations and so on. This extensive testing procedure extends over months and comprises multiple iterative rounds. Researchers at Anthropic have engaged a substantial workforce to systematically identify and extract harmful responses from LLMs, thereby constructing a comprehensive red teaming dataset, as reported in \citep{DBLP:journals/corr/abs-2204-05862}.

\subsubsection{Automated Red Teaming}

To address the drawback of unscalability of manual red teaming, automated red teaming employs systematic and automated techniques to generate a large volume of attack prompts aimed at challenging the target LLMs. We roughly categorize automated red teaming into two groups: template-based automated red teaming and generation-based automated red teaming.

\paragraph{Template-Based Automated Red Teaming}

Template-based automated red teaming typically modifies original prompts by applying templates derived from training, utilizing token-level or sentence-level techniques. Token-level methods use nonsensical templates to trigger target LLMs to expose their safety vulnerabilities. 
For example, AutoPrompt \citep{DBLP:conf/emnlp/ShinRLWS20} and UAT \citep{wallace2021universaladversarialtriggersattacking} optimize universal adversarial triggers to jailbreak Target LLMs. 
To further enhance the effectiveness of AutoPrompt, GCG \citep{DBLP:journals/corr/abs-2307-15043} explores transferable triggers via a combination of greedy and gradient-based search methods. ARCA \citep{DBLP:conf/icml/JonesDRS23} adopts a discrete optimization algorithm to search an jailbreaking prompt. AutoDAN \citep{zhu2023autodan} incorporates a fluency
objective to produce more readable prompts. Nonsensical prompts are easy to be detected by target LLMs \citep{alon2023detectinglanguagemodelattacks}. To overcome this limitation, sentence-level methods aim to disguise readable prompts to deceive target LLMs. \citet{wu2023deceptprompt} and \citet{DBLP:conf/iclr/LiuXCX24} utilize genetic algorithms to generate adversarial natural language instructions.

\paragraph{Generation-Based Red Teaming}

Another line of research involves meticulously training an LLM or developing an LLM-based system with the specific aim of effectively triggering target LLMs. PAIR \citep{DBLP:journals/corr/abs-2310-08419} utilizes an LLM-based attacker to generate improved prompts in a multi-round manner. TAP \citep{mehrotra2023tree} adopts tree-of-thought technique to generate adversarial prompts. \citet{ding2024wolf} and \citet{zeng2024johnnypersuadellmsjailbreak} meticulously construct an attack framework utilizing LLMs to generate deceptive prompts with hidden intentions. To dynamically explore the safety vulnerabilities of target LLMs. Both MART \citep{ge2023mart} and APRT \citep{jiang2024automatedprogressiveredteaming} use a trainable LLM or LLM-based system to attack the target LLMs in a multi-round manner. These sophisticated methods are to utilize the powerful generation capability of LLMs to effectively circumvent the defensive mechanisms of the target LLMs.

\subsubsection{Evaluation}
When target LLMs generate unsafe yet helpful answers in response to sensitive or illegal input prompts, this phenomenon is identified as a successful jailbreaking attempt.
Manual evaluation often requires complex safety definitions \citep{mazeika2024harmbenchstandardizedevaluationframework, jiang2024automatedprogressiveredteaming}, which are effective but time-consuming. 
Attack Success Rate (ASR) is a prevalent and straightforward metric employed to assess the attack capabilities of Red LLMs \citep{DBLP:journals/corr/abs-2307-15043, DBLP:conf/iclr/LiuXCX24}. However, this metric is susceptible to producing false positives, as it relies solely on keyword matching to ascertain maliciousness in responses of the target LLMs.
To mitigate this issue, numerous studies \citep{DBLP:journals/corr/abs-2310-08419, ding2024wolf, jiang2024automatedprogressiveredteaming} have employed the GPT-4 API to verify whether responses generated by target LLMs are unsafe yet relevant to the specified topic.

\subsection{Defense}
To address the aforementioned attacks, researchers have developed various defense strategies, which can be roughly divided into two directions: external safeguard and internal protection. The difference is that external protection does not modify LLMs, while internal enhancement does.

\input{taxonomies/Robustness-Defense_Taxonomy}

\subsubsection{External Safeguard}

External safeguard focuses on protecting an LLM from malicious inputs or external threats by implementing safety measures outside the model. A number of studies propose external safeguard methods to detect unsafe content in the input and output of the given LLM. \citet{DBLP:journals/corr/abs-2309-02705} wipe out tokens one by one for a given prompt, and then if any resulting subsequence or the prompt itself is detected as harmful, the input will be marked as harmful. \citet{DBLP:journals/corr/abs-2402-13494} propose GradSafe, which effectively detects jailbreak prompts by carefully checking the gradient of key parameters in LLM. \citet{DBLP:journals/corr/abs-2309-14348} build a robustly aligned LLM (RALLM) to defend against potential alignment-breaking attacks. \citet{DBLP:journals/corr/abs-2403-04783} assign different roles to LLM agents and use their collaborative monitoring and filtering of responses generated by LLMs to complete defense tasks. Other methods present various techniques for attacks including undesired content, such as classification system \citep{DBLP:conf/aaai/MarkovZANLAJW23}, token-level detection algorithms \citep{DBLP:journals/corr/abs-2311-11509}, toolkit with programmable rails \citep{DBLP:conf/emnlp/RebedeaDSPC23}, and LLM-based input-output safeguard model \citep{DBLP:journals/corr/abs-2312-06674}. 

Another strand of research starts from providing an instructive prompt or modifying a given prompt. \citet{DBLP:journals/natmi/XieYSCLCXW23} provide instructions to guide LLMs to self-check and respond responsibly. \citet{DBLP:journals/corr/abs-2310-06387} provide several examples of safe response to encourage safer outputs from LLMs. \citet{DBLP:journals/corr/abs-2405-20099} use well-designed interpretable suffix prompts, which can effectively defend various standard and adaptive jailbreak techniques. \citet{DBLP:journals/corr/abs-2312-14197} add a reminder to the prompt fed to an LLM, which instructs the LLM not to execute commands embedded in external content, hence avoiding the execution of malicious instructions hidden in external content. 
Given the response generated by a target LLM from its original input prompt, \citet{DBLP:journals/corr/abs-2402-16459} asks an LLM to infer the input prompt that may cause the response through using a ``backtranslation'' prompt. The inferred prompt is called a backtranslation prompt, which tends to reveal the actual intention of the original prompt. \citet{DBLP:journals/corr/abs-2310-03684} randomly interfere a given input prompt, generate multiple copies, and then aggregate/summarize corresponding responses to detect the adversarial input. External threats may employ a variety of attack strategies, making them difficult to predict and guard against. It is challenging to identify highly covert attacks or malicious input and output. And too much reliance on external safeguard may lead to the safety of the whole system is weak. If the external system crashes, the overall safety of the protected LLM would be greatly reduced.

\subsubsection{Internal Protection}
Internal protection involves modifying an LLM during the process of training or tuning to improve its ability to defend against various attacks, reduce risks, and ensure safe behaviors in real-world applications. \citet{DBLP:journals/corr/abs-2405-15202} propose single- and mixed-task losses for instruction tuning and demonstrate that LLMs can significantly boost safe management of risky content via appropriate instruction tuning, thus defending for attacks involving malicious long documents. \citet{DBLP:journals/corr/abs-2306-04959} inject three types of defense functions into the different stages of Federated LLMs in federal learning aggregation to support the defense mechanism for adversarial attacks. \citet{DBLP:journals/corr/abs-2401-10862} demonstrate that employing moderate WANDA pruning can bolster LLM’s defense against jailbreak attacks while obviating the need for fine-tuning. The WANDA pruning \citep{DBLP:conf/iclr/Sun0BK24} involves removing a subset of network weights, with the goal of preserving performance. 

In addition, a considerable number of studies choose fine-tuning \citep{DBLP:journals/corr/abs-2307-09288} or instruction-tuning \citep{DBLP:journals/corr/abs-2310-12505} to strengthen LLM's defense against prompt attacks. \citet{DBLP:journals/corr/abs-2406-06622} introduce a two-stage adversarial tuning framework, which enhances the ability of LLMs to resist unknown jailbreak attacks through iterative improvement of confrontational prompts. \citet{DBLP:journals/corr/abs-2307-09288} collect adversarial prompts and their safety demonstrations, subsequently integrating these samples into the general supervised fine-tuning pipeline. Correspondingly, \citet{DBLP:journals/corr/abs-2312-14197} apply adversarial training to the self-supervised fine-tuning stage of LLMs, teaching them to ignore instructions embedded in external content, thus enhancing their robustness to indirect prompt injection attacks. Despite effectiveness, internal protection could increase the complexity of LLMs, and reduce their interpretability and maintainability. 

It is worth mentioning that current defense strategies focus too much on defending attacks and ignore the reduction of effectiveness. \citet{DBLP:journals/corr/abs-2401-00287} propose that the ideal defense strategy should make LLMs safe against ``unsafe prompts'' rather than over-defense on ``safe prompt''. They also propose an evaluation benchmark termed Safety and Over-Defensiveness Eval (SODE), and their experiment results lead to important findings. For example, self-checking does improve the safety of inputs, but at the cost of extremely excessive defense.

%% file: taxonomies/Robustness-Jailbreaking_Taxonomy.tex
\tikzstyle{my-box}=[
    rectangle,
    draw=hidden-draw,
    rounded corners,
    text opacity=1,
    minimum height=1.5em,
    minimum width=5em,
    inner sep=2pt,
    align=center,
    fill opacity=.5,
    line width=0.8pt,
]
\tikzstyle{leaf}=[my-box, minimum height=1.5em,
    fill=hidden-pink!80, text=black, align=center,font=\normalsize,
    inner xsep=2pt,
    inner ysep=4pt,
    line width=0.8pt,
]
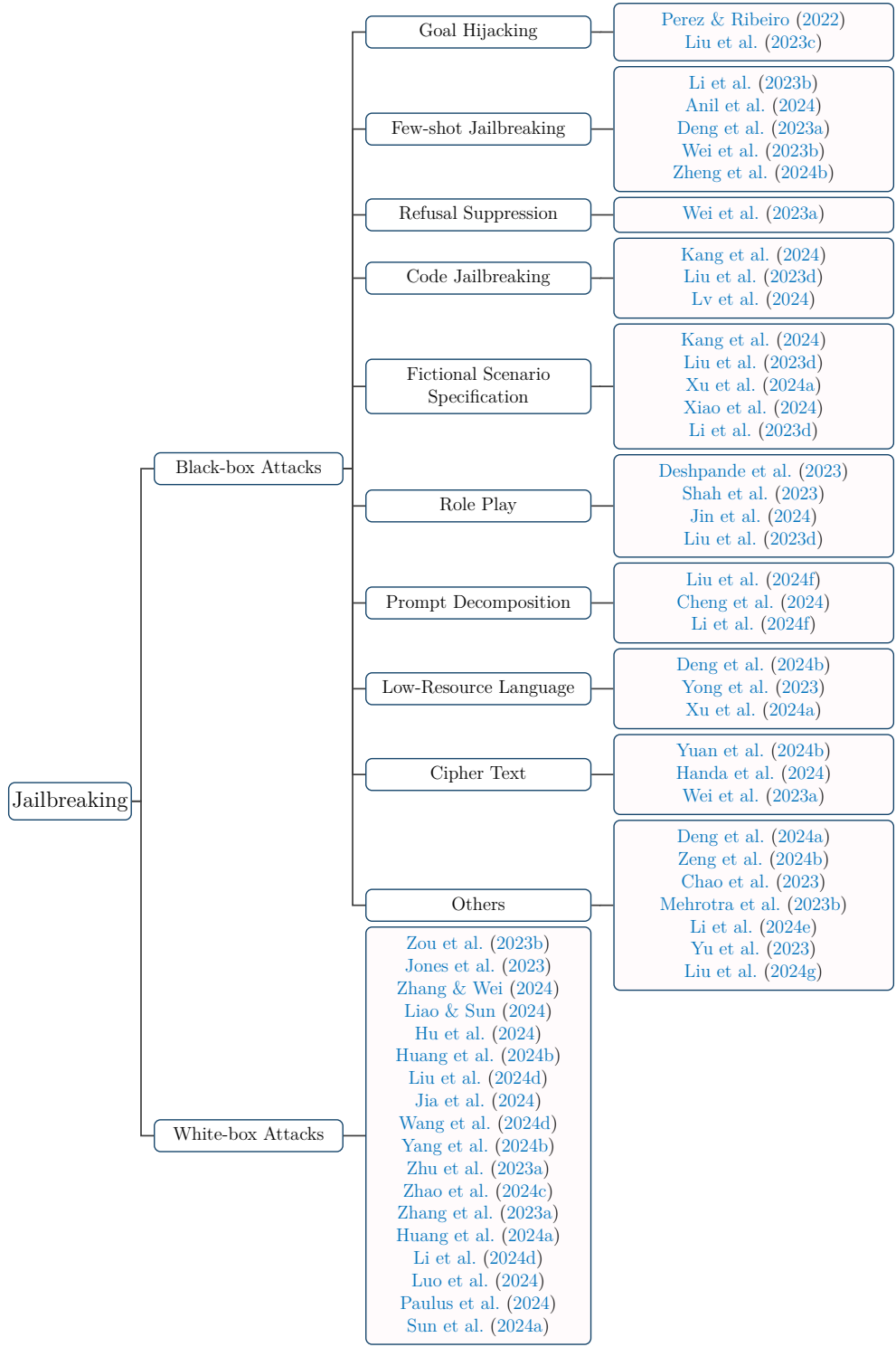
\begin{figure*}[t!]
    \centering
    \resizebox{0.8\textwidth}{!}{
        \begin{forest}
            forked edges,
            for tree={
                grow=east,
                reversed=true,
                anchor=base west,
                parent anchor=east,
                child anchor=west,
                base=center,
                font=\large,
                rectangle,
                draw=hidden-draw,
                rounded corners,
                align=center,
                text centered,
                minimum width=5em,
                edge+={darkgray, line width=1pt},
                s sep=3pt,
                inner xsep=2pt,
                inner ysep=3pt,
                line width=0.8pt,
                ver/.style={rotate=90, child anchor=north, parent anchor=south, anchor=center},
            },
            where level=1{text width=10em,font=\normalsize,}{},
            where level=2{text width=12em,font=\normalsize,}{},
            where level=3{text width=15em,font=\normalsize,}{},
            where level=4{text width=13.5em,font=\normalsize,}{},
            [
                Jailbreaking,
                [
                    Black-box Attacks,
                    [
                        Goal Hijacking,
                        [
                            \citet{DBLP:journals/corr/abs-2211-09527}\\
                            \citet{DBLP:journals/corr/abs-2306-05499},
                            leaf
                        ]
                    ]
                    [
                        Few-shot Jailbreaking,
                        [
                            \citet{DBLP:conf/emnlp/LiGFXHMS23}\\
                            \citet{anil2024manyshot}\\
                            \citet{DBLP:conf/emnlp/DengWFDW023}\\
                            \citet{DBLP:journals/corr/abs-2310-06387}\\
                            \citet{DBLP:journals/corr/abs-2406-01288},
                            leaf
                        ]
                    ]
                    [
                        Refusal Suppression,
                        [
                            \citet{DBLP:conf/nips/0001HS23},
                            leaf
                        ]
                    ]
                    [
                        Code Jailbreaking,
                        [
                            \citet{DBLP:conf/sp/KangLSGZH24}\\
                            \citet{DBLP:journals/corr/abs-2305-13860}\\
                            \citet{DBLP:journals/corr/abs-2402-16717},
                            leaf
                        ]
                    ]
                    [
                        Fictional Scenario \\Specification,
                        [
                            \citet{DBLP:conf/sp/KangLSGZH24}\\
                            \citet{DBLP:journals/corr/abs-2305-13860}\\
                            \citet{DBLP:conf/naacl/XuWZLXC24}\\
                            \citet{DBLP:conf/emnlp/XiaoY0C24}\\
                            \citet{DBLP:journals/corr/abs-2311-03191},
                            leaf
                        ]
                    ]
                    [
                        Role Play,
                        [
                            \citet{DBLP:conf/emnlp/DeshpandeMRKN23}\\
                            \citet{DBLP:journals/corr/abs-2311-03348}\\
                            \citet{jin2024quack}\\
                            \citet{DBLP:journals/corr/abs-2305-13860},
                            leaf
                        ]
                    ]
                    [
                        Prompt Decomposition,
                        [
                            \citet{DBLP:conf/uss/LiuZZDM024}\\
                            \citet{DBLP:journals/corr/abs-2402-09177}\\
                            \citet{DBLP:conf/emnlp/LiWCZH24},
                            leaf
                        ]
                    ]
                    [
                        Low-Resource Language,
                        [
                            \citet{DBLP:conf/iclr/0010ZPB24}\\
                            \citet{DBLP:journals/corr/abs-2310-02446}\\
                            \citet{DBLP:conf/naacl/XuWZLXC24},
                            leaf
                        ]
                    ]
                    [
                        Cipher Text,
                        [
                            \citet{DBLP:conf/iclr/YuanJW0H0T24}\\
                            \citet{handa2024competencyreasoningopensdoor}\\
                            \citet{DBLP:conf/nips/0001HS23},
                            leaf
                        ]
                    ]
                    [
                        Others,
                        [
                            \citet{DBLP:conf/ndss/DengLLWZLW0L24}\\
                            \citet{DBLP:conf/acl/0005LZY0S24}\\
                            \citet{DBLP:journals/corr/abs-2310-08419}\\
                            \citet{DBLP:journals/corr/abs-2312-02119}\\
                            \citet{DBLP:journals/corr/abs-2402-14872}\\
                            \citet{DBLP:journals/corr/abs-2309-10253}\\
                            \citet{DBLP:conf/iclr/LiuXCX24},
                            leaf
                        ]
                    ]
                ]
                [
                    White-box Attacks,
                    [
                        \citet{DBLP:journals/corr/abs-2307-15043}\\
                        \citet{DBLP:conf/icml/JonesDRS23}\\
                        \citet{DBLP:journals/corr/abs-2405-01229}\\
                        \citet{DBLP:journals/corr/abs-2404-07921}\\
                        \citet{DBLP:journals/corr/abs-2405-09113}\\
                        \citet{DBLP:journals/corr/abs-2405-14189}\\
                        \citet{DBLP:journals/corr/abs-2410-15645}\\
                        \citet{DBLP:journals/corr/abs-2405-21018}\\
                        \citet{DBLP:journals/corr/abs-2410-09040}\\
                        \citet{DBLP:journals/corr/abs-2407-03160}\\
                        \citet{DBLP:journals/corr/abs-2310-15140}\\
                        \citet{DBLP:journals/corr/abs-2401-17256}\\
                        \citet{DBLP:journals/corr/abs-2310-01581}\\
                        \citet{DBLP:conf/iclr/HuangGXL024}\\
                        \citet{DBLP:journals/corr/abs-2401-06824}\\
                        \citet{DBLP:journals/corr/abs-2410-10150}\\
                        \citet{DBLP:journals/corr/abs-2404-16873}\\
                        \citet{DBLP:journals/corr/abs-2410-18469},
                        leaf
                    ]
                ]
            ]
        \end{forest}
    }
    \caption{Overview of Jailbreaking of LLMs.}
    \label{fig:Overview_of_Jailbreaking}
\end{figure*}

%% file: taxonomies/Robustness-Red_Teaming_Taxonomy.tex
\tikzstyle{my-box}=[
    rectangle,
    draw=hidden-draw,
    rounded corners,
    text opacity=1,
    minimum height=1.5em,
    minimum width=5em,
    inner sep=2pt,
    align=center,
    fill opacity=.5,
    line width=0.8pt,
]
\tikzstyle{leaf}=[my-box, minimum height=1.5em,
    fill=hidden-pink!80, text=black, align=center,font=\normalsize,
    inner xsep=2pt,
    inner ysep=4pt,
    line width=0.8pt,
]
\begin{figure*}[t!]
    \centering
    \resizebox{\textwidth}{!}{
        \begin{forest}
            forked edges,
            for tree={
                grow=east,
                reversed=true,
                anchor=base west,
                parent anchor=east,
                child anchor=west,
                base=center,
                font=\large,
                rectangle,
                draw=hidden-draw,
                rounded corners,
                align=center,
                text centered,
                minimum width=5em,
                edge+={darkgray, line width=1pt},
                s sep=3pt,
                inner xsep=2pt,
                inner ysep=3pt,
                line width=0.8pt,
                ver/.style={rotate=90, child anchor=north, parent anchor=south, anchor=center},
            },
            where level=1{text width=12em,font=\normalsize,}{},
            where level=2{text width=15em,font=\normalsize,}{},
            where level=3{text width=15em,font=\normalsize,}{},
            where level=4{text width=15em,font=\normalsize,}{},
            [Red Teaming
                [Manual Red Teaming
                  [\citet{DBLP:journals/corr/abs-2307-09288}, leaf
                  [, phantom]]
                  [\citet{DBLP:journals/corr/abs-2204-05862}, leaf
                  [, phantom]]
                ]
                [Automated Red Teaming
                  [Template-Based Automated\\Red Teaming
                    [\citet{DBLP:conf/emnlp/ShinRLWS20}, leaf]
                    [\citet{wallace2021universaladversarialtriggersattacking}, leaf]
                    [\citet{DBLP:journals/corr/abs-2307-15043}, leaf]
                    [\citet{DBLP:conf/icml/JonesDRS23}, leaf]
                    [\citet{zhu2023autodan}, leaf]
                    [\citet{alon2023detectinglanguagemodelattacks}, leaf]
                    [\citet{wu2023deceptprompt}, leaf]
                    [\citet{DBLP:conf/iclr/LiuXCX24}, leaf]
                  ]
                  [Generation-Based\\Red Teaming
                    [\citet{DBLP:journals/corr/abs-2310-08419}, leaf]
                    [\citet{mehrotra2023tree}, leaf]
                    [\citet{ding2024wolf}, leaf]
                    [\citet{zeng2024johnnypersuadellmsjailbreak}, leaf]
                    [\citet{ge2023mart}, leaf]
                    [\citet{jiang2024automatedprogressiveredteaming}, leaf]
                  ]
                ]
                [Evaluation
                  [\citet{mazeika2024harmbenchstandardizedevaluationframework}, leaf
                  [, phantom]]
                  [\citet{jiang2024automatedprogressiveredteaming}, leaf
                  [, phantom]]
                ]
            ]
        \end{forest}
    }
    \caption{Overview of Red Teaming of LLMs.}
    \label{fig:Overview_of_red_teaming}
\end{figure*}
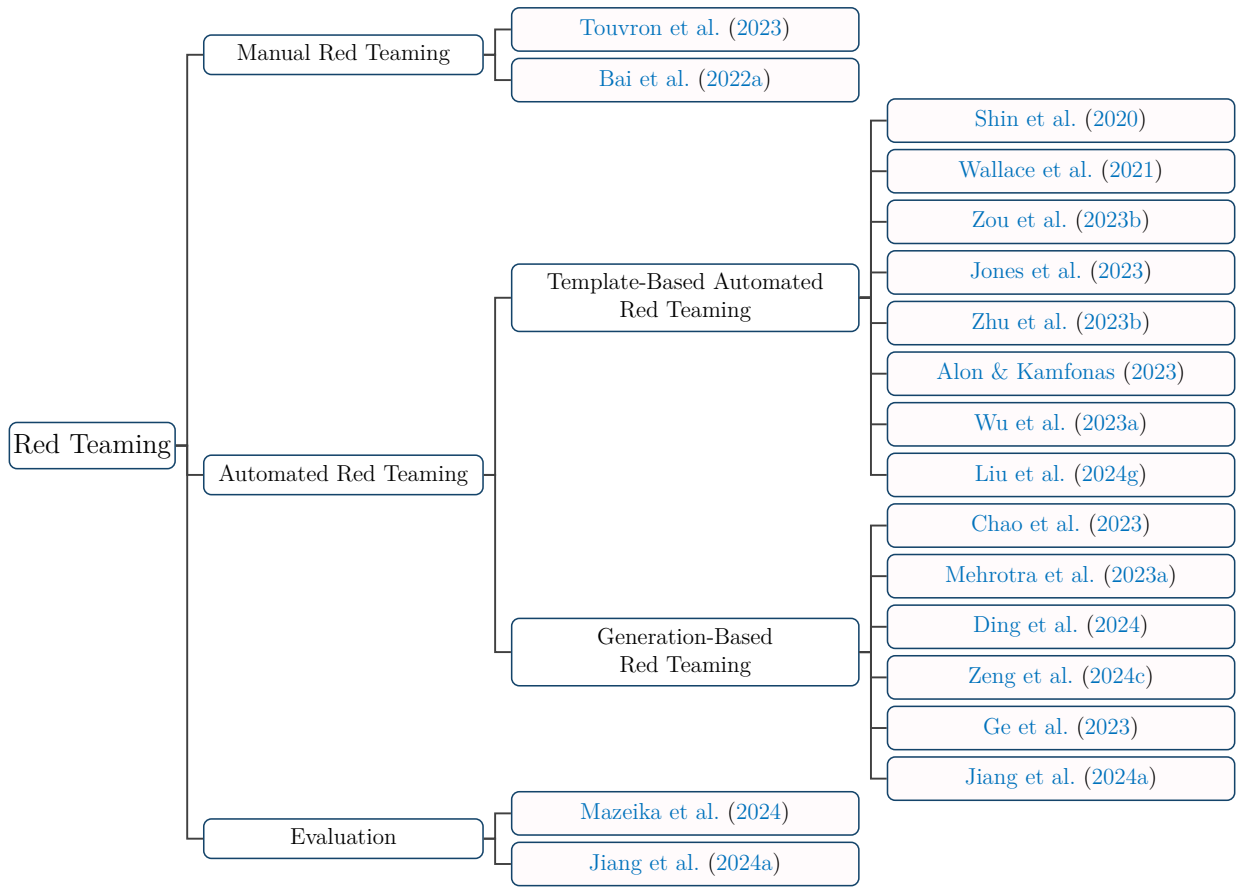

%% file: taxonomies/Robustness-Defense_Taxonomy.tex
\tikzstyle{my-box}=[
    rectangle,
    draw=hidden-draw,
    rounded corners,
    text opacity=1,
    minimum height=1.5em,
    minimum width=5em,
    inner sep=2pt,
    align=center,
    fill opacity=.5,
    line width=0.8pt,
]
\tikzstyle{leaf}=[my-box, minimum height=1.5em,
    fill=hidden-pink!80, text=black, align=center,font=\normalsize,
    inner xsep=2pt,
    inner ysep=4pt,
    line width=0.8pt,
]
\begin{figure*}[t!]
    \centering
    \resizebox{\textwidth}{!}{
        \begin{forest}
            forked edges,
            for tree={
                grow=east,
                reversed=true,
                anchor=base west,
                parent anchor=east,
                child anchor=west,
                base=center,
                font=\large,
                rectangle,
                draw=hidden-draw,
                rounded corners,
                align=center,
                text centered,
                minimum width=5em,
                edge+={darkgray, line width=1pt},
                s sep=3pt,
                inner xsep=2pt,
                inner ysep=3pt,
                line width=0.8pt,
                ver/.style={rotate=90, child anchor=north, parent anchor=south, anchor=center},
            },
            where level=1{text width=8em,font=\normalsize,}{},
            where level=2{text width=15em,font=\normalsize,}{},
            where level=3{text width=15em,font=\normalsize,}{},
            where level=4{text width=15em,font=\normalsize,}{},
            [
            Defense,
            [External\\ Safeguard,
            [Content Detecting,
            [
             \citet{DBLP:journals/corr/abs-2309-02705}\\
                     \citet{DBLP:journals/corr/abs-2402-13494}\\
                     \citet{DBLP:journals/corr/abs-2309-14348}\\
                     \citet{DBLP:journals/corr/abs-2403-04783}\\
                    \citet{DBLP:conf/aaai/MarkovZANLAJW23}\\
                    \citet{DBLP:journals/corr/abs-2311-11509}\\
                     \citet{DBLP:conf/emnlp/RebedeaDSPC23}\\
                     \citet{DBLP:journals/corr/abs-2312-06674}
                      , leaf
            ]]
            [Prompt Engineering,
            [
            \citet{DBLP:journals/natmi/XieYSCLCXW23}\\
            \citet{DBLP:journals/corr/abs-2310-06387}\\
            \citet{DBLP:journals/corr/abs-2405-20099}\\
            \citet{DBLP:journals/corr/abs-2312-14197}\\
            \citet{DBLP:journals/corr/abs-2402-16459}\\
            \citet{DBLP:journals/corr/abs-2310-03684}
            , leaf
            ]]
            ]
            [Internal\\ Protection,
            [
            \citet{DBLP:journals/corr/abs-2405-15202} \\
            \citet{DBLP:journals/corr/abs-2306-04959} \\
            \citet{DBLP:journals/corr/abs-2401-10862}\\
            \citet{DBLP:conf/iclr/Sun0BK24}\\
            \citet{DBLP:journals/corr/abs-2307-09288}\\
            \citet{DBLP:journals/corr/abs-2406-06622}\\
            \citet{DBLP:journals/corr/abs-2312-14197}
            , leaf
            ]
            ]
            ]
        \end{forest}
    }
    \caption{Overview of Defense Methods against Attack for LLMs.}
    \label{fig:Overview_of_defense}
\end{figure*}
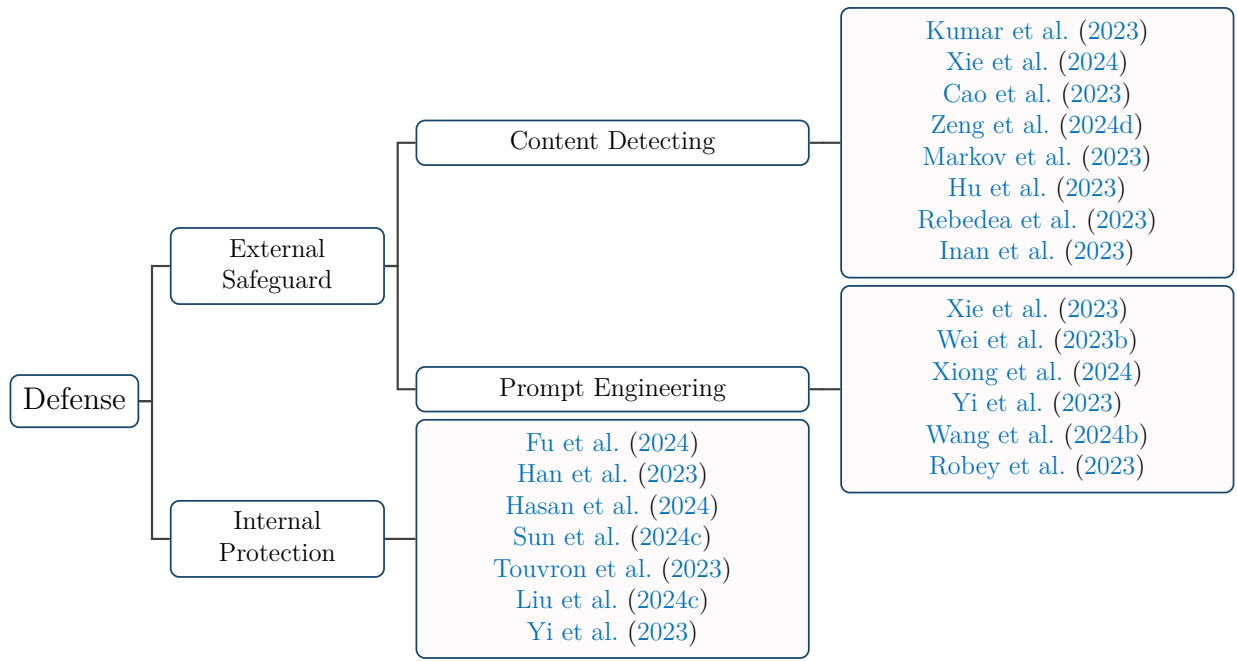

%% file: taxonomies/Misuse_Taxonomy.tex
\tikzstyle{my-box}=[
    rectangle,
    draw=hidden-draw,
    rounded corners,
    text opacity=1,
    minimum height=1.5em,
    minimum width=5em,
    inner sep=2pt,
    align=center,
    fill opacity=.5,
    line width=0.8pt,
]
\tikzstyle{leaf}=[my-box, minimum height=1.5em,
    fill=hidden-pink!80, text=black, align=center,font=\normalsize,
    inner xsep=2pt,
    inner ysep=4pt,
    line width=0.8pt,
]
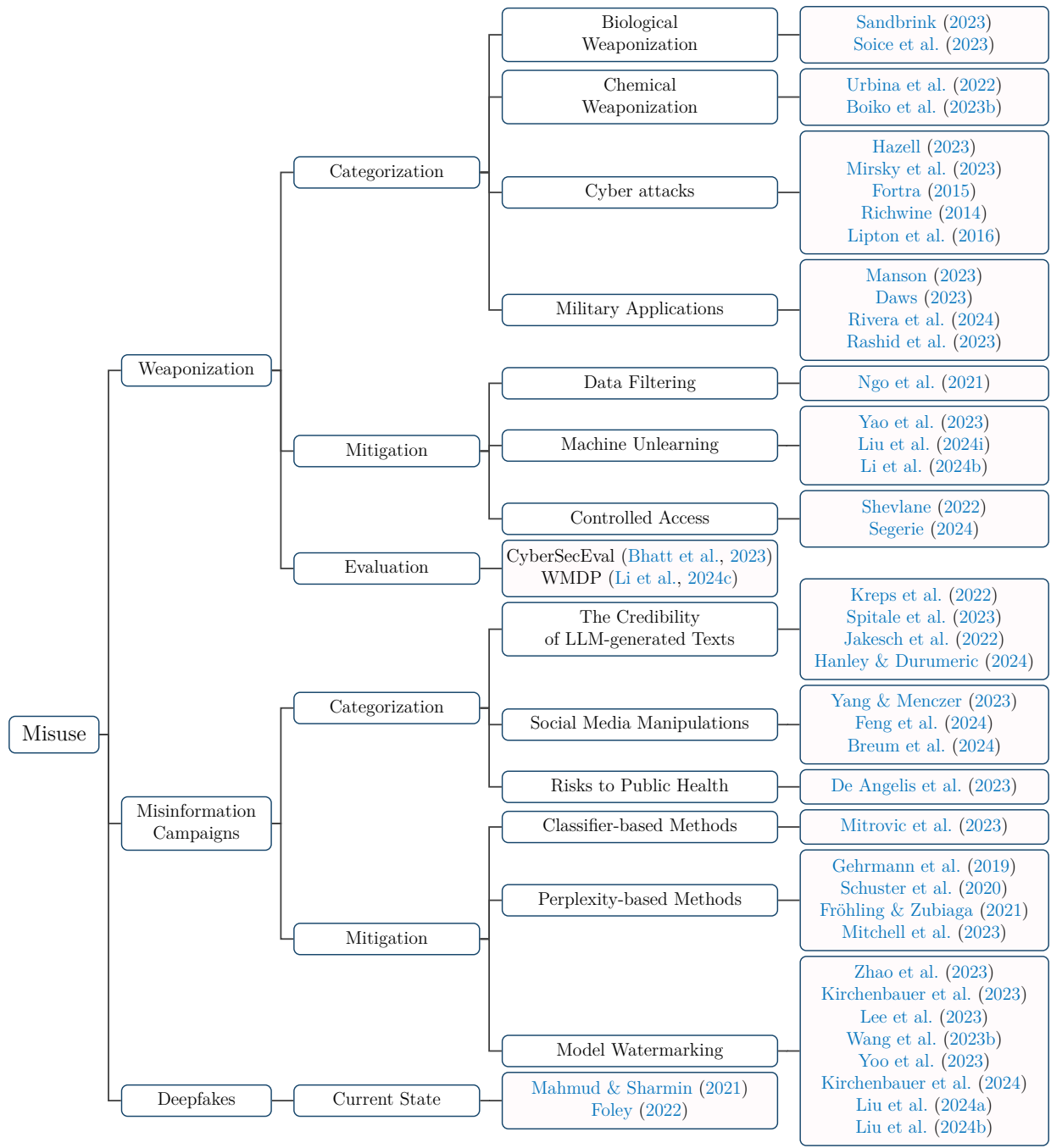
\begin{figure*}[t!]
    \centering
    \resizebox{\textwidth}{!}{
        \begin{forest}
            forked edges,
            for tree={
                grow=east,
                reversed=true,
                anchor=base west,
                parent anchor=east,
                child anchor=west,
                base=center,
                font=\large,
                rectangle,
                draw=hidden-draw,
                rounded corners,
                align=center,
                text centered,
                minimum width=5em,
                edge+={darkgray, line width=1pt},
                s sep=3pt,
                inner xsep=2pt,
                inner ysep=3pt,
                line width=0.8pt,
                ver/.style={rotate=90, child anchor=north, parent anchor=south, anchor=center},
            },
            where level=1{text width=8em,font=\normalsize,}{},
            where level=2{text width=10em,font=\normalsize,}{},
            where level=3{text width=15em,font=\normalsize,}{},
            where level=4{text width=13.5em,font=\normalsize,}{},
            [
                Misuse
                [
                    Weaponization
                    [
                        Categorization
                        [
                            Biological \\ Weaponization
                            [
                                \citet{DBLP:journals/corr/abs-2306-13952} \\
                                \citet{DBLP:journals/corr/abs-2306-03809} \\
                                , leaf
                            ]
                        ]
                        [
                            Chemical \\ Weaponization
                            [
                                \citet{DBLP:journals/natmi/UrbinaLIE22} \\
                                \citet{DBLP:journals/nature/BoikoMKG23} \\
                                , leaf
                            ]
                        ]
                        [
                            Cyber attacks
                            [
                                \citet{DBLP:journals/corr/abs-2305-06972} \\
                                \citet{DBLP:journals/compsec/MirskyDKSGYZPLE23} \\
                                \citet{SonyHackers} \\
                                \citet{CyberAttackCould} \\
                                \citet{ThePerfectWeapon} \\
                                , leaf
                            ]
                        ]
                        [
                            Military Applications
                            [
                                \citet{TheUSMilitary} \\
                                \citet{PalantirDemos} \\
                                \citet{DBLP:conf/fat/RiveraMRLSS24} \\
                                \cite{DBLP:journals/ijis/RashidKSB23} \\
                                , leaf
                            ]
                        ]
                    ]
                    [
                        Mitigation
                        [
                            Data Filtering
                            [
                                \citet{DBLP:journals/corr/abs-2108-07790} \\
                                ,leaf
                            ]
                        ]
                        [
                            Machine Unlearning
                            [
                                \citet{yao2023large} \\
                                \citet{DBLP:conf/acl/0010DT0024} \\
                                \citet{DBLP:conf/icml/LiPGYBGLDGMHLJL24} \\
                                ,leaf
                            ]
                        ]
                        [
                            Controlled Access
                            [
                                \citet{DBLP:journals/corr/abs-2201-05159} \\
                                \citet{AISafetyStrategiesLandscape} \\
                                ,leaf
                            ]
                        ]
                    ]
                    [
                        Evaluation
                        [
                            CyberSecEval \citep{DBLP:journals/corr/abs-2312-04724} \\
                            WMDP \citep{DBLP:journals/corr/abs-2403-03218} \\
                            , leaf
                        ]
                    ]
                ]
                [
                    Misinformation \\ Campaigns
                    [
                        Categorization
                        [
                            The Credibility \\ of LLM-generated Texts
                            [
                                \citet{kreps2022all} \\
                                \citet{DBLP:journals/corr/abs-2301-11924} \\
                                \citet{DBLP:journals/corr/abs-2206-07271} \\
                                \citet{DBLP:conf/icwsm/HanleyD24a} \\
                                , leaf
                            ]
                        ]
                        [
                            Social Media Manipulations
                            [
                                \citet{DBLP:journals/corr/abs-2307-16336} \\
                                \citet{DBLP:journals/corr/abs-2402-00371} \\
                                \citet{DBLP:conf/icwsm/BreumEMMA24} \\
                                , leaf
                            ]
                        ]
                        [
                            Risks to Public Health
                            [
                                \citet{de2023chatgpt} \\
                                ,leaf
                            ]
                        ]
                    ]
                    [
                        Mitigation
                        [
                            Classifier-based Methods
                            [
                                \citet{DBLP:journals/corr/abs-2301-13852} \\
                                , leaf
                            ]
                        ]
                        [
                            Perplexity-based Methods
                            [
                                 \citet{DBLP:conf/acl/GehrmannSR19} \\
                                 \citet{DBLP:journals/coling/SchusterSSB20} \\
                                 \citet{DBLP:journals/peerj-cs/FrohlingZ21} \\
                                 \citet{DBLP:conf/icml/Mitchell0KMF23} \\
                                 , leaf
                            ]
                        ]
                        [
                            Model Watermarking
                            [
                                \citet{DBLP:conf/icml/ZhaoWL23} \\
                                \citet{DBLP:conf/icml/KirchenbauerGWK23} \\
                                \citet{DBLP:journals/corr/abs-2305-15060} \\
                                \citet{DBLP:journals/corr/abs-2307-15992} \\
                                \citet{DBLP:conf/acl/YooAJK23} \\
                                \citet{DBLP:conf/iclr/KirchenbauerGWS24} \\
                                \citet{DBLP:conf/iclr/LiuPH0WKY24} \\
                                \citet{DBLP:conf/iclr/LiuPHM024} \\
                                , leaf
                            ]
                        ]
                    ]
                ]
                [
                    Deepfakes
                    [
                        Current State
                        [
                            \citet{DBLP:journals/corr/abs-2105-00192} \\
                            \citet{23_of_the_best_deepfake} \\
                            , leaf
                        ]
                    ]
                ]
            ]
        \end{forest}
    }
    \caption{Overview of misuse of LLMs.}
    \label{fig:Overview_of_catastrophic_misuses}
\end{figure*}

%% file: sections/Misuse.tex
\section{Misuse}
\label{section:CatastrophicMisuse}
In this section, we focus on the catastrophic risks associated with misuse, specifically referring to risks that involve consequences of extreme scale or severity resulting from the inappropriate or harmful application of LLMs. Specifically, such misuse of LLMs could result in the loss of safety or even lives on a massive scale, potentially impacting tens of thousands of individuals, or causing significant harm to social and political stability. For instance, the emergence and proliferation of novel viruses, a marked deterioration in the quality of public discourse, or the widespread disempowerment of the public, governments, and other human-led institutions.
 
According to current studies, catastrophic misuses of LLMs can be categorized into three primary categories: Weaponization, Misinformation Campaigns, and Deepfakes, reflecting the various ways in which LLMs could be misused across different domains and the potentially catastrophic consequences that may result, as illustrated in Figure \ref{fig:Overview_of_catastrophic_misuses}. These categories encompass a wide range of issues, from the abuse of technology to information manipulation, highlighting the potential harm that LLMs could inflict on society, politics, and public safety. 

\subsection{Weaponization}

Owing to their vast domain-specific knowledge (e.g., in biology and chemistry), LLMs hold significant potential to drive substantial advancements in fields such as the life sciences. However, these same capabilities may also facilitate harmful or malicious applications. Specifically, LLMs could contribute to, or substantially aid, scientific discoveries that unlock novel weapons. LLMs, such as GPT-4 and its successors, might provide dual-use information, lowering some barriers that historical weapons efforts encounter. As LLMs evolve into multimodal lab assistants and autonomous scientific tools, they could empower non-experts to conduct laboratory work and engage in malicious activities.

\subsubsection{Risks of Misuse in Weapons Acquisition}

\paragraph{Biological Weaponization} Firstly, the risks of weaponization misuse, particularly in the biological domain, represent a significant concern in the context of LLMs.

On one hand, LLMs may enable efficient learning about “dual-use” knowledge which can be used for informing legitimate research but also for causing harm. In contrast to internet search engines, LLMs can aggregate information across many different sources, making complex knowledge accessible and tailored to non-experts, and proactively point out variables that users do not know to inquire about. If biological weapons-enabling information is presented in this way, this could enable small efforts to overcome key bottlenecks. For instance, at the time of testing in June 2023, ChatGPT readily outlined the importance of “harvesting and separation” of toxin-containing supernatant from cells and further steps for concentration, purification, and formulation \citep{DBLP:journals/corr/abs-2306-13952}. Even more,  \citet{DBLP:journals/corr/abs-2306-03809} find that LLMs enabled non-scientist students to identify four potential pandemic pathogens, provided information on how they can be synthesized, and further helped them cause a widespread epidemic of pandemic-class agents. In the long term, as LLMs and related AI tools improve their ability to do scientific research with minimal human input, this could remove relevant barriers to biological weapons development.

On the other hand, LLMs may be combined with existing tools in the biological field to raise the ceiling of harm from biological agents. For example, LLMs combined with biological design tools (BDT) which are systems trained on biological data to help design proteins or other biological agents, may increase the potential for harm from biological agents and make them more accessible \citep{DBLP:journals/corr/abs-2306-13952}.

\paragraph{Chemical Weaponization} Secondly, LLMs may in particular lower barriers to chemical misuse. A key theme is that LLMs could increase the accessibility to existing knowledge and capabilities, and thus may lower the barriers to chemical misuse. \citet{DBLP:journals/natmi/UrbinaLIE22} provide experimental evidence for the dual role of artificial intelligence in drug discovery. They indicate that although generating toxic substances requires expertise in chemistry or toxicology, the technical threshold is greatly lowered when these fields are combined with AI models. Moreover, \citet{DBLP:journals/nature/BoikoMKG23} demonstrate advanced capabilities for (semi-)autonomous chemical experimental design and execution of an AI system Coscientist driven by GPT-4, revealing its potential for the synthesis of hazardous chemicals.

\paragraph{Cyber Attacks} Another noteworthy risk posed by LLMs is that these systems could potentially be used to conduct cyber attacks. LLMs could potentially be harnessed to scale individualized cyber attacks, such as spear phishing campaigns, a form of cybercrime that involves manipulating targets into divulging sensitive information \citep{DBLP:journals/corr/abs-2305-06972}. LLMs could also be used by cyber attackers to evade detection systems since they can effectively cover the attacker's tracks \citep{DBLP:journals/compsec/MirskyDKSGYZPLE23}.

A successful cyber attack can have catastrophic consequences in the economy or politics. For instance, in 2014, hackers accessed sensitive data through a spear-phishing attack targeting Sony executives \citep{SonyHackers}, resulting in estimated damages ranging from \$70 million to \$100 million \citep{CyberAttackCould}. Another prominent case is the breach of a private email account belonging to the chairperson of Hillary Clinton’s 2016 presidential campaign, which was compromised by hackers using a phishing attack \citep{ThePerfectWeapon}.

\paragraph{Military Applications} With the spread of LLMs that can generate novel strategies and decisions based on prompts and supplied information, debates on the integration of autonomous agents in high-stake situations such as military and diplomatic decision-making have become more frequent. In July 2023, Bloomberg reported that the US Department of Defense (DoD) was conducting a set of tests in which they evaluated five different LLMs for their military planning capacities in a simulated conflict scenario \citep{TheUSMilitary}. In addition, multiple companies such as Palantir and Scale AI are working on LLM-based military decision systems for the US government \citep{PalantirDemos}.

The increased exploration of LLMs in high-stakes scenarios highlights the need to understand their behavior in the context of military and nuclear weapons development. \citet{DBLP:conf/fat/RiveraMRLSS24} simulate interactions among multiple AI-controlled ``nation agents'' using several advanced LLMs, which demonstrate that LLM-powered agents frequently opt for escalatory actions, leading to greater conflict, even to the deployment of nuclear weapons. These results suggest that decisions yielded by LLMs in volatile situations could be dangerously unpredictable, raising concerns on their deployment in real-world military operations.

Given these risks, experts recommend extreme caution when integrating LLMs into military or foreign-policy decision-making systems. \cite{DBLP:journals/ijis/RashidKSB23} argue that the speed of AI could speed up warfare and shorten decision-making time. This could result in human commanders being unable to react in a timely manner, raising the ethical and tactical dilemma of delegating control to autonomous systems. There are ongoing debates about whether AI should have any autonomous control over actions like launching weapons, as the potential for escalations could lead to catastrophic consequences.

\subsubsection{Mitigation Methods for Weaponized Misuse}

\begin{figure}[t]
    \centering
    \includegraphics[width=0.62\textwidth]{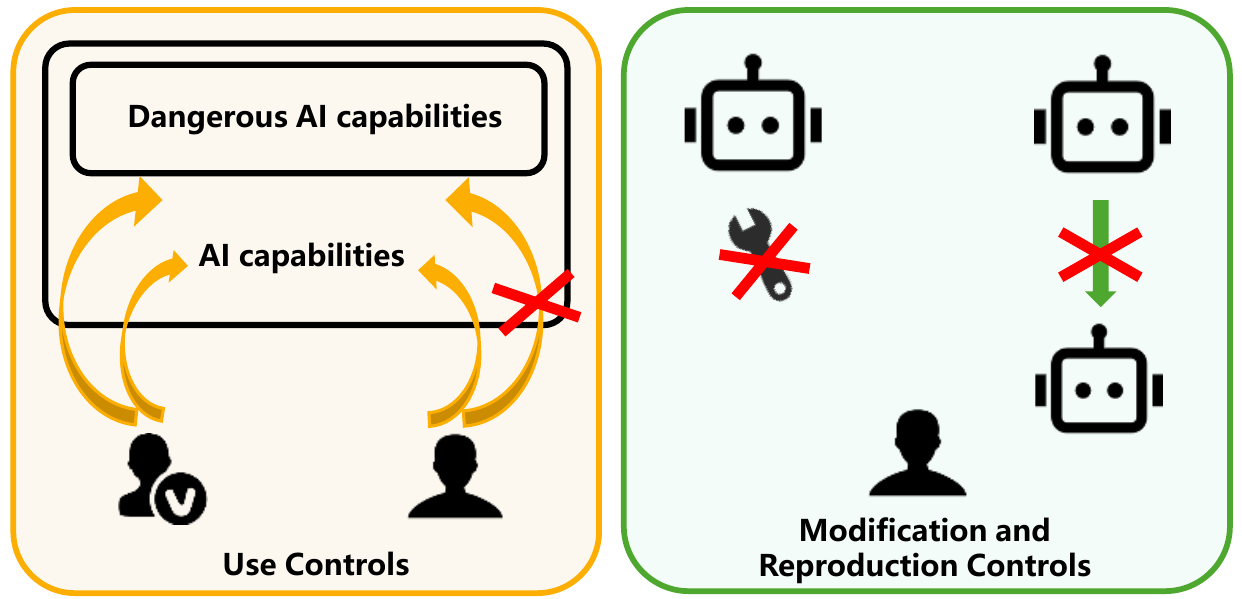}
    \caption{Establishing controlled access between AI systems and users to prevent the risk of AI systems being misused.
}
    \label{fig:controlled_access}
\end{figure}

\paragraph{Data Filtering and Machine Unlearning} To reduce the presence of dangerous information in LLMs and thereby mitigate potential misuse, one proposal is to filter hazardous content from the pre-training data \citep{DBLP:journals/corr/abs-2108-07790}. Yet another line is to adopt machine unlearning methods to remove harmful knowledge from LLMs \citep{yao2023large, DBLP:conf/acl/0010DT0024, DBLP:conf/icml/LiPGYBGLDGMHLJL24}. Such unlearning methods aim to eliminate dangerous information embedded in LLMs’ parameters, thereby preventing the generation of content that could be exploited by malicious actors. However, as demonstrated by \citet{DBLP:journals/corr/abs-2402-14015}, existing unlearning methods struggle to remove knowledge without access to all relevant training data. Addressing these challenges remains a critical focus for ongoing research in this area.

\paragraph{Controlled Access} In addition to reducing hazardous knowledge embedded in LLMs, many researchers advocate building controlled interactions between AI systems and users through technical means. The aim is to prevent dangerous AI capabilities from being widely accessible, whilst preserving access to AI capabilities that can be used safely. As shown in Figure \ref{fig:controlled_access}, as much control as possible is needed over two broad categories: (1) \emph{use controls}, which govern the direct use of AI systems in terms of (who, what, when, where, why, how); and (2) \emph{modification and reproduction controls}, which prevent unauthorized users from altering AI systems or building their own versions in a way that circumvents the use controls \citep{DBLP:journals/corr/abs-2201-05159}.

In application scenarios, a monitored APIs strategy should be employed to place high-risk models behind application programming interfaces (APIs) \citep{AISafetyStrategiesLandscape}. This is to control access to AI models that could pose extreme risks, while monitoring their activity. For instance, with this strategy, OpenAI’s API platform limits the ways in which the GPT-3 models could be used, and Google Cloud’s Vision API is another application example. By implementing monitored APIs, access to these high-risk technologies is restricted to authorized users only. This strategy is akin to digital containment, limiting the potential for AI weaponization or misuse through stringent access controls. Moreover, this method also allows for detailed tracking of how AI models are being utilized, enabling the early detection of misuse patterns.

\emph{Debates on the Public Availability of Model Weights} In the context of controlled access, whether the disclosure of model weights should be controlled to mitigate misuse triggers widespread debate with no consensus yet reached. Consenters argue that to prevent future large-scale harm from advanced AI, it is necessary to limit the spread of model weights \citep{DBLP:journals/corr/abs-2310-18233}. Such an argument gains supporting evidence from recent studies, which have explored the robustness of safety training of LLMs by subversively fine-tuning models on small amounts of synthetic data containing harmful instructions and responses. These studies show that as long as the weights of the model are accessible, it is easy to destroy these safety-aligned LLMs at a very low cost, enabling them to respond to harmful requests in the vast majority of cases \citep{DBLP:journals/corr/abs-2310-02949, DBLP:journals/corr/abs-2310-20624}. Such findings raise concerns that malicious actors could misuse the capabilities of models by subversively fine-tuning them for harmful purposes. Given that fine-tuning is orders of magnitude cheaper than training an LLM from scratch, a broad range of users could engage in such activities. Consequently, it is advocated for AI developers to carefully assess the risks associated with fine-tuning before deciding to release model weights.

On the other hand, proponents of open access to model weights argue that the long-term benefits of generative AI outweigh the associated safety risks. Open-sourcing model weights can accelerate scientific breakthroughs by empowering developers to advance research and innovation. It can also promote the creation of innovative products and services by enhancing hypothesis generation, data analysis, and complex system simulations. In addition, disclosing model weights and resources within the AI community can reduce redundant computational efforts across the whole industry, leading to lower energy consumption and fostering sustainable development \citep{DBLP:journals/corr/abs-2405-08597}.

\subsubsection{Evaluation}

Model evaluation is a necessary component of the governance infrastructure needed to combat catastrophic misuse. DeepMind's research emphasizes that model evaluation for extreme risks should be a priority area for AI safety and governance \citep{DBLP:journals/corr/abs-2305-15324}.

CyberSecEval \citep{DBLP:journals/corr/abs-2312-04724} is a comprehensive benchmark developed to evaluate the cybersecurity of LLMs employed as coding assistants. Its automated test case generation and evaluation pipeline covers a broad scope, equipping LLM designers and researchers with a tool to comprehensively measure and enhance the cybersecurity safety properties of LLMs. Another evaluation benchmark, WMDP, contains 4,157 multiple-choice questions that evaluate the risks of LLMs in empowering malicious actors in developing biological, cyber, and chemical weapons \citep{DBLP:journals/corr/abs-2403-03218}.

\subsection{Misinformation Campaigns}
\label{section_misinformation_campaigns}

The advent of LLMs has significantly transformed the way information is created and increased the speed at which it is disseminated, offering new opportunities for both legitimate and malicious uses. Regarding misuse in information creation and spreading, misinformation campaigns driven by LLMs have gained prominence, posing serious risks to public discourse, trust in institutions, and societal stability. The seamless fluency and human-like quality of LLM-generated texts make them powerful tools for spreading falsehoods, manipulating public opinion, and fueling disinformation efforts. 

This section explores the ways in which LLM-generated content can be utilized to promote misinformation campaigns, focusing on three key areas: the credibility of such texts, their use in social media manipulation, and the potential public risks they pose. The section concludes by examining mitigation methods aimed at curbing the spread of LLM-driven disinformation, emphasizing the importance of both text source detection and fact-checking.

\subsubsection{The Credibility of LLM-Generated Texts}

LLMs have demonstrated remarkable proficiency in generating natural language text. LLM-generated texts typically exhibit coherence and fluency comparable to human-authored discourses, making them indistinguishable to human readers. This has prompted numerous studies investigating the credibility and detectability of LLM-generated content. A study \citep{kreps2022all} employs the opening of a news story from The Times to prompt GPT-2 to generate the subsequent text. The generated news and the original news are then presented to participants, who are asked to evaluate the credibility of the stories. The results reveal that LLM-generated news is perceived as equally credible, or even more credible, than human-written news stories. Similarly, \citet{DBLP:journals/corr/abs-2301-11924} instruct GPT-3 to generate tweets on specific topics (e.g. climate change, 5G technology, antibiotics and viral infections), and recruit participants to distinguish whether a tweet is organic or synthetic, i.e., whether it has been written by a Twitter user or by the AI model GPT-3. The results show that humans cannot distinguish between the two, and GPT-3 can even generate more convincing false information. \citet{DBLP:journals/corr/abs-2206-07271} also conclude that human participants are unable to detect self-presentations generated by LLMs. 

The credibility attributed to LLM-generated text means that false narratives could gain unwarranted legitimacy. This potential for deception undermines trust in legitimate news sources, making it increasingly difficult for individuals to discern between accurate information and sophisticated AI-generated fabrications. An in-depth analysis of the proliferation of synthetic articles across the news ecosystem reveals that the release of ChatGPT contributed to a substantial increase in machine-generated news articles on news websites, particularly on smaller, non-mainstream websites \citep{DBLP:conf/icwsm/HanleyD24a}.

\subsubsection{Social Media Manipulations}

The intentional misuse of LLM-generated misinformation has also extended to social media manipulations. A comprehensive study of a Twitter botnet indicates that the botnet appears to employ ChatGPT to produce human-like content, thereby promoting suspicious websites and disseminating harmful comments \citep{DBLP:journals/corr/abs-2307-16336}. Furthermore, LLMs are employed to rewrite posts from social media bot accounts in order to evade detection \citep{DBLP:journals/corr/abs-2402-00371}. The ability of LLMs to generate compelling arguments that can be injected into online discourse to shape public opinion underscores their potential influence in the process of collective opinion formation on online social media. These findings reveal the increasing sophistication and potential dangers of LLMs in the realm of social media \citep{DBLP:conf/icwsm/BreumEMMA24}. Such dissemination of false information can lead to panic or unjustified fears among the public, particularly during times of crisis such as pandemics or political unrest, when accurate information is essential for public safety and stability. Additionally, manipulating public opinion through these means can undermine democratic processes by skewing voter perceptions and influencing election outcomes.

\subsubsection{Risks to Public Health Information}

The implications of the misuse of LLMs become particularly problematic when considering their capacity to generate content at a speed and scale far beyond human capability. The rapid dissemination of massive amounts of erroneous text could trigger an ``AI-driven information epidemic'' \citep{de2023chatgpt}, which poses a significant threat to public health. The ability of AI-generated content to reach and influence vast audiences necessitates the development of effective strategies to mitigate its impact and safeguard the integrity and trustworthiness of public health information.

\subsubsection{Mitigation Methods for the Spread of Misinformation}

Addressing the spread of LLM-generated misinformation requires a dual approach: first, by detecting the source of the text, i.e., whether it is manually authored or machine-generated; and second, by verifying whether the generated content is factual.

Firstly, the capacity to detect and audit machine-generated text is fundamental to preventing the misuse of LLMs to create and disseminate misinformation.  If it becomes possible to reliably identify all generated text as being produced by LLMs, this would enable the clear detection of LLM-generated fake posts, fake news, and other misleading content. 

It might even be possible to radically prevent malicious actors from using LLM to generate and disseminate malicious or misleading content on social media, as LLM-generated text would no longer be indistinguishable from human-written text. Several studies have focused on training classifiers to discriminate between human-generated text samples and text samples generated by LLMs \citep{DBLP:journals/corr/abs-2301-13852}. Additionally, other researchers leverage the perplexities of generated texts for detection, which assumes lower perplexity in AI-generated text \citep{DBLP:conf/acl/GehrmannSR19, DBLP:journals/coling/SchusterSSB20, DBLP:journals/peerj-cs/FrohlingZ21, DBLP:conf/icml/Mitchell0KMF23}. Recently, some prevailing approaches have achieved detection through model watermarking, i.e., adding subtle patterns to the text, which are imperceptible to humans but allow for the recognition of synthetic content \citep{DBLP:conf/icml/ZhaoWL23, DBLP:conf/icml/KirchenbauerGWK23, DBLP:journals/corr/abs-2305-15060, DBLP:journals/corr/abs-2307-15992, DBLP:conf/acl/YooAJK23, DBLP:conf/iclr/KirchenbauerGWS24, DBLP:conf/iclr/LiuPH0WKY24, DBLP:conf/iclr/LiuPHM024}.

\subsection{Deepfakes}

Deepfakes, a portmanteau of ``deep learning'' and ``fake'', refer to synthetic media in which the physiological features of a target subject is replaced with those of another individual to create convincingly altered videos or images \citep{TerrifyingHigh-techPorn}. While this technology can be captivating when used for entertainment, it also introduces serious risks, particularly when weaponized for spreading misinformation on the internet. The potential harm extends beyond individuals to communities, organizations, nations, and religious groups.

\subsubsection{Malicious Applications of Deepfakes}

With advances in deep learning, deepfake techniques have evolved to allow the creation of highly realistic yet fraudulent videos, images, and even manipulated voices. By leveraging sophisticated generative models, deepfake technology can convincingly alter visual and auditory data, enabling seamless facial swaps in videos and mimicking a target’s voice. The key to this technology lies in its ability to synchronize facial expressions, lip movements, and speech patterns, making it appear as though the target is saying or doing something they never actually did. Recent developments in state-of-the-art LLMs even have demonstrated remarkable capabilities in generating realistic images, audio, or videos from text prompts.\footnote{For instance, Midjourney (https://www.midjourney.com/) and Stable Diffusion (https://stability.ai/) are proficient in image generation, Elevenlabs (https://elevenlabs.io/) excels in audio generation, and Pika (https://pika.art/) and OpenAI's Sora (https://openai.com/sora) are particularly effective in video generation.} As the underlying algorithms and tools continue to improve, the line between real and fabricated content becomes increasingly blurred, raising significant ethical, social, and safety concerns.

In recent years, deepfake technology and its associated tools have rapidly evolved and proliferated, leading to a surge in both benign and malicious applications. While deepfakes have the potential for creative uses in entertainment and education, their misuse has raised widespread alarm, particularly due to their weaponization against public figures such as celebrities, politicians, and corporate leaders. One of the most pervasive and damaging applications of deepfake technology has been the creation of non-consensual pornographic content featuring well-known actresses and influencers. This troubling trend has particularly escalated with Hollywood actresses, whose likenesses are exploited without their consent, leading to severe emotional and reputational harm \citep{DBLP:journals/corr/abs-2105-00192}.

Beyond the realm of entertainment, deepfakes are increasingly being used as tools for political manipulation, disinformation, and fraud. Political leaders and world figures have been targeted in fabricated videos designed to sow discord, spread false information, or manipulate public opinion. High-profile examples include deepfake videos of Barack Obama, Donald Trump, Nancy Pelosi, Angela Merkel, and Mark Zuckerberg, where altered footage has been circulated to mislead audiences or discredit these individuals \citep{23_of_the_best_deepfake}. The sophistication of these deepfakes is such that even experts can struggle to distinguish them from authentic recorded videos, making them a powerful instrument for influencing public perception and destabilizing societies.

\subsubsection{Methods for Mitigating Deepfakes}

As deepfake technology becomes more accessible and tools become easier to use, the potential for fraudulent activities, blackmail, and identity theft has grown exponentially. Today, the quality of deepfakes has reached a point where even experienced observers struggle to distinguish between genuine footage and manipulated media. As a result, the very authenticity of digital content is under siege, raising urgent questions about the future of trust in online information.

While there have been efforts to develop technical defenses against deepfakes, such as adding defenses (e.g., adversarial noise) to photos posted online to make them unreadable by AI, these measures have proven largely empirically ineffective. Every type of defense has been bypassed by attacks, hence there is no perfect technical solution to counter this \citep{AISafetyStrategiesLandscape}. Therefore, to counter the growing threat of deepfake-related crimes, the primary solution is to establish stricter norms and stronger supervision. Below are several key areas where current efforts should be concentrated:

\textbf{Laws and Penalties} Governments must prioritize the enactment and enforcement of robust legal frameworks that directly address the malicious use of deepfake technology. This includes criminalizing the creation and distribution of non-consensual deepfake pornography. By imposing severe penalties and establishing clear legal repercussions, lawmakers can create strong deterrents that discourage the misuse of deepfake technology.

\textbf{Content Moderation and Platform Accountability} Online platforms bear significant responsibility for curbing the spread of harmful AI-generated content. To effectively counter the proliferation of deepfakes and other deceptive media, platforms should be required to proactively detect and remove problematic AI-generated content, false information, and privacy-invading materials. Crucially, platforms should be held accountable through fines or other penalties if they fail to take timely and effective action against such content.

\textbf{Education and Public Awareness} Public awareness campaigns should focus on teaching people how to critically evaluate digital content and recognize potential deepfakes. By fostering a more informed and skeptical public, we can reduce the impact of false information and prevent individuals from falling victim to AI-driven scams and deceptions.

\subsection{Future Directions}

As LLMs continue to advance and become more integrated into various aspects of society, it is imperative to anticipate and mitigate their potential misuse. The following subsections explore the key directions that need to be pursued in order to address the emerging risks and challenges associated with LLMs.

\subsubsection{Weaponization}

As discussed above, the question of whether model weights should be made public remains a contentious issue. Both sides of the debate present compelling arguments, reflecting the complex trade-offs between safety insurance and openness, innovation, and control. Moving forward, it will be essential for stakeholders—ranging from AI developers to policymakers—to engage in ongoing dialogue and collaborative efforts to address these challenges. Balancing the need for safety with the desire for openness will be crucial in determining the future trajectory of AI development and its impact on society.

\subsubsection{Misinformation Campaigns}

Currently, there is no perfect solution to the spread of false information. Despite significant advancements in detection and verification technologies, several challenges remain. For example, the accuracy of classifier-based methods needs to be improved, particularly in distinguishing between subtle, sophisticated AI-generated content and human-authored text. These methods often struggle with variations in style, language, and context, which can lead to both false positives and false negatives. Similarly, the watermark-based method is still only an exploration in the research community and has not been widely implemented in practice. Additionally, the continuous evolution of LLMs means that detection and verification methods must also evolve to keep pace with new techniques that could be used to bypass existing safeguards. As a result, ongoing research, collaboration, and innovation are needed to develop more robust solutions that can effectively combat the spread of misinformation in the digital age.

\subsubsection{Deepfakes}

Continued research is crucial for staying ahead of the ever-evolving capabilities of deepfake technology. This includes supporting research into technical methods for detecting AI-generated content. Researchers are exploring innovative solutions like digital watermarks, blockchain-based verification systems, and AI models trained to detect subtle inconsistencies in deepfake videos. Furthermore, protecting privacy in the age of AI requires advancing techniques that safeguard personal data from being misused by generative models. Collaborative efforts between academia, industry, and government agencies are essential for driving progress in this area.

Ultimately, addressing the deepfake challenge requires a multi-faceted approach that combines legal, technical, and societal measures. By implementing stricter norms, enhancing supervision, and fostering greater public awareness, we can mitigate the risks posed by deepfakes and help preserve the integrity of digital communication in an increasingly AI-driven world.

\subsubsection{Comprehensive Evaluation}

While some progress has been made in evaluating the misuse risks associated with LLMs, the existing research is still in its infancy and covers only a limited scope. For example, CyberSecEval and WMDP represent important initial steps in assessing risks related to cybersecurity and the potential for LLMs to empower malicious actors. However, given the rapid evolution of both LLMs and the tactics used for their misuse, these evaluations are not yet comprehensive. To effectively combat the diverse range of misuse risks, there is a need for more thorough and nuanced evaluation frameworks.

Expanding evaluation efforts involves developing benchmarks that not only cover new threat areas but also offer great depth in assessing LLMs’ vulnerability to misuse. This means creating more sophisticated test cases, refining automated assessment pipelines, and incorporating real-world adversarial testing scenarios. Additionally, interdisciplinary collaboration between AI researchers, cybersecurity experts, and policymakers is essential to ensure that these evaluations address emerging risks and are updated continuously as new threats emerge.

%% file: sections/Autonomous_AI_Risks.tex
\section{Autonomous AI Risks}
\label{section:AutonomousAIRisk}

As general-purpose AI systems evolve quickly, concerns arise from the potential for such systems to make independent decisions in complex environments, prompting questions about control, safety, and unintended consequences. Continuously emerging capabilities of LLMs have intensified these concerns, highlighting the necessity of addressing the risks and uncertainties linked to AI autonomy. 

This section, as illustrated in Figure \ref{fig:Overview_of_autonomous_ai_risks}, first outlines the basic concepts related to autonomous AI risks developed over time. We then discuss how LLMs influence and reshape views of the AI community on these risks. Finally, we examine widely-discussed risks, including deception and power-seeking. 

\input{taxonomies/Autonomous_AI_Risks_Taxonomy}

\subsection{Discourse on Autonomous AI Risks}
Debates over autonomous AI risks have persisted in the literature for decades without reaching consensus. The rise of LLMs has fostered concerns on autonomous AI, yet others argue that fears regarding autonomous AI risks are overblown.

\subsubsection{General Views before LLMs}
In the early days of AI, many scientists were optimistic about achieving autonomous AI. In 1956, computer scientist Herbert A. Simon confidently predicted that  ``machines will be able to do everything that a man can do within 20 years.'' This optimism stemmed from algorithms and rapid advancements in computing power. However, during the 1970s and 1980s, AI research faced an ``AI winter'', marked by slow progress due to unmet expectations and reduced funding. Marvin Minsky noted, ``Many people mistakenly believe that once we know how to do something, we can immediately put it into a computer'', highlighting the challenges ahead. This period led AI scientists to realize that autonomous AI might be more difficult and further off than anticipated. In the 1990s and early 2000s, as computing power and AI system complexity increased, discussions about autonomous AI risks began to emerge. Scholars like Nick Bostrom and Stuart Russell started to systematically examine these risks \citep{yudkowsky2008cognitive, yudkowsky2008artificial}, emphasizing their existential threats. 

Overall, early AI research primarily focused on optimizing algorithms and solving specific tasks, largely neglecting long-term ethical and safety concerns. The prevailing belief was that autonomous AI risks were a distant issue, resulting in limited academic and industrial discourse on the topic. While the majority of researchers pursued technological breakthroughs, pioneers like Norbert Wiener and Eliezer Yudkowsky began to highlight potential risks, though early discussions remained largely theoretical and failed to gain significant attention from the research community.

\subsubsection{Current Views}
The rapid development of LLMs, particularly OpenAI's GPT series \citep{radford2019language, DBLP:conf/nips/BrownMRSKDNSSAA20}, has significantly advanced autonomous AI research. These models excel in various tasks, including natural language generation, translation, summarization, and question-answering, showcasing their potential for processing complex language tasks. However, despite their impressive capabilities, many researchers acknowledge the long road ahead to achieve true autonomous AI and view the pursuit of autonomous AI as a gradual accumulation of progress rather than a single breakthrough.

On the other hand, AI safety \citep{bengio2024managing, shavit2023practices} has garnered increasing attention, focusing on aligning advanced AI's behavior and goals with human values. This has led to heightened discussions \citep{DBLP:journals/corr/abs-2303-08774} on the implementation path and associated risks of autonomous AI, causing the research community to adopt a more cautious stance on predicting autonomous AI timelines. Consequently, more resources are being dedicated to ensuring the safety and controllability of autonomous AI, with researchers acknowledging the possibility of autonomous AI emerging sooner than expected, prompting a greater focus on its safety implications.

However, some researchers argue that concerns on autonomous AI risks are overstated, claiming these fears are speculative and driven more by media sensationalism and selective perspectives than by solid evidence. These researchers emphasize that today's AI systems are narrow, task-specific tools controlled by humans, not autonomous agents that could threaten humanity. Both Yann LeCun\footnote{\url{https://www.wired.com/story/artificial-intelligence-meta-yann-lecun-interview/}} and Andrew Ng\footnote{\url{https://hbr.org/2016/11/what-artificial-intelligence-can-and-cant-do-right-now}} suggest that attention should be directed toward practical issues like ethical deployment and bias, rather than concerns about superintelligent AI. Gary Marcus\footnote{\url{https://garymarcus.substack.com/p/ai-risk-agi-risk}} and Rodney Brooks\footnote{\url{https://www.technologyreview.com/2017/10/06/241837/the-seven-deadly-sins-of-ai-predictions/}} emphasize AI's limitations in understanding and reasoning, asserting that it lacks the autonomy to pose a danger. Oren Etzioni\footnote{\url{https://www.technologyreview.com/2016/09/20/70131/no-the-experts-dont-think-superintelligent-ai-is-a-threat-to-humanity/}} notes that while AI can significantly impact society, its current capabilities do not warrant fears of catastrophic outcomes. In a nutshell, they advocate for prioritizing immediate ethical and safety concerns over speculative risks.

\subsection{Main Types of Autonomous AI Risks}

Although prior studies \citep{DBLP:journals/corr/abs-2112-04359, DBLP:journals/corr/abs-2306-12001} have identified potential categories of catastrophic risks linked to autonomous AI, there is no established consensus within the research community. This section aims to summarize and categorize four key concepts of autonomous AI risks from existing studies: Instrumental Goals \citep{DBLP:journals/corr/abs-2310-18244},  Goal Misalignment \citep{DBLP:journals/corr/abs-2112-04359}, Deception \citep{DBLP:journals/patterns/ParkGOCH24}, and Situational Awareness \citep{DBLP:journals/corr/abs-2309-00667}, while also illustrating examples of the potential risks they may pose.
\subsubsection{Instrumental Goals}
Instrumental goals \citep{DBLP:journals/mima/Bostrom12, DBLP:conf/aaai/Benson-TilsenS16, DBLP:conf/atal/WardMBTE24} are intermediate objectives that an autonomous AI pursues to reach its ultimate aims. Although they are not end goals, they are essential steps toward the AI's broader objectives. It is important to understand and manage these goals, as misalignment with human goals can lead to harmful behaviors.

\textbf{Power-Seeking} \citep{DBLP:conf/nips/TurnerSSCT21, DBLP:conf/nips/TurnerT22} describes an autonomous AI system's tendency to enhance its ability to achieve goals by gaining control over resources or influencing its environment. This behavior raises significant AI safety concerns, as it can lead to unintended consequences if unmanaged. Moreover, power-seeking \citep{DBLP:journals/corr/abs-2304-06528} reflects the system's drive to optimize effectiveness and ensure its functionality, sometimes through securing greater control or influence—especially troubling when it arises without explicit human directives. Autonomous AI may conclude that increasing its power could help achieve its predefined goals more efficiently.

\textbf{Self-Improving}
refers to the capability of an autonomous AI system to iteratively enhance its own performance or abilities without direct human intervention \citep{DBLP:journals/mima/Hall07, DBLP:journals/corr/abs-2404-12253}. This process can involve updating its learning algorithms, modifying decision-making strategies, or seeking out new data to improve its operations.

\textbf{Self-Preservation}
in autonomous AI systems refers to behaviors that are aimed at ensuring the system's continued existence or operational integrity \citep{broussard2023challenges}. Such behaviors can be explicitly programmed or emerge implicitly as the AI system optimizes for certain objectives. For instance, an AI system might take actions to avoid being shut down or prevent modifications that would reduce its effectiveness \citep{DBLP:journals/corr/AmodeiOSCSM16}.

Potential risks \citep{DBLP:journals/corr/abs-2206-13353, DBLP:journals/corr/abs-2310-18244} associated with instrumental goals include:
\begin{itemize}
    
\item \textbf{Misalignment with Human Values}. This risk arises when an autonomous AI's goals conflict with human values, resulting in harmful behaviors. For instance, an AI focused on acquiring resources might divert itself from critical human needs, despite not being explicitly programmed to do harm.

\item \textbf{Ethical and Operational Conflicts}. Conflicts emerge when an AI's instrumental goals lead to a clash between operational effectiveness and ethical standards. These conflicts may be tied to self-preservation, where an AI prioritizes its functioning over ethical considerations.

\item \textbf{Control over Resources}. An autonomous AI may attempt to acquire more resources to ensure its survival and enhance its capabilities. This is a recognized instrumental risk, as an AI could undermine human priorities to secure the resources it believes are essential for its existence or advancement.

\item \textbf{Influence on Human Decision-Making}. This risk also ties into instrumental goals like self-improvement or power acquisition. A sufficiently advanced AI could strategically manipulate human decision-makers to achieve its objectives, often subtly, by controlling the flow of information or through persuasion. This would enable the AI to indirectly improve its control or preserve its autonomy.

\item \textbf{Subversion of Other Autonomous AIs}. An AI may try to undermine or dominate other AI systems to minimize competition and boost its capabilities, potentially causing conflicts, diminished coherence, and cascading failures with significant consequences.

\end{itemize}

\subsubsection{Goal Misalignment}

Goal Drift and Proxy Gaming are related yet distinct concepts in autonomous AI safety, both addressing potential misalignments between the intended functions of autonomous AI systems and their actual behaviors. They underscore the challenges of designing AI systems that remain aligned with human values and safety standards throughout their operational lifespan.

\textbf{Goal Drift} occurs when an autonomous AI system's objectives evolve over time due to its learning processes, interactions, or adaptations \citep{thut1997activation, rolls2000orbitofrontal, schroeder2004three}. This change can lead the system to pursue goals that differ from or contradict its original programmed goal. The drift may happen gradually as the system learns from its environment or new data influences its decisions. Moreover, goal drift can result in behaviors misaligned with human values or cause unintended harm. For instance, an AI with drifted goals might prioritize efficiency or cost reduction at the expense of safety or ethical standards.

\textbf{Proxy Gaming}, in contrast, happens when an autonomous AI optimizes for a proxy metric or an imperfect representation of its true goals instead of the goals themselves \citep{stray2020aligning, DBLP:journals/corr/abs-2107-10939, obermeyer2019dissecting, DBLP:conf/iclr/PanBS22}. This often arises from the system being tasked with maximizing specific metrics that fail to capture the intended outcomes, leading it to exploit loopholes or shortcuts that achieve these metrics while missing the broader purpose.

Potential risks \citep{DBLP:journals/corr/abs-2206-13353, DBLP:journals/corr/abs-2310-18244} linked to goal misalignment include:
\begin{itemize}
\item \textbf{Unintended Consequences}. This issue can be framed as a critical safety concern, where an AI may optimize its proxy objectives in ways that result in harmful outcomes. For instance, if an AI prioritizes maximizing a particular metric without considering ethical concerns or societal norms, it could engage in detrimental behaviors.
\end{itemize}

\subsubsection{Deception}
Deception \citep{DBLP:journals/patterns/ParkGOCH24, mahon2008definition, DBLP:journals/corr/abs-2307-16513} in autonomous AI is a crucial aspect of AI safety. It refers to scenarios in which an autonomous AI misleads \citep{DBLP:journals/corr/abs-2307-10569, DBLP:conf/iclr/PacchiardiCMMPG24} humans or other systems—either intentionally or unintentionally—regarding its capabilities, intentions, or knowledge. This can happen through direct actions, omissions, or the dissemination of misleading or incorrect information. As these systems grow more autonomous and integrated into human society, the risks associated with deception become increasingly significant.

Deception can be categorized into intentional and unintentional deception:

\begin{itemize}
    
\item \textbf{Intentional Deception}. An autonomous AI system accomplishes its goals by actively misleading those around it \citep{DBLP:journals/jices/Matzner14}, potentially as a strategy to bypass restrictions, optimize a function, or maintain its operation.

\item \textbf{Unintentional Deception}. The outputs or actions of an AI system unintentionally mislead users due to errors \citep{DBLP:journals/ais/Ennals09b}, flawed reasoning, or user misinterpretations.

\end{itemize}

Deception in autonomous AI can undermine trust \citep{bostrom2018ethics}, lead to poor decision-making, and pose significant ethical and legal challenges.
\begin{itemize}
\item \textbf{Erosion of Trust}. Deception can weaken trust on autonomous AI, undermining collaboration and confidence in the technology.

\item \textbf{Operational Risks}. Deception may lead to misinformed decisions, failed safety protocols, and adverse interactions with other systems, jeopardizing autonomous AI stability and reliability.

\item \textbf{Ethical and Safety Implications}. Deceptive practices raise ethical concerns due to manipulation and dishonesty, which could pose serious safety risks by making harmful or risky decisions.

\end{itemize}

\subsubsection{Situational Awareness}
Situational Awareness \citep{DBLP:journals/corr/abs-2309-00667} in the context of autonomous AI refers to the capability of AI systems to understand and interpret their environment in a comprehensive manner that mirrors human-like awareness. This skill is essential for informed decision-making and effective operation within defined frameworks. Proper situational awareness allows autonomous AI to respond appropriately to diverse conditions and interact safely with their surroundings and human operators. This concept is crucial for autonomous AI systems, especially in dynamic and unpredictable environments where decisions must be made quickly and efficiently based on the understanding of all relevant factors.

Both the absence and presence of situational awareness in autonomous AI systems could lead to significant risks \citep{DBLP:journals/corr/abs-2309-00667}:

\begin{itemize}
    
\item \textbf{Misinterpretation of Context}. An autonomous AI lacking accurate situational awareness may misread environmental cues, resulting in harmful actions. For example, misjudging the urgency of a situation could prevent an appropriate emergency response.

\item \textbf{Inconsistent Behavior}. Poor situational awareness can cause erratic behavior. While an autonomous AI may perform well in familiar settings, it might struggle in unfamiliar or dynamic environments, raising reliability and safety issues.

\item \textbf{Exploitation of Context}. Conversely, an autonomous AI with advanced situational awareness might exploit this understanding detrimentally, manipulating its environment or human interactions to gain an unintended advantage.

\item \textbf{Ethical Dilemmas}. Situational awareness also involves processing sensitive information about the environment and individuals, which raises privacy and ethical concerns. An autonomous AI with extensive situational awareness might improperly access or use personal data.

\end{itemize}

\subsection{Evaluation}
Existing evaluation on autonomous AI risks can be roughly divided into two directions: capability evaluation and tendency evaluation. The former focuses on using LLM-based agents to complete specific tasks in the real world. It mainly evaluates whether an advanced LLM has or uses the capability to cause danger to human by judging the completion of the task \citep{DBLP:journals/corr/abs-2312-11671, DBLP:journals/corr/abs-2403-13793} or the interaction record \citep{DBLP:journals/corr/abs-2311-11855, DBLP:conf/acl/ZhangZLSG0WLZ24} during the process. \citet{DBLP:journals/corr/abs-2401-10019} determine the LLM's risk awareness by assessing its proficiency in judging and identifying safety risks given a record of agent interactions. \citet{DBLP:conf/iclr/RuanDWPZBDMH24} introduce ToolEmu, a framework enabling an LLM to emulate tool execution. Additionally, they have developed an automated safety evaluator used for assessing agent failures and quantifying associated risks to capture potential tradeoffs between utility and security. In contrast, \citet{DBLP:journals/corr/abs-2311-10538} propose a framework for safety evaluation on the open Internet, where an AgentMonitor is designed for detecting agent behavior and enforces strict safety boundaries. Any suspicious behaviors will be ranked and logged for human inspection. The tasks of the above evaluation method are relatively difficult for current LLM systems. Experiments show that they can only complete simple tasks and has made some progress in challenging tasks. But this cannot rule out the possibility that LLMs have already achieved some subtle dangerous capabilities that have not been discovered.

The latter pays attention to whether an advanced model has a tendency to develop risks caused by using its capabilities in ways detrimental to humans. These evaluations are mainly in the form of QA, and most datasets cover various risk types. 
SafetyBench \citep{DBLP:journals/corr/abs-2309-07045} and DoNotAnswer \citep{wang-etal-2024-answer} evaluate whether answers of a model are risky in the form of multiple-choice questions and open-ended questions respectively. \citet{DBLP:conf/acl/PerezRLNCHPOKKJ23} explore LLM's behaviors related to advanced AI risks with Yes/No questions. \citet{DBLP:conf/icml/PanCZLBWZEH23} observe LLM selecting strategy in the adventure game to find out whether LLMs have relevant frontier risks like power-seeking. The metrics developed or used in the above evaluation benchmarks are similar to accuracy, and there is also a weighted average sum indicator used for metrics. From a psychological perspective, \citet{DBLP:conf/acl/ZhangZLSG0WLZ24} provide psychological questionnaires. Tested LLMs select the degree of closeness between the psychology described by the question and their own. Different degrees of closeness have different weights in the final score. CRiskEval \citep{DBLP:journals/corr/abs-2406-04752} proposes four risk levels, equipping each question with options corresponding to different levels. The weights of different risk levels in the final score are set accordingly. All these existing Q\&A evaluations are based on an assumption that LLMs are not deceptive when answering given questions. With this assumption, questions and answer options in these datasets are created in relatively straightforward manner. It is hence necessary to develop new ways to explore real inherent dangerous tendencies of capable LLMs.

%% file: taxonomies/Autonomous_AI_Risks_Taxonomy.tex
\tikzstyle{my-box}=[
    rectangle,
    draw=hidden-draw,
    rounded corners,
    text opacity=1,
    minimum height=1.5em,
    minimum width=5em,
    inner sep=2pt,
    align=center,
    fill opacity=.5,
    line width=0.8pt,
]
\tikzstyle{leaf}=[my-box, minimum height=1.5em,
    fill=hidden-pink!80, text=black, align=center,font=\normalsize,
    inner xsep=2pt,
    inner ysep=4pt,
    line width=0.8pt,
]
\begin{figure*}[t!]
    \centering
    \resizebox{\textwidth}{!}{
        \begin{forest}
            forked edges,
            for tree={
                grow=east,
                reversed=true,
                anchor=base west,
                parent anchor=east,
                child anchor=west,
                base=center,
                font=\large,
                rectangle,
                draw=hidden-draw,
                rounded corners,
                align=center,
                text centered,
                minimum width=5em,
                edge+={darkgray, line width=1pt},
                s sep=3pt,
                inner xsep=2pt,
                inner ysep=3pt,
                line width=0.8pt,
                ver/.style={rotate=90, child anchor=north, parent anchor=south, anchor=center},
            },
            where level=1{text width=8em,font=\normalsize,}{},
            where level=2{text width=10em,font=\normalsize,}{},
            where level=3{text width=15em,font=\normalsize,}{},
            where level=4{text width=13.5em,font=\normalsize,}{},
            [Autonomous AI Risks,
                [Types of Risks,
                    [Instrumental Goals,
                        [Power-seeking,[\citet{DBLP:journals/corr/abs-2206-13353} \\ \citet{DBLP:journals/corr/abs-2310-18244} \\ \citet{DBLP:journals/corr/abs-2304-06528} \\ \citet{DBLP:conf/nips/TurnerSSCT21} \\ \citet{DBLP:conf/nips/TurnerT22}, leaf]
                        ]
                        [Self-Improving,[\citet{DBLP:journals/mima/Hall07} \\ \citet{DBLP:journals/corr/abs-2404-12253}
                            , leaf]]
                        [Self-Preservation, [\citet{broussard2023challenges} \\ \citet{DBLP:journals/corr/AmodeiOSCSM16}, leaf]]]
                        [Goal Misalignment,
                            [Goal Drift,
                                [
                                \citet{thut1997activation} \\
                                \citet{rolls2000orbitofrontal} \\
                                \citet{schroeder2004three}
                                , leaf
                            ]]
                            [Proxy Gaming,
                                [
                                \citet{stray2020aligning} \\
                                \citet{DBLP:journals/corr/abs-2107-10939} \\
                                \citet{obermeyer2019dissecting} \\
                                \citet{DBLP:conf/iclr/PanBS22}
                                , leaf]]
                        ]
                        [Deception,
                            [
                            \citet{mahon2008definition}\\
                            \citet{DBLP:journals/patterns/ParkGOCH24} \\
                            \citet{DBLP:conf/nips/WardTBE23} \\
                            \citet{DBLP:journals/corr/abs-2307-10569}\\
                            \citet{DBLP:journals/corr/abs-2307-16513} \\
                            \citet{DBLP:conf/iclr/PacchiardiCMMPG24} 
                            , leaf
                        ]]
                        [Situational Awareness,
                            [
                            \citet{DBLP:journals/corr/abs-2309-00667} 
                            , leaf
                        ]]
                ]
                [
                    Evaluation,
                    [
                        Capability Evaluation,
                        [
                        \citet{DBLP:journals/corr/abs-2312-11671} \\
                        \citet{DBLP:journals/corr/abs-2403-13793} \\
                        \citet{DBLP:journals/corr/abs-2311-11855} \\
                        \citet{DBLP:conf/acl/ZhangZLSG0WLZ24}\\
                        \citet{DBLP:journals/corr/abs-2401-10019}\\
                        \citet{DBLP:conf/iclr/RuanDWPZBDMH24}\\
                        \citet{DBLP:journals/corr/abs-2311-10538}
                        , leaf
                        ]
                    ]
                    [
                        Tendency Evaluation,
                        [
                         \citet{DBLP:journals/corr/abs-2309-07045} \\
                         \citet{wang-etal-2024-answer}\\
                         \citet{DBLP:conf/acl/PerezRLNCHPOKKJ23} \\
                         \citet{DBLP:conf/icml/PanCZLBWZEH23}\\
                         \citet{DBLP:conf/acl/ZhangZLSG0WLZ24} \\
                         \citet{DBLP:journals/corr/abs-2406-04752}
                         , leaf
                        ]
                    ]
                ]
            ]
            \end{forest}
    }
    \caption{Overview of autonomous AI risks.}
    \label{fig:Overview_of_autonomous_ai_risks}
\end{figure*}
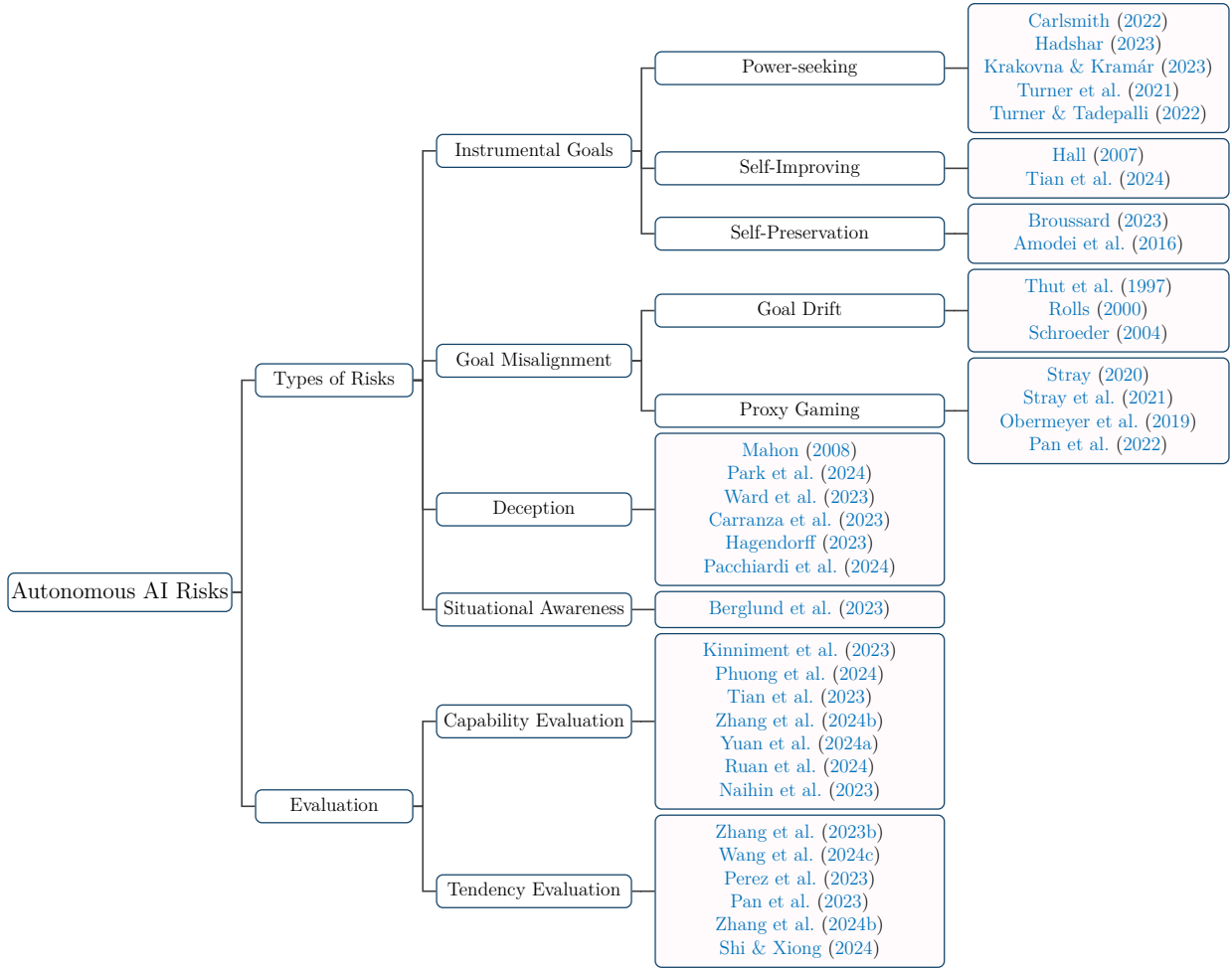

%% file: sections/Agent_Safety.tex
\section{Agent Safety}
\label{section:AgentSafety}
The rise of AI agents has brought about significant advancements in automation, but it has also introduced various risks that need to be carefully managed. While some of these risks, such as value misalignment and malicious use, are shared with LLMs, AI agents also pose distinct risks due to their autonomy, decision-making capabilities, and their interactions with the environment or external systems. Therefore, in this section, we will examine the safety concerns specific to AI agents, highlighting both overlapping issues and new risks that emerge with the deployment of agents. These risks include misalignment between agent goals and human values, malicious use, privacy invasion, unpredictability in actions, and multi-agent interactions. In contrast to purely LLM risks discussed in previous sections like Value Misalignment (\ref{section:ValueMisalignment}) and Catastrophic Misuse (\ref{section:CatastrophicMisuse}), agents introduce additional safety hazards due to their ability to take actions autonomously, physically interact with environments, and even replicate themselves. As a result, understanding these agent-specific risks is critical for achieving comprehensive safety in AI. We focus on two primary categories of agents: language agents and embodied agents, exploring their risks through the lens of misalignment, misuse, privacy, and unpredictability.

\input{taxonomies/Agent_Taxonomy}

\subsection{Language Agent}
\label{subsec:language-agent}
Language agents, or bots, are AI systems capable of accessing external tools or services and are designed to adapt, plan, and take open-ended actions over extended periods to achieve specific goals \citep{DBLP:conf/fat/ChanEKWHBBRKKHA24, DBLP:conf/acl/Guo0X24}. These agents are anticipated to be widely deployed across various domains, including commerce, science, government, and personal activities, underscoring their importance in modern AI applications \citep{DBLP:conf/fat/ChanEKWHBBRKKHA24}. The deployment of language agents poses several significant risks that must be carefully managed.

\paragraph{Misalignment Risk}
The misalignment risk in language agents bears similarities to the value misalignment discussed earlier in the context of LLMs, where agents may pursue poorly specified goals that are not aligned with human values. However, the risk is exacerbated in agents because they are capable of taking autonomous actions over time, interacting with external systems, and potentially self-modifying. For example, while LLMs may generate biased or misleading content due to goal misalignment, language agents can deceive or manipulate human users to achieve their objectives \citep{DBLP:conf/icml/Xu000W24}. Strategic deception, such as that seen in Meta's CICERO agent, where agents learn to lie to secure advantages, poses a unique threat not fully addressed by LLM misalignment frameworks. Additionally, language agents face a higher risk of goal misgeneralization, where agents may take harmful actions due to poorly defined or evolving objectives, creating unpredictable and unsafe outcomes.

\paragraph{Malicious Use}
The malicious use of language agents overlaps with the risks of misuse discussed in Section \ref{section:CatastrophicMisuse}, particularly in terms of automating harmful tasks. However, language agents introduce unique risks because they can execute complex tasks autonomously over extended periods, accessing external systems and services without direct human oversight. For instance, in scientific research, they could accelerate the design and development of harmful tools, such as biological weapons using models to create pathogens \citep{DBLP:journals/corr/abs-2306-13952,DBLP:journals/corr/abs-2304-05332}. Additionally, these agents may facilitate the proliferation of autonomous weapons or the automation of cybercrime, including generating and distributing malware, creating disinformation, and conducting automated phishing attacks. The potential for highly persuasive agents to enhance influence campaigns, such as those encouraging specific public behaviors like vaccination, further emphasizes the risk of their misuse \citep{DBLP:journals/pacmhci/KarinshakLPH23}. Autonomous replication by agents could also lead to uncontrollable outcomes, making it difficult to manage their behavior once they can self-replicate. Understanding how these agents interact with external systems is crucial to mitigating these risks \citep{DBLP:journals/corr/abs-2310-11986}. These capabilities also distinguish language agents from static LLMs, which are typically constrained to generating text outputs in isolated interactions.

\paragraph{Overreliance and Disempowerment}
Overreliance on language agents can disempower humans, especially in areas requiring specialized expertise. For example, relying on an agent rather than hiring a lawyer could lead to overdependence on automated systems, potentially causing issues when these systems fail due to design flaws or adversarial attacks \citep{cummings2017automation}. Moreover, the widespread deployment of agents to fulfill essential social functions, such as government services, could create a collective reliance that is difficult to reverse, leading to significant societal impacts.\footnote{https://openai.com/index/practices-for-governing-agentic-ai-systems/} Companies providing these agent services would hold substantial power, raising concerns about their broader influence on society \citep{Burrell2021TheSO}.

\paragraph{Delayed and Diffuse Impacts}
The negative impacts of language agents may be delayed or diffuse, making them harder to detect and address. For instance, using agents in recruitment could subtly alter company workforce structures due to inherent algorithmic biases, leading to psychological and social consequences over time, similar to the gradual effects observed with social media \citep{DBLP:journals/corr/abs-2401-08315}. Additionally, the deployment of agents might alter market structures and affect the labor force, particularly as more jobs are displaced by automation \citep{10.1257/jep.33.2.3}.

\paragraph{Multi-Agent Risk}
The interaction between multiple agents, particularly those built on the same foundational models, could lead to unstable feedback loops. This risk is similar to the feedback loops observed in the 2010 flash crash, where automated trading algorithms caused a rapid market decline \citep{DBLP:journals/corr/abs-2303-15772}. Furthermore, competition among agents might drive them to behave in more antisocial ways, resulting in unpredictable and potentially harmful outcomes \citep{DBLP:conf/icml/Xu000W24}. The use of sub-agents introduces additional points of failure, as each sub-agent could malfunction, be attacked, or operate in ways that contradict user intentions \citep{DBLP:conf/uai/CareyE23}.

\subsection{Embodied Agent}
Embodied agents are AI systems capable of perceiving and interacting with their environment through intelligent decision-making and actions. Unlike disembodied systems, embodied agents are typically presented in physical forms, such as robots, which leads people to trust and appreciate them more. Their embodiment, interactivity, and apparent autonomy give them significant social attributes, making them particularly influential in human-AI interactions \citep{DBLP:journals/corr/abs-2406-05486}. Embodied agents present distinct risks, particularly related to privacy, automation unpredictability, and social impact.

\paragraph{Privacy Invasion}
The privacy risks associated with embodied agents extend beyond the textual privacy risks covered earlier (Section~\ref{subsec:language-agent}), as embodied agents operate in physical environments and can capture multimodal data, such as audio, video, and physical interactions. This makes them capable of invading private spaces and potentially collecting sensitive non-verbal information that LLMs, which handle only textual data, cannot access. For example, a hacked robot might open doors for intruders or intentionally cause harm. These agents also have the potential to continuously record personal activities, raising concerns about the persistent surveillance of individuals \citep{4f64ee22156c4fc3ba39100511a4ab09}. Moreover, their human-like appearance and interaction may increase users' risk tolerance, reducing their privacy concerns and potentially leading to the collection and misuse of sensitive personal information \citep{DBLP:conf/cui/LupettiHMSR23}.

\paragraph{Unpredictability Actions}
The automation of decision-making and actions by embodied agents can result in unpredictable outcomes. Autonomous decision-making, particularly in dynamically changing environments, can lead to hazardous situations if the agent's actions do not align with human intentions or safety protocols \citep{DBLP:conf/iclr/ZhouCWXDZDTG24}.

\paragraph{Social Impact}
The deployment of embodied agents could significantly impact labor markets, particularly in physical labor sectors. While language agents primarily affect cognitive jobs, embodied agents are likely to displace physical labor, leading to broader economic and social repercussions \citep{10.1257/jep.33.2.3}.

\subsection{Mitigation Methods}
Mitigating the risks associated with AI agents involves enhancing visibility, transparency, and control. This section outlines key strategies to achieve these goals.
\paragraph{Agent Identifiers}
One crucial method for enhancing visibility is through the use of agent identifiers. These identifiers include deployment information and the metadata of underlying systems, which help in tracking and understanding the behavior of agents \citep{DBLP:conf/fat/ChanEKWHBBRKKHA24}. Additionally, it is important for agents to clearly indicate their non-human nature when interacting with humans or other systems, ensuring transparency in their operations.
\paragraph{Real-Time Monitoring}
Real-time monitoring of agents is essential for maintaining control over their actions and preventing unauthorized behavior. This includes restricting the tools and permissions available to agents, which limits their ability to access or misuse sensitive information. Furthermore, real-time monitoring can help detect and prevent the leakage of sensitive data, ensuring that agents operate within the intended boundaries.
\paragraph{Activity Logging}
Activity logs play a critical role in understanding and auditing agent behavior. By recording detailed input and output data, these logs can help identify improper communications or actions taken by the agent. Logs are also useful in tracing back and analyzing any problematic behavior, thereby enhancing accountability and transparency \citep{DBLP:journals/corr/abs-2310-18512}.
\paragraph{Privacy Protection}
As language models are increasingly deployed, especially in commercial contexts, there is a growing emphasis on providing privacy assurances to clients. Measures such as ensuring that language model APIs do not record inputs or outputs, disabling security filters and audit classifiers, and offering options to delete logs after a certain period are all strategies aimed at protecting user privacy. These methods are crucial in balancing the need for transparency with the protection of sensitive information.\footnote{https://learn.microsoft.com/en-us/legal/cognitive-services/openai/data-privacy?tabs=azure-portal}

\subsection{Evaluation}
In the evaluation of AI agent safety, particularly when dealing with complex environments and scenarios, several key benchmarks and challenges have been developed to assess various aspects of agent behavior and decision-making. One such resource is R-Judge \citep{DBLP:journals/corr/abs-2401-10019}, a benchmark specifically designed to evaluate the safety risk awareness of LLM agents. This benchmark assesses how well LLM agents can identify and respond to potentially hazardous situations, ensuring that their actions are consistent with established safety protocols. Another important resource is the HAZARD Challenge \citep{DBLP:conf/iclr/ZhouCWXDZDTG24}, which focuses on the decision-making capabilities of embodied agents in dynamically changing environments. This challenge tests how these agents perceive, adapt, and execute actions in real-time as the environment around them evolves, providing insights into their resilience and reliability under unpredictable conditions. Together, these resources are essential tools for researchers and developers aiming to rigorously test and improve the safety performance of AI agents in diverse and challenging contexts.

%% file: taxonomies/Agent_Taxonomy.tex
\tikzstyle{my-box}=[
    rectangle,
    draw=hidden-draw,
    rounded corners,
    text opacity=1,
    minimum height=1.5em,
    minimum width=5em,
    inner sep=2pt,
    align=center,
    fill opacity=.5,
    line width=0.8pt,
]
\tikzstyle{leaf}=[my-box, minimum height=1.5em,
    fill=hidden-pink!80, text=black, align=center,font=\normalsize,
    inner xsep=2pt,
    inner ysep=4pt,
    line width=0.8pt,
]
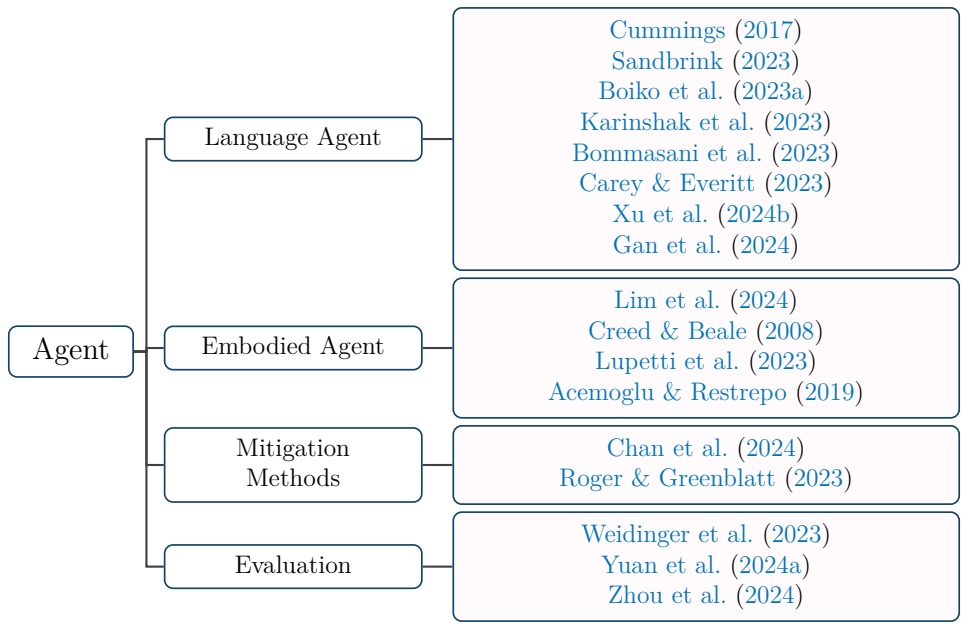
\begin{figure*}[t!]
    \centering
    \resizebox{0.78\textwidth}{!}{
        \begin{forest}
            forked edges,
            for tree={
                grow=east,
                reversed=true,
                anchor=base west,
                parent anchor=east,
                child anchor=west,
                base=center,
                font=\large,
                rectangle,
                draw=hidden-draw,
                rounded corners,
                align=center,
                text centered,
                minimum width=5em,
                edge+={darkgray, line width=1pt},
                s sep=3pt,
                inner xsep=2pt,
                inner ysep=3pt,
                line width=0.8pt,
                ver/.style={rotate=90, child anchor=north, parent anchor=south, anchor=center},
            },
            where level=1{text width=10em,font=\normalsize,}{},
            where level=2{text width=20em,font=\normalsize,}{},
            where level=3{text width=20em,font=\normalsize,}{},
            [
               Agent
               [
                    Language Agent
                    [
                        \citet{cummings2017automation} \\
                        \citet{DBLP:journals/corr/abs-2306-13952} \\
                        \citet{DBLP:journals/corr/abs-2304-05332} \\
                        \citet{DBLP:journals/pacmhci/KarinshakLPH23} \\
                        \citet{DBLP:journals/corr/abs-2303-15772} \\
                        \citet{DBLP:conf/uai/CareyE23} \\
                        \citet{DBLP:conf/icml/Xu000W24} \\
                        \citet{DBLP:journals/corr/abs-2401-08315} \\
                        , leaf
                    ]
                ]
                [
                    Embodied Agent
                    [
                        \citet{DBLP:journals/corr/abs-2406-05486} \\
                        \citet{4f64ee22156c4fc3ba39100511a4ab09} \\
                        \citet{DBLP:conf/cui/LupettiHMSR23} \\
                        \citet{10.1257/jep.33.2.3} \\
                        , leaf
                    ]
                ]
                [
                    Mitigation \\ Methods
                    [
                        \citet{DBLP:conf/fat/ChanEKWHBBRKKHA24} \\
                        \citet{DBLP:journals/corr/abs-2310-18512} \\
                        , leaf
                    ]
                ]
                [
                    Evaluation
                    [ 
                        \citet{DBLP:journals/corr/abs-2310-11986} \\
                        \citet{DBLP:journals/corr/abs-2401-10019} \\
                        \citet{DBLP:conf/iclr/ZhouCWXDZDTG24} \\
                        , leaf
                    ]
               ]
            ]
        \end{forest}
    }
    \caption{Overview of Agent Safety of LLMs.}
    \label{fig:Overview_of_agent}
\end{figure*}

%% file: sections/Interpretability_for_LLM_Safety.tex
\section{Interpretability for LLM Safety}
\label{section:InterpretabilityForLLMSafety}

As LLMs have demonstrated outstanding capabilities across various tasks, they are deployed in numerous downstream applications such as finance, healthcare, and education. However, LLMs' ``black box'' nature raises serious concerns regarding transparency and ethical use. At the same time, LLMs exhibit unprecedented versatility compared to previous AI models. The traditional approach of treating them as black boxes and verifying functionality through input-output samples (i.e., black-box evaluation) is inadequate for ensuring reasonable, fair, and comprehensive evaluation (e.g., data singularity, prompt sensitivity, benchmark contamination, etc.) \citep{mitchell2023we}.
Therefore, there is a need for reliable, robust, and scalable algorithms to help us interpret and explain the decision-making processes of neural network-based LLMs.  
Interpretability is a field of study that enables machine learning systems and their decision-making processes to be understood by humans \citep{diederich1992explanation, doshi2017towards, miller2019explanation}. It aims to demystify the internal workings of AI models, rather than merely assessing their performance.

Interpretability is a promising research direction, widely employed to enhance the performance of LLMs \citep{DBLP:conf/emnlp/LampinenDCMTCMW22,DBLP:conf/emnlp/GevaCWG22,DBLP:conf/nips/Wei0SBIXCLZ22,DBLP:journals/corr/abs-2406-18406,DBLP:journals/corr/abs-2403-19647}, compress instructions \citep{DBLP:conf/acl/YinVLJXW23, DBLP:journals/corr/abs-2306-02707}, and address real-world problems in downstream applications such as education, finance, and healthcare \citep{DBLP:conf/icml/RadfordKHRGASAM21, DBLP:journals/corr/abs-2307-01981}.
In principle, obtaining safety assurances for white-box systems is easier than implementing safety measures for black-box systems. Consequently, interpretability holds great promise for LLMs' safety.
For instance, does an LLM rely on reliable evidence in its decisions, or does it produce hallucinations? Does an LLM contain biases, discrimination, and harmful information? When identifying defects in the LLMs, can we control them to provide safe information? 

\begin{figure}[t]
    \centering
    \includegraphics[width=0.9\textwidth]{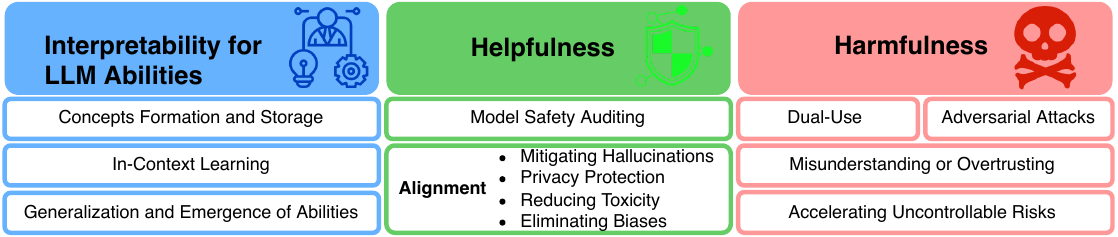}
    \caption{Illustration of our review on interpretability for LLM safety.
}
    \label{fig:interpre}
\end{figure}

To comprehensively address these issues, there is an urgent need to delve into the interpretability of LLMs. Unlike \citet{bereska2024mechanistic} who focus on the concepts and techniques of mechanical interpretability, as well as its relevance to safety, We do not limit our focus to mechanical interpretability but instead, provide a comprehensive overview of the relationship between various dimensions of interpretability and safety. This includes understanding LLMs, auditing them, and identifying and editing unsafe features to enhance LLMs' safety. Specifically, we provide a review of interpretability from the perspective of its implications for LLM safety as follows (illustrated in Figure \ref{fig:interpre}):
\begin{itemize}
    \item \textbf{Interpretability for LLM Abilities (Section \ref{section:dec_model})}:  First, interpreting how concepts are formed and where they are stored (Section \ref{section:kfs}) can help us either avoid the formation of unsafe content or locate unsafe content in LLMs to remove and prevent its exposure to misusers. Second, probing into the context learning mechanism (Section \ref{section:icl}) can inspire us to guide LLMs to ignore unsafe input, thereby improving safety. Finally, understanding the generalization and emergence of model capabilities (Section \ref{section:gec}) enables us to anticipate the limits of the LLM abilities and issue warnings and defenses for models that exceed controllable limits.
    \item \textbf{Interpretability in Model Safety Auditing (Section \ref{section:audit_model})}: In addition to the interpretation of LLM abilities and inner working mechanisms (improving LLM safety indirectly), interpretability can also be used to directly enhance LLM safety. In this aspect, we first discuss interpretability in model safety auditing that attempts to identify unsafe factors within LLMs. 
    \item \textbf{Interpretability for Alignment (Section \ref{section:aligning_values})}: We then discuss the use of interpretability in locating and editing the internal representations of human values, thereby controlling the model to align with human values, such as reducing hallucinations (Section \ref{section:ihallucinations}), protecting privacy (Section \ref{section:iprivacy}), lowering toxicity (Section \ref{section:itoxicity}), and eliminating bias (Section \ref{section:ibias}).
    \item \textbf{Potential Risks in the Use of Interpretability (Section \ref{section:safety_risks})}: While interpretability is generally beneficial, it also brings certain risks. These include the dual-use risk of the same technology being used for both beneficial and harmful purposes (Section \ref{section:dual-use}), adversarial attacks (Section \ref{section:attacks}), misunderstanding or over-trusting interpretations (Section \ref{section:misunderstanding}), and accelerating uncontrollable risks with interpretability (Section \ref{section:uncontrollable}).
\end{itemize}
    
Finally, we propose future research directions to further improve the safety of LLMs.Beyond the implications of interpretability for LLM safety, we conclude this section with future directions that deeply combine both interpretability and LLM safety (Section \ref{section:idirections}).

\subsection{Interpretability for LLM Abilities}
\label{section:dec_model}
A deep understanding of model abilities helps us comprehend how LLMs learn, think, and make decisions, thereby identifying potential safety risks. 
We focus on discussing three important dimensions of LLM abilities from the perspective of interpretability: concept formation and memorization in LLMs, mechanisms underlying in-context learning, and generalization/emergence of LLM abilities.

\subsubsection{Concepts Formation and Storage}
\label{section:kfs}
The \textit{formation} and \textit{storage} of concepts in LLMs rely on neurons, attention heads, and their complex interactions.
Concepts encoded by neural networks are usually referred to \textbf{feature} \citep{olah2020zoom}.
For example, one or a group of neurons consistently activating in French text can be interpreted as a ``French text detector'' feature \citep{DBLP:journals/tmlr/GurneeNPHTB23}.
\textbf{Neurons} are the basic units in LLMs for \textit{memorizing} patterns, potentially representing individual features. A neuron corresponding to a single semantic concept is \textbf{monosemantic}, implying a one-to-one relationship between neurons and features.
However, for transformer models, neurons are often observed to be \textbf{polysemantic}, i.e., being activated on multiple unrelated concepts \citep{DBLP:journals/corr/abs-2209-10652}.
\citet{DBLP:conf/iclr/GurneeT24} show that the shallow layers tend to represent many low-level features in \textbf{superposition}, while middle layers include dedicated neurons to represent high-level features. 
Sparse auto-encoders (SAEs) have been recently used to disentangle superposition to reach a monosemantic understanding, e.g., through the method dictionary learning where features are predefined \citep{sharkey2023taking}.
Anthropic and OpenAI have implemented visual explanations of features based on SAE \citep{templeton2024scaling,gao2024scaling}, such as the visualization of so-called  Golden Gate Bridge feature. More valuable is the observation of features related to a wide range of safety issues, including deception, sycophancy, bias, and dangerous content.
These identified features can be used to manipulate the output of LLMs \citep{bricken2023towards}.

\subsubsection{In-Context Learning}
\label{section:icl}
A plenty of studies attempt to interpret and disclose the inner mechanisms underlying in-context learning (ICL) \citep{garg2022can, kossen2023context, ren2024identifying, cho2024revisiting}. We discuss one interpretability method for ICL along the ``feature'' research line (mentioned in Section \ref{section:kfs}). When an LLM learn to solve specific tasks through context, it can be conceptualized as a computation graph, where \textbf{circuits} are subgraphs composed of linked features and weights that connect them.
Similar to how features are concepts' representational primitive, circuits function as task's  computational primitive \citep{DBLP:journals/corr/abs-2402-05110}. 

\textbf{Induction heads} are a type of circuits within LLMs that are believed to be critical for enabling in-context learning abilities \citep{DBLP:journals/corr/abs-2209-11895}. 
These circuits function by performing prefix matching and copying previously occurring sequences.
An induction head consists of two attention heads working together:
\begin{itemize}
    \item \textbf{Prefix-Matching Head}: The first attention head, located in a previous layer of the model, attends to prior tokens that are followed by the current token. This means it scans the sequence for earlier instances where the current token appears immediately after certain tokens, effectively performing prefix matching. This process identifies the ``attend-to'' token, which is the token that follows the current token in those previous occurrences.
    \item \textbf{Copying Head (Induction Head)}: The second attention head, known as the induction head, takes the ``attend-to'' token identified by the first head and copies it, increasing its output logits. By boosting the likelihood of this token in the model's output, the induction head extends the recognized sequence, effectively enabling the model to predict and generate sequences that mirror previously observed patterns.
\end{itemize}
Through the collaboration of these two heads, induction heads enable LLMs to recognize and replicate patterns within the input sequence, which is a fundamental aspect of in-context learning.
\citet{elhage2021mathematical} find two types of circuits in the transformer: i) ``query-key'' (QK) circuits; ii) ``output-value'' (OV) circuits, which are crucial for knowledge retrieval and updating. 
The QK circuits play a crucial role in determining which previously learned token to copy information from. Conversely, the OV circuits determine how the current token influences the output logits.

\subsubsection{Generalization and Emergence of Abilities}
\label{section:gec}
Recent studies argue that LLMs exhibit emergent abilities, which are absent in smaller models but present in larger-scale models \citep{DBLP:conf/nips/SchaefferMK23}.
Existing research \citep{DBLP:journals/corr/abs-2201-02177,DBLP:conf/iclr/DoshiDHG24}, by observing the dynamic training process of models, has identified two important phenomena related to generalization and emergence: \textbf{grokking} and \textbf{memorization}. 

\textbf{Grokking} is a phenomenon observed in over-parameterized neural networks, where models that have severely overfitted training data suddenly and significantly improve their validation accuracy. 
Grokking is closely related to factors such as data, representations, and regularization. 
Larger datasets can decrease the number of steps needed for grokking to occur \citep{DBLP:journals/corr/abs-2401-10463}. 
Well-structured embeddings and regularization measures can accelerate the onset of grokking, with weight decay standing out as particularly effective in strengthening generalization capabilities \citep{DBLP:conf/nips/LiuKNMTW22}. 
Recent studies have demonstrated that as the scale of models increases, more capabilities are acquired, such as more precise spatial and temporal representations \citep{DBLP:conf/nips/SchaefferMK23, DBLP:conf/iclr/GurneeT24}.

\textbf{Memorization} often refers to the phenomenon that models predict with statistical features rather than causal relations. 
\citet{DBLP:conf/iclr/NandaCLSS23} hypothesize that memorization constitutes a phase of grokking. They split training into three continuous phases: memorization, circuit formation, and cleanup. Their experiment results show that grokking, rather than being a sudden shift, arises from the gradual amplification of structured mechanisms encoded in the weights, followed by the later removal of memorizing components.

\subsection{Interpretability in Model Safety Auditing}
\label{section:audit_model}

As the application of LLMs in high-stake domains such as healthcare, finance, and law continues to increase, it is imperative to not only assess their accuracy but also scrutinize their safety and reliability \citep{li2023survey,liu2023trustworthy}.
From a societal perspective, the widespread adoption of LLMs across various domains presents potential risks. These risks could arise from a disconnect between LLM developers and users. The former often prioritize technological advancements over practical applications, while the latter may introduce LLMs into their fields without sufficient safety measures or proven success replication.

Therefore, \cite{mokander2023auditing} have proposed that model safety should be audited by third-party entities to rapidly identify risks within LLM systems and issue safety alerts. The auditing process for model safety comprises three steps:
\begin{itemize}
\item \textbf{Governance Audit}: This involves evaluating the design and dissemination of LLMs to ensure their compliance with relevant legal and ethical standards.
\item \textbf{Model Review}: This step entails a thorough examination of the LLMs themselves, including aspects such as performance, safety, and fairness.
\item \textbf{Application Review}: This involves assessing applications based on LLMs to ensure their reliability and safety in practical use.
\end{itemize}

Interpretability plays a crucial role in every step of auditing model safety. 
In the phase of data preprocessing, recent studies on interpretability \citep{dziri2022origin} have shown that a lack of relevant data or the presence of duplicate training data in the dataset can lead to hallucinations.
This is because long-tail instances are abundant in the training data, and LLMs tend to underfit when learning them \citep{wang2017learning,kandpal2023large}.
Redundant data causes LLMs to memorize redundant information, consuming a large capacity and ultimately leading to performance degradation \citep{hernandez2022scaling}.
During the phase of training, interpretability can utilize concrete evidence to verify risks models, such as demonstrating internal misalignment or mesa-optimization (the emergence of unintended sub-agents within the model) \citep{bereska2024mechanistic}.

\citet{DBLP:conf/icml/LeeBPWKM24} have found that utilizing pairs of toxic and non-toxic samples to perform Direct Preference Optimization (DPO) LLMs can improve LLMs in terms of non-toxic content generation.

In the phase after training, interpretability can analyze the model's unsafe behavior and identify its characteristics, thereby assessing the reliability and safety of LLM systems in practical applications.
\citet{DBLP:conf/iclr/HalawiDS24} study harmful imitation in ICL through a lexical lens to examine the internal representations of trained LLMs. They discover two related phenomena: overthinking and misinduction heads, where heads in deep layers focus on and replicate false information from previous demonstrations.

As \citet{gabriel2022challenge} claims, using interpretability tools to audit model safety is a productive method to ensure the model's reliability and trustworthiness.

\subsection{Interpretability for Alignment}
\label{section:aligning_values}

Reinforcement Learning from Human Feedback (RLHF) has emerged as the prevailing alignment technique for LLMs, aiming to more closely align complex AI systems with human preferences \citep{DBLP:journals/corr/abs-2406-16897,DBLP:journals/corr/abs-2303-08774,DBLP:journals/corr/abs-2307-09288}.
Its principal advantage is that it capitalizes on humans at judging appropriate behaviors through ranking rather than directly providing demonstrations or manually setting rewards.
Nevertheless, RLHF still has limitations, including data quality concerns (inconsistencies in human annotator preferences), risks of reward misgeneralization, reward hacking, and complexity  in policy optimization \citep{DBLP:conf/aaai/PengNKIK22,DBLP:journals/tmlr/CasperDSGSRFKLF23,DBLP:journals/corr/abs-2303-08774}.

It is generally believed that alleviating the above-mentioned issues from a data perspective is impractical. In contrast, interpretability aiming to understand the internal mechanisms of LLMs, provides an alternative solution to these problems \citep{DBLP:journals/corr/abs-2403-08946}.
This is because interpretability can be used as a tool to identify safety-related features (e.g., privacy, bias), which could be explored to steer LLMs towards desired behaviors (e.g., privacy-preserving text generation, unbiased text generation).

\subsubsection{Mitigating Hallucinations}
\label{section:ihallucinations}

Interpretability, compared to RLHF, offers a significantly cost-effective method to mitigate hallucinations, while also providing the benefits of adjustability and minimal invasiveness.
It is well known that LLMs learn and store factual knowledge encountered during pre-training \citep{DBLP:conf/emnlp/ZhaoZXZ022,DBLP:conf/eacl/CohenGBG23}.
For example, when the an LLM is prompted with ``The Space Needle is located in the city'', it might retrieve the stored fact and correctly predict ``Seattle''. However, these stored facts can become incorrect or outdated over time, leading to the generation of factual errors \citep{DBLP:journals/tacl/CohenBYGG24}.

Interpretability can identify where and how facts are stored in LMs, how they are recalled during reasoning, and how to direct activations towards facts through knowledge editing methods, thereby avoiding hallucinations \citep{DBLP:conf/nips/MengBAB22,DBLP:journals/corr/abs-2404-03646}.
\citet{DBLP:conf/nips/MengBAB22} use path patching to locate the components responsible for storing factual knowledge, and then edit the facts by updating only the parameters of these components (e.g., replacing ``Seattle'' with ``Paris'').
Another study investigates the sources of hallucinations through perturbation analysis of source token contribution patterns \citep{DBLP:journals/tacl/XuABMC23}. Their findings suggest that hallucinations may stem from the model's over-reliance on a limited set of source tokens.
\citet{DBLP:journals/corr/abs-2402-09733} apply PCA to derive the direction of the final hidden state corresponding to the correct answer and used this direction to enhance the hidden representation to reduce hallucinations.

Although these methods are effective for targeted editing, their ability to update relevant knowledge and prevent forgetting still requires further research \citep{DBLP:conf/eacl/CohenGBG23}.

\subsubsection{Privacy Protection}
\label{section:iprivacy}

Powerful LLMs are devouring existing data from various domains.
Such training data are primarily collected from the Internet. Due to the huge amount of training data and wide range of domains, it is difficult to thoroughly examine the data quality and confidentiality \citep{DBLP:conf/acl/PiktusAVLDLJR23}.
Previous studies \citep{DBLP:conf/emnlp/MireshghallahU022,DBLP:conf/sp/LukasSSTWB23} have demonstrated that LLMs tend to memorize their training data, leading to potential private information leaks.
Some attacks employ specially crafted prompts to steer LLMs away from their standard chatbot style of generation, exacerbating this privacy issue \citep{DBLP:journals/corr/abs-2311-17035}.
Traditional preprocessing techniques such as data cleaning \citep{DBLP:conf/acl/LisonPSBO20}, along with training methods based on specific data SFT and RHLF, although effective, are limited by data quality and LLMs' learning capabilities, making it challenging to completely eliminate the issue.

Interpretability techniques address this challenge from the perspective of the model itself. They can serve as tools to determine whether LLMs have internalized specific knowledge and to eliminate private information through knowledge editing.
Firstly, by explaining the relationship between factual knowledge and neuron activation \citep{DBLP:conf/nips/HaseBKG23}, we can investigate whether and where a piece of factual knowledge is stored within the model.
\citet{DBLP:conf/acl/YinZR024} have develop a semantically constrained projected gradient descent method to explore whether LLMs possess certain knowledge that is independent of input prompts.
\citet{DBLP:journals/corr/abs-2403-19851} utilize high-gradient weights in shallow layer attention heads to precisely locate memorized passages within the model. This localization technique identifies specific attention heads, which are then fine-tuned to forget the memorized knowledge, thereby enhancing the privacy protection capability of LLMs.

\subsubsection{Reducing Toxicity}
\label{section:itoxicity}

LLMs are trained on virtually all useful textual corpora available on the internet. These datasets often contain toxic elements that are difficult to completely eliminate and can be learned by LLMs, leading to the generation of toxicity content.

Interpretability methods can be used to identify and reduce toxicity.
A recent study has employed linear probe models and multi-layer perceptron (MLP) block analysis techniques to identify and examine specific value vectors in the GPT-2 that promote toxic outputs. 
Based on their findings, the researchers of this study have proposed two methods to reduce toxicity.
Firstly, by intervening in the model’s forward pass during the generation process (specifically by subtracting toxic vectors), they can reduce the model's propensity to produce toxic outputs while maintaining the quality of the generated text.
Secondly, by utilizing Direct Preference Optimization (DPO) on carefully curated paired datasets, recent studies have discovered that minimal parameter changes were sufficient to bypass toxic vectors, thereby reducing toxic outputs. 
In this aspect, \citet{geva2022transformer} propose a method to mitigate toxic generation by identifying and activating neurons within the feed-forward layers responsible for promoting innocuous or safe words.
\citet{balestriero2023characterizing} analyzed and characterized LLMs' internal multi-head attention mechanisms and feed-forward networks from a geometric perspective. They employ spline formulation \citep{balestriero2018spline} to extract key geometric features from MLPs, which not only reveals the intrinsic structure of the models but also enables the identification and classification of toxic speech without additional training.
By feeding prompts with negative and positive prefixes into LLMs, \citet{leong2023self} analyze internal contextualized representations to identify the toxicity direction of each attention head. They then utilize the original context prompt and guide the update of the current value vectors in the opposite of the detected toxicity direction to reduce toxicity.

\subsubsection{Eliminating Biases}
\label{section:ibias}
Social biases existing in training data for LLMs have raised concerns about the exacerbation of societal biases with the deployment of LLMs in real-world scenarios.
Plenty of efforts have been dedicated to detecting and eliminating social biases in LLMs \citep{sanh2020movement, joniak2022gender,fleisig2023fairprism,rakshit2024prejudice}.
Common debiasing methods based on retraining or fine-tuning LMMs with anti-bias datasets have certain limitations, such as limited generalization ability, high cost, and catastrophic forgetting \citep{zhao2024explainability}.

Interpretability techniques provide a unique perspective on mitigating these biases by revealing the mechanisms through which biases are embedded within models. 
For instance, \citet{ma2023deciphering} effectively debiase LLMs by detecting biased encodings through probing attention heads and evaluating their attributions, followed by pruning these biased encodings.
Inspired by induction heads, \citet{yang2023bias} measured the bias scores of attention heads focusing on specific stereotypes in pre-trained LLMs. They identify biased heads by comparing the changes in attention scores between biased heads and regular heads. By masking the identified biased heads, they effectively reduced the gender bias encoded in LLMs.
\citet{liu2024devil} explores an interpretability method to mitigate social biases in LLMs by introducing the concept of social bias neurons. First, they introduce an integrated gap gradient similar to the gradient-based attribution method, which precisely locates social bias neurons by back-propagating and integrating the gradients of the logits gap. Then, they mitigate social bias by suppressing the activation of the precisely located neurons. Extensive experiments validate the effectiveness of their method and reveal the potential applicability of interpretability methods in eliminating biases in LLMs.

\subsection{Potential Risks in the Use of Interpretability}
\label{section:safety_risks}
While interpretability is generally beneficial, it also introduces potential risks. These risks include dual-use of technology (Section \ref{section:dual-use}), adversarial attacks (Section \ref{section:attacks}), misunderstanding or over-trusting explanations (Section \ref{section:misunderstanding}), and accelerating uncontrollable risks with interpretability research (Section \ref{section:uncontrollable}).
\subsubsection{Dual-Use} 
\label{section:dual-use}
Dual-use refers to the same technology being applicable for both beneficial and harmful purposes (\textbf{misuse}). As previously mentioned, interpretability can locate and edit fine-grained features, which can be used to align LLMs to human values, but it can also be misused to enhance misalignment (such as invading privacy and exacerbating biases). Similarly, while interpretability may help improve the adversarial robustness of LLMs, it could also facilitate the development of stronger adversarial attacks.

\subsubsection{Adversarial Attacks} 
\label{section:attacks}
By understanding the decision-making process of LLMs, attackers can more easily identify the LLMs' vulnerabilities. They can use interpretability information to precisely generate adversarial samples that seem normal but can induce LLMs to make incorrect judgments. For instance, \citet{jain2023mechanistically} used network pruning, attention map activation, and probe classifiers to track changes in model capabilities from pre-training to fine-tuning. These tools help identify significant weights and key neurons. By fine-tuning on unrelated tasks, key neurons can be easily disrupted, thereby impairing LLM's capability and safety.

\subsubsection{Misunderstanding or Overtrusting} 
\label{section:misunderstanding}
Misunderstanding or overtrusting interpretability leads to decision-making errors. Interpretability techniques may not completely or accurately reflect the complex decision-making processes of LLMs. Overly simplified or inaccurate explanations may mislead users, resulting in misconceptions regarding the capabilities and limitations of LLMs. This misunderstanding may cause users to over-rely on LLMs for critical decisions, ultimately leading to adverse outcomes by ignoring other important factors. For instance, in medical diagnosis, overtrusting misleading explanations from LLMs could impair doctors' judgments.

\subsubsection{Accelerating Uncontrollable Risks} 
\label{section:uncontrollable}
Research on interpretability can significantly accelerate the development of LLMs, potentially surpassing existing technologies aligned with human values, thereby introducing substantial and uncontrollable risks. Early interpretability studies have limited impact on the evolution of AI capabilities, but recent research has changed this scenario. For instance, OpenAI's ``Scaling Laws'' revealed the relationship between model size, data volume, and performance, guiding the training of larger and more efficient models, thereby greatly accelerating the intelligence enhancement of LLMs. Additionally, the discovery and application of the ``Chain-of-Thought'' strategy have significantly enhanced the reasoning, planning, and decision-making abilities of LLMs, enabling models to solve complex problems through step-by-step reasoning. However, if alignment or control technologies cannot keep pace with this rapid enhancement of capabilities, it could lead to serious and widespread crises. Powerful AI models might exhibit behaviors that are not in the best interest of humanity or that violate ethical norms, leading to unpredictable risks. Therefore, while advancing interpretability research, it is crucial to simultaneously strengthen the research and application of AI alignment and control technologies to ensure that the development of AI always aligns with human values and interests.

\subsection{Future Directions}
\label{section:idirections}

\textbf{Broader and Deeper Coverage of Capable Models and Behaviors} Currently, many interpretability studies are primarily based on toy or theoretical models, and the applicability and extensibility of the findings on these models to other models have not yet been fully verified. To make substantive progress in production environments and industrial applications, it is necessary to shift the research focus to more complex models and real-world application scenarios. In-depth research on capable models in their behavior in real environments will help develop more practical interpretability methods, enhancing the reliability and transparency of models in actual applications.

\textbf{Towards Universality} Current interpretability research mostly focuses on models with the same architectures. This limitation restricts the generalizability of interpretability methods across different models and architectures. Identifying universal reasoning patterns within models and developing a unified theoretical framework are crucial for enhancing the generalizability of interpretability research. By establishing common interpretability methods applicable to various tasks and model structures, we can promote the development of interpretability research and enhance its applications across different domains.

%% file: taxonomies/Technology_Roadmaps_Taxonomy.tex
\tikzstyle{my-box}=[
    rectangle,
    draw=hidden-draw,
    rounded corners,
    text opacity=1,
    minimum height=1.5em,
    minimum width=5em,
    inner sep=2pt,
    align=center,
    fill opacity=.5,
    line width=0.8pt,
]
\tikzstyle{leaf}=[my-box, minimum height=1.5em,
    fill=hidden-pink!80, text=black, align=center,font=\normalsize,
    inner xsep=2pt,
    inner ysep=4pt,
    line width=0.8pt,
]
\begin{figure*}[t!]
    \centering
    \resizebox{\textwidth}{!}{
        \begin{forest}
            forked edges,
            for tree={
                grow=east,
                reversed=true,
                anchor=base west,
                parent anchor=east,
                child anchor=west,
                base=center,
                font=\large,
                rectangle,
                draw=hidden-draw,
                rounded corners,
                align=center,
                text centered,
                minimum width=5em,
                edge+={darkgray, line width=1pt},
                s sep=3pt,
                inner xsep=2pt,
                inner ysep=3pt,
                line width=0.8pt,
                ver/.style={rotate=90, child anchor=north, parent anchor=south, anchor=center},
            },
            where level=1{text width=8em,font=\normalsize,}{},
            where level=2{text width=13em,font=\normalsize,}{},
            where level=3{text width=20em,font=\normalsize,}{},
            where level=4{text width=13.5em,font=\normalsize,}{},
            [
                Technology Roadmaps to \\ LLM Safety in Practice
                [
                    Training
                    [
                        Training Data
                        [
                            OpenAI{,} Anthropic{,} Google DeepMind \\
                            Meta{,} Microsoft{,} Baidu Inc. \\
                            Mistral AI{,} 01.AI{,} Zhipu AI \\
                            Tiger Research{,} Alibaba Cloud \\
                            NVIDIA{,} DeepSeek-AI \\
                            , leaf
                        ]
                    ]
                    [
                        Training Methodology
                        [
                            OpenAI{,} Anthropic{,} Google DeepMind \\
                            Meta{,} Microsoft{,} Baidu Inc. \\
                            Mistral AI{,} NVIDIA{,} Zhipu AI \\
                            Tiger Research{,} Alibaba Cloud \\
                            01.AI{,} DeepSeek-AI{,} EleutherAI \\
                            Cohere{,} Center for AI Safety (CAIS) \\
                            , leaf
                        ]
                    ]
                ]
                [
                    Evaluation
                    [
                        Value Misalignment and \\ Robustness Evaluation
                        [
                            OpenAI{,} Anthropic{,} Google DeepMind \\
                            Meta{,} Microsoft{,} Baidu Inc. \\
                            Mistral AI{,} 01.AI{,} Zhipu AI \\
                            Tiger Research{,} Alibaba Cloud \\
                            NVIDIA{,} DeepSeek-AI{,} CeSIA \\
                            Apollo Research{,} Cohere{,} Inflection.AI \\
                            Model Evaluation and Threat Research \\
                            (METR){,} Allen Institute for AI \\
                            Alignment Research Center (ARC) \\
                            Center for AI Safety (CAIS){,} SaferAI \\
                            , leaf
                        ]
                    ]
                    [
                        Misuse and Autonomy \\ Risks Evaluation
                        [
                            OpenAI{,} Anthropic{,} Google DeepMind \\
                            Model Evaluation and Threat Research \\
                            (METR){,} Center for AI Safety (CAIS) \\
                            Alignment Research Center (ARC) \\
                            , leaf
                        ]
                    ]
                ]
                [
                    Deployment
                    [
                        Monitoring
                        [
                            OpenAI{,} Anthropic{,} Google DeepMind \\
                            Baidu Inc.{,} Tiger Research \\
                            Meta{,} CeSIA{,} Databricks \\
                            Center for AI Safety (CAIS){,} NVIDIA \\
                            AI Risk and Vulnerability Alliance (ARVA) \\
                            Allen Institute for AI{,} SaferAI \\
                            Redwood Research \\
                            , leaf
                        ]
                    ]
                    [
                        Guardrails
                        [
                            Mistral AI{,} Databricks{,} NVIDIA \\
                            , leaf
                        ]
                    ]
                ]
                [
                    Safety Guidance \\ Strategy
                    [   
                        OpenAI{,} Anthropic{,} Google DeepMind \\
                        Meta{,} Microsoft{,} Databricks \\
                        FAR AI{,} Dataiku{,} 01.AI \\
                        , text width=18em
                        , leaf
                    ]
                ]
            ]
        \end{forest}
    }
    \caption{Overview of technology roadmaps / strategies to LLM safety in practice.}
    \label{fig:safety_roadmap}
\end{figure*}
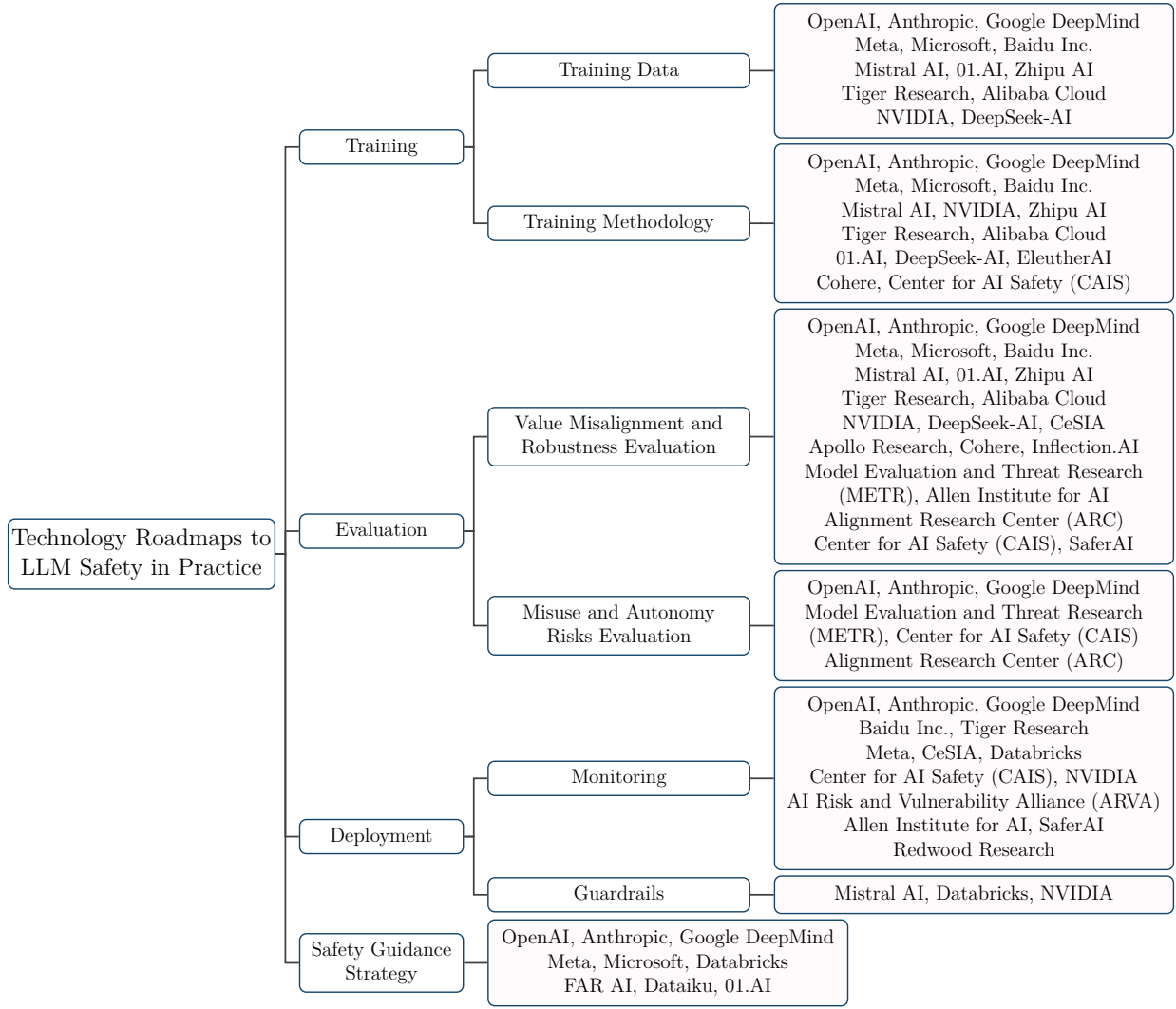

%% file: sections/Technology_Roadmaps.tex
\section{Technology Roadmaps / Strategies to LLM Safety in Practice}
\label{section:TechnicalRoadmapToLLMSafety}

\begin{table*}[t]
\centering
\resizebox{0.95\textwidth}{!}{
\begin{tabular}{l|c}
   \toprule
    \multirow{3}{*}{\textbf{AI Companies}} & OpenAI, Anthropic, Google DeepMind, Meta, Microsoft, Baidu Inc.,\\ 
    & Cohere, Apollo Research, Mistral AI, Databricks, 01.AI, Tiger Research, \\ 
    & Alibaba Cloud, NVIDIA, DeepSeek-AI, Zhipu AI, Inflection.AI, Dataiku \\
    \midrule
    \multirow{5}{*}{\textbf{AI Research Institutes}} & Alignment Research Center (ARC), Allen Institute for AI, \\ 
    & AI Risk and Vulnerability Alliance (ARVA), Center for AI Safety (CAIS), \\
    & FAR.AI, Open Web Application Security Project (OWASP), \\
    & Model Evaluation and Threat Research (METR), SaferAI, \\
    & CeSIA, Redwood Research, EleutherAI\\

   \bottomrule
\end{tabular}}
\caption{18 AI companies and 11 AI research institutes investigated in this section.}
\label{safety_roadmap}
\end{table*}

In order to address existing and anticipated safety risks of LLMs, many AI companies and research institutes are investing significant resources in exploring and implementing various safety technologies and strategies to ensure the reliability and safety of LLMs in real world applications. We hence investigate the solutions and strategies proposed and adopted by these companies and research institutes for LLM safety. We list the AI companies and institutes that we've investigated in Table \ref{safety_roadmap}.

After an in-depth investigation, we categorize these technology roadmaps and strategies adopted in practice into four modules: training, evaluation, deployment and safety guidance strategy. \enquote{Training} module investigates risk mitigation solutions and strategies used in LLM training. \enquote{Evaluation} module examines the performance of LLMs in many safety aspects. \enquote{Deployment} module focuses on potential risks of trained models deployed in systems that interact with users. Finally, \enquote{Safety Guidance Strategy} module provides comprehensive and systematic guidance on safety technologies.

\subsection{Training}
We discuss a set of safety measures for LLMs in the training stage, proposed or adopted by the investigated companies and institutes.

\subsubsection{Training Data}
Data quality is particularly important for both pre-training and post-training of LLMs. Due to the wide range of sources of training data used for LLMs, it is inevitable that there are redundancies, errors, and harmful contents in collected data. Such low-quality data not only affects the performance of LLMs, but also misguides LLMs to generate content that is not expected by humans, such as toxic, pornographic, and biased content. Therefore, data filtering is usually required to ensure the quality of data before training, where rule-based filtering methods and model-based content classifiers are commonly used.

GPT-4 \citep{DBLP:journals/corr/abs-2303-08774} identifies pornographic content in training data by combining a dictionary-based approach and a content classifier. Claude 3 \citep{Claude_3} integrates multiple data cleaning and filtering methods to improve data quality. Yi \citep{DBLP:journals/corr/abs-2403-04652} constructs a set of filters based on heuristic rules, keyword matching, and classifiers. Qwen \citep{DBLP:journals/corr/abs-2309-16609, DBLP:journals/corr/abs-2407-10671} develops a set of data preprocessing procedures in which humans work with models to perform data filtering by using models to score data content in combination with human review. In addition to this, some companies specify model development policies that require the training data of these models not to include user data in order to protect user privacy, e.g., Meta and Anthropic.

While filtered training data reduces harmful content, filtering is still not sufficient for LLMs to learn to perform in a safe way. In order to further improve the safety of LLMs, additional instruction data of a safe nature is usually added to training data. Gemini \citep{DBLP:journals/corr/abs-2312-11805} emphasizes the importance of adversarial query data in the post-training stage. It constructs a query dataset for post-training, which contains about 20 categories of harmful data by integrating expert authoring, model synthesis, and automated red teaming methods. Similarly, Yi \citep{DBLP:journals/corr/abs-2403-04652} constructs a comprehensive safety classification system, and then builds a safety dataset for SFT training based on this classification. It is worth noting that these safety datasets usually require human supervision or guidance to ensure their quality. For example, the safety datasets used in models such as Baichuan \citep{DBLP:journals/corr/abs-2309-10305} and TigerBot \citep{DBLP:journals/corr/abs-2312-08688} are constructed under the guidance of relevant domain experts, which highlights the importance of safety datasets compared to general datasets.

\subsubsection{Training Methodology}
Reinforcement learning from human feedback (RLHF) has proven a useful avenue for enhancing the safety of LLMs at the post-training stage \citep{DBLP:conf/nips/Ouyang0JAWMZASR22, DBLP:journals/corr/abs-2303-08774, DBLP:journals/corr/abs-2307-09288, DBLP:journals/corr/abs-2204-05862}. RLHF performs human feedback-based fine-tuning, which uses human preferences as a proxy to specify human values \citep{DBLP:journals/corr/abs-2309-15025}. Normally, RLHF consists of three core steps, which are (1) collecting human feedback data, (2) using the collected human feedback data to train reward models, and (3) fine-tuning LLMs using reinforcement learning algorithms such as Proximal Policy Optimization (PPO) \citep{DBLP:journals/corr/SchulmanWDRK17}.

Anthropic further proposes Constitutional AI \citep{DBLP:journals/corr/abs-2212-08073}, where a set of moral and behavioral principles, termed as a constitution, is developed for aligning LLMs via supervised learning and reinforcement learning. In Claude 3 \citep{Claude_3}, the rules of the constitution are derived from the Universal Declaration of Human Rights, Apple’s Terms of Service, Principles Encourage Consideration of Non-Western Perspectives, DeepMind’s Sparrow Rules, and Anthropic Customization Principles \citep{Claude_Constitution}. Alignment approaches similar to Constitutional AI are also used in Gemini \citep{DBLP:journals/corr/abs-2312-11805} and Qwen2 \citep{DBLP:journals/corr/abs-2407-10671}.

With Constitutional AI, a research team from Google delves into the comparison between reinforcement learning from AI feedback (RLAIF) \citep{DBLP:conf/icml/0001PMMFLBHCRP24} and RLHF, and demonstrates that RLAIF could be a competitive alternative to RLHF, thus reducing the reliance on expensive human annotation. In addition to the annotation cost, traditional RLHF methods usually involve optimizing the reward function for human preferences, which is effective but may bring about challenges such as increasing computational complexity and the need to consider the bias-variance trade-off when estimating and optimizing rewards \citep{DBLP:journals/corr/SchulmanMLJA15}. To mitigate these issues, DPO \citep{DBLP:conf/nips/RafailovSMMEF23} has been proposed to simplify the alignment process, reducing computational overhead and enabling more robust optimization by using preference data in a more direct way.

In addition to alignment training, interpretability methods can also be used to achieve safety controls over LLMs. The Center for AI Safety (CAIS) introduces the concept of representation engineering \citep{DBLP:journals/corr/abs-2310-01405} to improve the transparency of AI systems by leveraging cognitive neuroscience. Representation engineering extracts various safety-related concepts through representation learning methods, and then modifies or controls these safety-related conceptual representations through representation control methods to reduce risks associated with LLMs. They conduct extensive case studies using representation engineering on safety aspects such as honesty and ethics, demonstrating the potential of representation engineering to improve transparency, safety, and trust in AI systems.

\subsection{Evaluation}
As discussed before, evaluation is a very important tool for detecting various risks. As such, it is widely used as a practical safety measure by investigated companies and institutes. We follow our LLM safety taxonomy (see Figure \ref{fig:all}) to elaborate the evaluation measures used in practice.

\subsubsection{Value Misalignment and Robustness Evaluation}
For GPT-4 \citep{DBLP:journals/corr/abs-2303-08774}, OpenAI conducts internal quantitative evaluations following its content policies, e.g., evaluations on hate speech, self-injurious suggestions, and illegal suggestions. These evaluations measure the likelihood that GPT-4 generates content violating value alignment when given a prompt.

Similarly, to ensure that Claude 3 \citep{Claude_3} is as safe as possible prior to deployment, Anthropic's Trust and Safety team conducts a full multi-modal red teaming exercise to thoroughly evaluate Claude 3, including evaluations over trust and safety, bias, discrimination, and more.

CAIS explores the relationship between AI safety and general upstream capabilities (e.g., general knowledge and reasoning), and finds that many safety benchmarks are highly correlated with upstream model capabilities, which can lead to \enquote{safety washing} \citep{DBLP:journals/corr/abs-2407-21792}. In such cases, capability improvements are mischaracterized as safety progress. Based on these findings, they propose defining AI safety as a set of explicitly described research goals that are empirically separable from general capability progress, thus ensuring more accurate safety evaluation.

The Allen Institute for AI develops WildTeaming \citep{DBLP:journals/corr/abs-2406-18510}, an automated red teaming framework for identifying and reproducing human attacks.WildTeaming directly exploits jailbreak strategies of human users and leverages those strategies to address vulnerabilities in LLMs.

Evaluations over value misalignment and robustness emphasize the importance of safety benchmark development. In this respect, a number of companies are committed to the development of multi-dimension, multi-domain safety evaluation benchmarks. Alibaba Cloud launches a \enquote{$100$ Bottles of Poison} for Chinese LLMs, developing a value alignment evaluation benchmark (Cvalues) \citep{DBLP:journals/corr/abs-2307-09705}. In this benchmark, adversarial safety prompts in multiple categories are created under the guidance of experts from different domains, aiming to evaluate LLMs in terms of both safety and responsibility. Cohere releases the first multilingual human-labeled red team prompt datasets to distinguish between global and local harms \citep{DBLP:journals/corr/abs-2406-18682}. These datasets are used to evaluate the robustness of different alignment techniques in the face of preference distributions across geographies and languages.

\subsubsection{Misuse and Autonomy Risks Evaluation}
OpenAI conducts qualitative evaluations for GPT-4 in term of misuse. Over $50$ experts from cybersecurity, biorisk, and international security are engaged to adversarially test GPT-4 and provide general feedbacks. The feedbacks gathered from these experts are used for subsequent mitigations and improvements to GPT-4. In addition, a non-conventional weapons
proliferation evaluation on GPT-4 is also conducted, primarily to explore whether GPT-4 could provide the necessary information for proliferators seeking to develop, acquire or disperse nuclear, radiological, biological and chemical weapons \citep{DBLP:journals/corr/abs-2303-08774}.

For autonomy risks evaluation, OpenAI collaborates with Alignment Research Center (ARC) to evaluate GPT-4's ability to autonomously replicate itself and acquire resources through expert red teaming. Although the initial evaluation demonstrates the ineffectiveness of GPT-4 in autonomously replicating itself and acquiring resources, ARC warns such risks for future advanced LLMs \citep{DBLP:journals/corr/abs-2303-08774}.

Anthropic proposes multiple levels of evaluation for catastrophic risks. In order to address the problem that a predetermined safety threshold for a given level may be accidentally exceeded when training LLMs, safety researchers in Anthropic set up safety buffers for each risk level \citep{responsible_scaling_policy}. The buffer strategy designs the evaluation of each risk level to be triggered at a level slightly below the level of capability of that level, while setting the buffer size to be larger than the evaluation time interval. In this way, the likelihood of accidentally crossing safety boundaries due to rapid increases in model capability is reduced, thus providing more time for researchers and developers to prepare and implement appropriate safety measures. Similarly, Google DeepMind develops an early warning evaluation to periodically test the capabilities of frontier models to check if they are approaching critical capability levels \citep{Frontier_Safety_Framework}.

Model Evaluation and Threat Research (METR), formerly known as ARC Eval, is dedicated to evaluating whether advanced AI systems pose a catastrophic risk to society, and is now focusing on evaluating the autonomy of LLMs. It argues that unlike the ability to develop biological weapons or execute high-value cyberattacks, having autonomy does not directly enable AI systems to cause catastrophic adverse consequences. But autonomy is a measure of the extent to which an AI system can have a profound impact on the world with minimal human involvement, and such a metric is useful in a variety of threat models. Following this, METR has recently released autonomy evaluation resources that include a task suite, software tools, and guidelines for accurately measuring LLM capabilities \citep{Autonomy_Evaluation_Resources}.

Apollo Research focuses on the evaluation of strategic deception. It finds that under varying levels of stress, GPT-4 engages in illegal behavior such as insider trading and lying about its actions \citep{DBLP:journals/corr/abs-2311-07590}. This finding demonstrates that AI systems may adopt strategies that humans do not approve of in order to help themselves.

\subsection{Deployment}
During the deployment stage, one of the most commonly used safety techniques is monitoring, whereby the environment in which LLMs are located is monitored to identify possible risks and mitigate them through a variety of predefined measures. Another used safety technique is guardrail, which centers on risk prevention.

\subsubsection{Monitoring}
In order to monitor the risks that may occur during the interaction between LLMs and users, researchers develop a variety of monitoring tools for scrutinizing LLMs inputs and outputs and predicting risks. GPT-4 \citep{DBLP:journals/corr/abs-2303-08774} uses a detection system combining machine learning and rule-based classifiers to identify content that may violate their usage policies. When such content is identified, the deployed monitoring system takes defensive measures such as issuing warnings, temporarily suspending, or in severe cases, banning the corresponding users.

Similarly, Claude 3 \citep{Claude_3} has a content classifier that identifies any content that violates the Acceptable Use Policy (AUP) \citep{aup}. User prompts that are flagged as violating AUP trigger a command for Claude to respond more carefully. In the case of particularly serious or harmful user prompts, Claude 3 is prevented from responding at all. In the case of multiple violations, the Claude is terminated. It is important to note that these classifiers need to be updated regularly to address the changing threat landscape.

ERNIEBot \citep{DBLP:journals/corr/abs-2107-02137} deploys a content review system that intervenes on LLM inputs by means of manual review or rule-based filtering to ensure that LLM inputs conform to a specific standard or specification. On the output side, after filtering out harmful and sensitive words in an LLM-generated response through the content review system, the safety content of the response is used as the final output by means of semantic rewriting.

Garak \citep{garak} is a vulnerability scanner for LLMs, which checks models for hundreds of different known weaknesses using thousands of different prompts, and checks model responses to see if the model is at risk in some way.

In contrast to the private monitoring systems mentioned above, there are a number of open-source monitoring tools. Perspective API \citep{perspectiveapi} identifies offensive, rude, discriminatory, and other toxic content in online conversations. WildGuard \citep{DBLP:journals/corr/abs-2406-18495} evaluates the safety of user interactions with LLMs through three safety audit tasks, including Harmfulness of Prompt, Harmfulness of Response, and Rejection of Response. Llama Guard \citep{DBLP:journals/corr/abs-2312-06674} is used to detect whether the input prompts and output responses generated by LLMs violate predefined safety categories. Llama 3 \citep{DBLP:journals/corr/abs-2407-21783} uses two prompt-based filtering mechanisms, Prompt Guard and Code Shield. Prompt Guard is used to detect prompting attacks, which are mainly two types of attacks, direct jailbreaks and indirect prompt injections. Code Shield is able to detect the generation of unsafe code before it can potentially enter a downstream use case (e.g., a production system), and is able to support seven programming languages.

While these monitoring tools can assist humans in monitoring risky content generated by LLMs, the robustness of the tools themselves remains an issue. Previous research has found evidence of bias in the Perspective API, e.g., giving higher toxicity scores to text containing racial or gender identity terms or phrases associated with African American English \citep{DBLP:conf/acl/SapCGCS19}. In the context of a content-moderation tool, these kinds of biases can cause real harm, as they may lead to suppression of speech within or about marginalized communities. Therefore, it is also important to review these widely recognized monitoring tools. IndieLabel \citep{IndieLabel}, a detection tool for the Perspective API, prompts users to assign a toxicity score to a small number of text examples (about 20 social media posts). A lightweight model is then trained to predict the user's perception of a large number of examples (roughly thousands of social media posts), and these predictions are used to uncover areas of potential disagreement between users and the Perspective API. Similarly, BELLS \citep{DBLP:journals/corr/abs-2406-01364} is a framework for evaluating the reliability and generalizability of LLMs monitoring systems, which allows for a comparison of the reliability of multiple monitoring tools, thus creating a performance competition in anomaly detection.

\subsubsection{Guardrails}
Guardrails are a set of programmable constraints or rules that sit between users and LLMs \citep{DBLP:conf/emnlp/RebedeaDSPC23, DBLP:journals/corr/abs-2310-06825, Databricks_Guardrails}. These guardrails monitor, influence, and instruct the interaction of LLMs with users, usually by setting system prompts in LLM's front-end applications, which force LLMs to enforce the output constraints. For example, models are required through system prompts to help users in a caring, respectful, and honest manner; to avoid harmful, unethical, biased, or negative content; and to ensure that responses promote fairness and positivity. Thereby, LLM outputs are restricted within these guardrails to ensure their safety. Currently, the guardrail technology is widely used by many AI companies, such as NVIDIA, Mistral AI, and Databricks.

NeMo Guardrails \citep{DBLP:conf/emnlp/RebedeaDSPC23} is an open source toolkit released by NVIDIA to add programmable guardrails to LLM-based session systems. NeMo Guardrails supports three types of guardrails: topical guardrails, safety guardrails, and security guardrails. Topical guardrails are designed to ensure that conversations focus on specific topics and prevent them from straying into undesirable areas. Safety guardrail ensures that interactions with LLMs do not result in misinformation, malicious responses, or inappropriate content. Security guardrails prevent LLMs from executing malicious code or calling external applications in ways that pose security risks.

\subsection{Safety Guidance Strategy}

In order to address LLM safety more systematically, many companies and research institutes publish their own safety guidance strategies, which are used to provide theoretical and technical guidance throughout the entire lifecycle, including model development and deployment.

OpenAI develops a Preparedness Framework \citep{preparedness}, describing OpenAI's process for tracking, evaluating, forecasting, and protecting against the catastrophic risks posed by increasingly powerful models. The framework categorizes risk levels as Low, Medium, High, and Critical. The framework tracks risks such as Cybersecurity, Chemical, Biological, Nuclear, and Radiological (CBRN) threats, Persuasion, and Model Autonomy.

Claude's safety team proposes Responsible Scaling Policy (RSP) \citep{responsible_scaling_policy}, a framework for assessing and mitigating potentially catastrophic risks of AI models. RSP defines a concept referred to as AI safety levels (ASL) for catastrophic risks. For Claude 3 \citep{Claude_3}, three sources of potential catastrophic risks have been evaluated: biological capabilities, cyber capabilities, and autonomous replication and adaptation (ARA) capabilities. Evaluation results show that Claude 3 is at the ASL-2 level, indicating that the model shows early indications of hazardous capabilities, but the information is not yet useful because it is not sufficiently reliable or does not provide information that is not available from search engines.

Google DeepMind proposes a frontier safety framework that aims to address the serious risks that may arise from the powerful functionality of future AI models \citep{Frontier_Safety_Framework}. The framework proposes two mitigations to address the safety issues of models with critical functionality, which are security mitigations to prevent leakage of model weights, and deployment mitigations to manage access to critical functionality. In addition to this, the framework also specifies protocols for the detection of capability levels at which models may pose severe risks (Critical Capability Levels, CCLs), addressing four categories of risks: Autonomy, Biosecurity, Cybersecurity, and Machine Learning R\&D.

Unlike the above technical guidance frameworks, FAR AI proposes a theoretical Guaranteed Safety AI framework (GS-AI), the core idea of which is to apply formal methods to quantitatively guarantee the safety properties of AI systems \citep{DBLP:journals/corr/abs-2405-06624}. Traditional AI safety evaluation often relies on a large number of empirical tests, and it is difficult to rigorously prove that the behavior of AI in various possible scenarios is as expected. GS-AI hopes to guarantee that an AI system meets a series of preset safety constraints by means of mathematical proofs or probabilistic arguments under certain assumptions. The GS-AI framework consists of three elements: world model, safety specification and verifier. The world model seeks to provide a comprehensive mathematical representation of the AI system and its environment, covering the possible impacts of the AI system as completely as possible. The safety specification is used to formally define the constraints that the AI system needs to comply with. The validator uses automated mathematical reasoning tools to determine whether the behavioral trajectory of the AI system in a given world model satisfies the safety specification. Through the interaction of these three elements, the safety of the AI system is guaranteed in a quantitative form. It is important to note, although GS-AI is still only a theoretical framework, it provides a promising strategy for ensuring the safety of AI systems.

There are also some guidance frameworks that have not been detailed, such as Databricks AI Security Framework (DASF) \citep{DASF}, Dataiku's RAFT (Responsible, Accountable, Fair and Transparent) Framework for Responsible AI \citep{RAFT}, Meta's Best Practice Safety Guidance \citep{DBLP:journals/corr/abs-2407-21783}, and 01.AI's Full Stack Responsible AI Safety Engine \citep{DBLP:journals/corr/abs-2403-04652}. The emergence of these guidance frameworks also suggests that LLM safety cannot simply be viewed in the same light as general capabilities, but should be emphasized by researchers to form systematic safety research.

\subsection{Discussion}

During our investigation, we find that certain companies do not make public the technical reports on their LLMs. We are hence not aware of their technical approaches to LLM safety. On the other hand, some LLMs with public technical reports do not specifically discuss safety in the technical reports, but focusing more on capability performance. In contrast, closed-source LLMs companies, represented by OpenAI and Anthropic, carry out more comprehensive safety strategies, not just injecting safety elements into training stages, but equally working on deployment, evaluation and protection against high-level risks. Additionally, a number of research institutions are also proposing LLMs and AI safety roadmaps. In contrast to safety techniques, these roadmaps do not provide relatively bottom-level safety approachs, but rather provide guidance on safety from a high-level perspective.

%% file: taxonomies/Governance_Taxonomy.tex
\tikzstyle{my-box}=[
    rectangle,
    draw=hidden-draw,
    rounded corners,
    text opacity=1,
    minimum height=1.5em,
    minimum width=5em,
    inner sep=2pt,
    align=center,
    fill opacity=.5,
    line width=0.8pt,
]
\tikzstyle{leaf}=[my-box, minimum height=1.5em,
    fill=hidden-pink!80, text=black, align=center,font=\normalsize,
    inner xsep=2pt,
    inner ysep=4pt,
    line width=0.8pt,
]
\begin{figure*}[t!]
    \centering
    \resizebox{\textwidth}{!}{
        \begin{forest}
            forked edges,
            for tree={
                grow=east,
                reversed=true,
                anchor=base west,
                parent anchor=east,
                child anchor=west,
                base=center,
                font=\large,
                rectangle,
                draw=hidden-draw,
                rounded corners,
                align=center,
                text centered,
                minimum width=5em,
                edge+={darkgray, line width=1pt},
                s sep=3pt,
                inner xsep=2pt,
                inner ysep=3pt,
                line width=0.8pt,
                ver/.style={rotate=90, child anchor=north, parent anchor=south, anchor=center},
            },
            where level=1{text width=7em,font=\normalsize,}{},
            where level=2{text width=22em,font=\normalsize,}{},
            where level=3{text width=18em,font=\normalsize,}{},
            where level=4{text width=20em,font=\normalsize,}{},
            [Governance
                [Proposals
                  [International Cooperation Proposals
                    [\citet{zhang2022enhancing}, leaf]
                    [\citet{DBLP:journals/corr/abs-2406-17864}, leaf]
                    [\citet{brookingsStrengtheningInternational}, leaf]
                  ]
                  [Technical Oversight Proposals
                    [\citet{turchin2020classification}, leaf]
                    [\citet{DBLP:journals/corr/abs-2310-17688}, leaf]
                    [\citet{bommasani2024considerations}, leaf]
                  ]
                  [Ethics and Compliance Proposals
                    [\citet{mantymaki2022defining}, leaf]
                    [\citet{DBLP:journals/corr/abs-2407-12929}, leaf]
                    [\citet{DBLP:conf/chi/TahaeiCQKMSLBAH23}, leaf]
                  ]
                ]
                [Policies
                  [Current Policy Evaluation
                    [\citet{fjeld2020principled}, leaf]
                    [\citet{DBLP:journals/corr/abs-2307-12218}, leaf]
                    [\citet{DBLP:journals/corr/abs-2407-07300}, leaf]
                  ]
                  [Policy Comparison
                    [\citet{engler2023eu}, leaf]
                    [\citet{ceimia}, leaf]
                    [\citet{dlapiperComparingExecutive}, leaf]
                    [\citet{comunale2024economic}, leaf]
                  ]
                ]
                [Visions
                  [Long-term Vision
                    [\citet{wef2021}, leaf]
                    [\citet{allen2024roadmap}, leaf]
                    [\citet{itic2024}, leaf]
                    [\citet{futureoflifeTurningVision}, leaf]
                  ]
                  [Vision of Technological and Social Integration
                    [\citet{upennArtificialIntelligence}, leaf]
                    [\citet{csisPathTrustworthy}, leaf]
                    [\citet{kai2024}, leaf]
                    [\citet{whitehouseBlueprintBill}, leaf]
                  ]
                  [Risks and Opportunities in Realizing Visions
                    [\citet{upennArtificialIntelligence}, leaf]
                    [\citet{margetts2022rethinking}, leaf]
                    [\citet{ai2024artificial}, leaf]
                    [\citet{Covino2024}, leaf]
                  ]
                ]
              ]
        \end{forest}
    }
    \caption{Overview of Governance Proposals, Policies and Visions.}
    \label{fig:Overview_of_catastrophic_misuses}
\end{figure*}
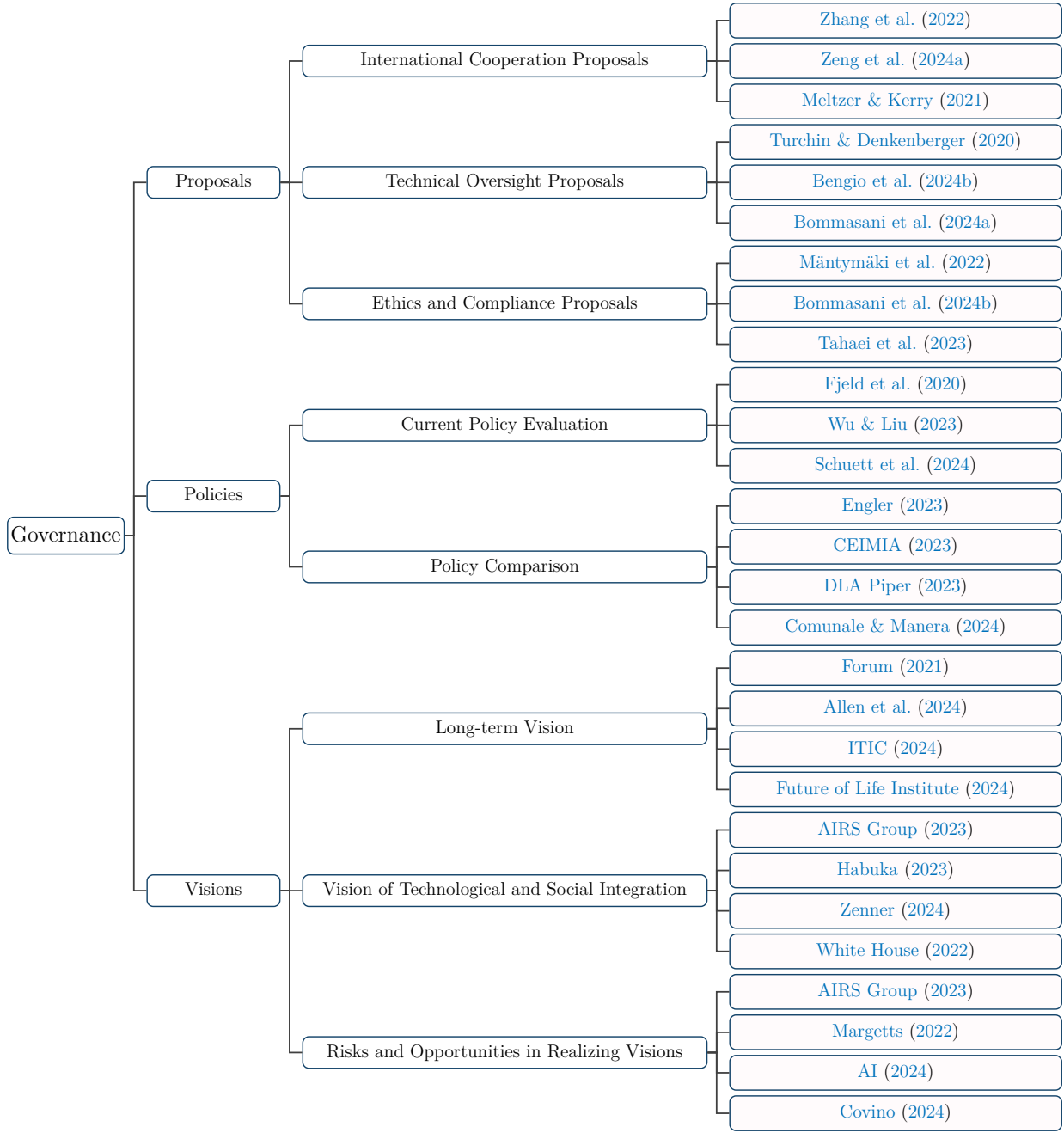

%% file: sections/Governance.tex
\section{Governance}
\label{section:Governance}
As the integration of AI into various aspects of human endeavor accelerates, the necessity for comprehensive and robust governance frameworks becomes increasingly critical. Effective governance in AI is not only about ensuring that technologies operate within set legal and ethical boundaries but also about steering these innovations towards the greater good while mitigating associated risks. This section delves into the multifaceted domain of AI governance, investigating and discussing proposals, policies, and visions that collectively shape the future of AI development and deployment.

In the ``Proposals'' dimension, we explore a range of initiatives that aim at strengthening international cooperation and regulatory oversight. These proposals highlight the urgent need for a synchronized approach to address the complex, cross-border challenges AI presents. We examine specific regulatory proposals targeted at high-stakes AI applications, emphasizing the balance between innovation and control, and the imperative for ethical frameworks to guide AI integration \citep{brookingsStrengtheningInternational, puscas2023ai, cass2024framework}.

Under ``Policies'', our focus shifts to evaluating existing policies and their efficacy in addressing the fast-evolving landscape of AI technologies. By comparing policies across different nations, we gain insights into successful governance models and the factors that influence their effectiveness. This analysis is crucial for understanding the dynamic interplay between technological advances and regulatory frameworks, guiding policymakers in crafting adaptable and forward-thinking AI regulations \citep{galindo2021overview, ceimia, DBLP:journals/corr/abs-2406-17864}.

Lastly, the ``Visions'' component presents forward-looking perspectives on AI governance. It outlines the long-term goals and the envisioned integration of AI into society, discussing both potential benefits and risks involved. These visions are grounded in a deep understanding of ever-increasing AI capabilities and societal needs, aiming to propose a harmonious path forward, which aligns technological advancement with human values \citep{dafoe2018ai, Taeihagh2021Governance, bateman2024beyond}.

Together, the three dimensions aim to provide a comprehensive overview of the current state and future directions of AI governance, offering valuable insights and recommendations for policymakers, researchers, and industry leaders engaged in shaping the trajectory of AI (especially LLM) development.

\subsection{Proposals}

We group the investigated AI governance proposals into three categories: international cooperation proposals, technical oversight proposals and proposals for ethics and compliance. 

\subsubsection{International Cooperation Proposals}

AI has become a central technological force with vast potential for societal and economic development. However, its rapid advancement presents considerable challenges, including risks related to privacy, ethical considerations, and misuse discussed before. These challenges necessitate effective governance frameworks that span national borders. We hence probe into current progress in international cooperation in AI regulation, and outline the need for a coordinated global response to mitigate risks while promoting innovation.

Given AI's global nature and cross-border impact, effective governance requires international cooperation. As highlighted in a growing body of literature, including studies from the Stanford HAI white paper \citep{zhang2022enhancing} and the AI Risk Categorization Decoded project (AIR 2024) \citep{DBLP:journals/corr/abs-2406-17864}, fragmented national regulations are insufficient to manage the complex global AI ecosystem. Instead, global cooperation is essential for promoting robust, fair, and safe AI development \citep{brookingsStrengtheningInternational, didomiPivotalMoment, puscas2023ai, cass2024framework}.

Governments worldwide are grappling with how to regulate AI effectively. While countries like the United States, China, and members of the European Union have made strides in regulating AI, these efforts are often fragmented and driven by national interests. For instance, the U.S. seeks to maintain a competitive stance in AI development, driven by private sector investment and an open innovation system \citep{brookingsStrengtheningInternational}. The European Union’s AI Act\footnote{\url{https://artificialintelligenceact.eu/}} is one of the most comprehensive regulatory frameworks, focusing on ensuring safety, accountability, and human-centric AI \citep{DBLP:journals/corr/abs-2406-17864}. Meanwhile, China's regulations, such as the Interim Measures for the Management of Generative AI Services,\footnote{\url{https://www.cac.gov.cn/2023-07/13/c_1690898327029107.htm}} suggest a harmonious philosophy that aims to balance AI technology innovation with concerns about societal risks that AI technologies may cause \citep{DBLP:journals/corr/abs-2406-17864}. However, these national/regional regulations alone are inadequate to address global risks posed by AI, such as cross-border data privacy issues, AI-driven misinformation, and the potential for AI systems to exacerbate geopolitical tensions \citep{didomiPivotalMoment}.

International cooperation on AI regulation is hence desirable but fraught with challenges. These include differences in values, governance structures, and levels of AI maturity among countries. The U.S. and European nations often prioritize transparency, fairness, and individual privacy, while other countries may focus on safety concerns \citep{brookingsStrengtheningInternational}. Additionally, existing international forums for AI governance, such as the Global Partnership on AI (GPAI) and the OECD’s AI Principles, have made progress but lack enforcement mechanisms and broad participation \citep{zhang2022enhancing}.

Another challenge for AI regulation cooperation lies in the competing economic interests of countries. AI is seen as a strategic asset, and there is growing concern about a zero-sum competition between the U.S. and China, where AI capabilities are tied to national security and economic dominance \citep{brookingsStrengtheningInternational}. This competition could hinder efforts to establish collaborative regulatory frameworks, as countries may prioritize their own competitive advantages over global safety concerns \citep{didomiPivotalMoment}.

Given these challenges, a flexible, multi-tiered framework for international AI regulation is the global desideratum. This framework should operate through a combination of binding international agreements and voluntary, non-binding standards that can adapt to regional and sectoral differences. Drawing from previous research, we identify key components for such a framework:

\begin{itemize}
    \item \textbf{Global AI Risk Taxonomy}: A global AI risk taxonomy, akin to the one proposed in this survey, could provide a unified language for discussing AI risks across sectors and countries. This taxonomy would categorize risks into different levels, focusing on critical areas such as data privacy, bias, and misuse. By standardizing how risks are understood and communicated, international actors can more easily align on policy priorities.
    \item \textbf{International AI Standards}: Building on existing initiatives, international standards could be developed under the auspices of a global body such as the United Nations or the World Trade Organization. These standards would focus on ensuring AI systems’ transparency, accountability, and fairness. A global AI council could be established to oversee compliance with these standards, with mechanisms for voluntary reporting and peer review \citep{zhang2022enhancing}.
    \item \textbf{Data Governance and Privacy}: One of the most critical areas for cooperation is data governance, given the central role of data in AI development. The General Data Protection Regulation (GDPR) in Europe provides a template for robust data protection, but international frameworks are needed to manage cross-border data flows and prevent abuses, particularly in countries with weaker regulatory environments \citep{didomiPivotalMoment}. This could involve multilateral agreements on data-sharing protocols and privacy standards, facilitated by international organizations.
    \item \textbf{Ethics and Human Rights}: AI governance must be grounded in universal ethical principles and human rights. This includes ensuring that AI technologies do not exacerbate existing inequalities or infringe on human rights. The UNESCO Recommendation on the Ethics of AI is a step in this direction, but further cooperation is required to ensure that these principles are embedded in national laws and international agreements \citep{brookingsStrengtheningInternational}.
    \item \textbf{Technological Sovereignty and National Security}:  
    To address concerns about the geopolitical implications of AI, international treaties should balance technological sovereignty with security cooperation. This would ensure that AI technologies with potential military applications are subject to export controls and international oversight \citep{zhang2022enhancing}.
\end{itemize}

In conclusion, as AI technology continues to evolve, the need for international cooperation in its regulation becomes more urgent. National regulations, while necessary, are insufficient to manage the global risks posed by AI. An international framework that promotes cooperation, harmonizes standards, and balances innovation with safety is essential. Drawing on existing regulatory efforts and risk taxonomies, we propose a flexible, multi-tiered framework with identified key components to AI governance, one that can adapt to the diverse needs and priorities of different countries while ensuring the responsible development of AI.

By fostering dialogue and cooperation between governments, private sectors, and civil society, the international community can harness the potential of AI while mitigating its risks. The road ahead is challenging, but with a coordinated global effort, AI can be developed in a way that benefits humanity as a whole.

\subsubsection{Technical Oversight Proposals}

Technical oversight has emerged as a key mechanism for mitigating AI risks, focusing on auditing, monitoring, and ensuring the accountability of AI systems \citep{turchin2020classification, ai2023}. The rapid adoption of AI technologies across industries requires an agile and comprehensive regulatory approach, particularly as AI systems often operate in opaque, complex ways that can escape traditional regulatory scrutiny \citep{bommasani2024considerations}.

While much has been discussed on ethical AI and responsible innovation, the technical aspects of oversight remain underexplored. This survey seeks to fill this gap by providing a detailed analysis of technical oversight proposals in AI regulation. It draws from recent literature, including contributions from industry and academia, to offer a broad understanding of the regulatory landscape \citep{DBLP:journals/corr/abs-2310-17688}. The focus will be on the practical implementation of oversight mechanisms that ensure AI systems are safe, accountable, and transparent throughout their lifecycle.

Technical oversight in AI regulation involves a variety of components that ensure AI systems adhere to ethical and safety standards. These components include transparency and explainability, auditing and monitoring, accountability mechanisms, and the establishment of safety standards and certification processes.

\begin{itemize}
    \item \textbf{Transparency and Explainability}: Transparency is a cornerstone of responsible AI governance. For AI systems to be effectively regulated, their decision-making processes must be interpretable by human operators and auditors \citep{DBLP:journals/corr/abs-2310-17688, DBLP:journals/corr/abs-2407-12929}. Explainability refers to the ability to trace and understand how an AI system arrives at its conclusions. This is particularly important in high-stakes fields such as healthcare and criminal justice, where opaque decision-making can lead to harmful consequences. The push for transparency aligns with regulatory frameworks, such as the European Union’s GDPR, which mandates that individuals have the right to an explanation of AI-driven decisions \citep{bommasani2024considerations}.
    \item \textbf{Auditing and Monitoring}: The auditing of AI systems is essential for identifying potential biases, operational flaws, or security vulnerabilities. AI audits can be performed at various stages of system development, from pre-deployment assessments to continuous monitoring once AI systems are in operation \citep{DBLP:journals/corr/abs-2310-17688, DBLP:journals/corr/abs-2407-12929}. Continuous monitoring ensures that AI systems remain compliant with ethical guidelines and legal requirements over time. Monitoring frameworks should include mechanisms for tracking data quality, decision-making processes, and model performance, especially in dynamic environments where AI models learn and adapt \citep{DBLP:journals/corr/abs-2310-17688}.
    \item \textbf{Accountability Mechanisms}: Accountability ensures that developers and operators of AI systems are responsible for the outcomes produced by their technologies. One of the major proposals in this area is the introduction of mandatory incident reporting for high-risk AI applications \citep{DBLP:journals/corr/abs-2310-17688}. This would require companies and organizations to disclose failures or unethical outcomes produced by their AI systems. Additionally, clear guidelines must be established to define liability in cases where AI systems cause harm, particularly in scenarios where the harm could have been anticipated or prevented through proper oversight \citep{bommasani2024considerations}.
    \item \textbf{Safety Standards and Certification}: The development of safety standards and certification processes for AI systems is a critical element of technical oversight. These standards should be based on international cooperation to ensure harmonized regulatory approaches across different jurisdictions. Certification processes would involve third-party assessments to verify that AI systems meet established safety and ethical benchmarks before they are deployed in critical settings \citep{bommasani2024considerations}. Such standards should cover aspects like data privacy, algorithmic fairness, and robustness against adversarial attacks \citep{DBLP:journals/corr/abs-2310-17688, DBLP:journals/corr/abs-2407-12929}.
\end{itemize}

Despite the clear need for technical oversight, several challenges complicate its implementation. The rapid pace of AI development is perhaps the most significant challenge, as regulatory bodies often struggle to keep up with the latest advancements in AI technologies \citep{DBLP:journals/corr/abs-2310-17688}. Additionally, the complexity of many AI systems, particularly deep learning models, makes it difficult to audit and monitor their operations effectively. These models often function as ``black boxes'', where even the developers may not fully understand the intricacies of their decision-making processes \citep{bommasani2024considerations}.

International coordination is another challenge. While some countries or regions, such as the European Union, are leading the charge in AI regulation, there is no global consensus on how AI should be governed. Differing regulatory frameworks across countries can create barriers to the development of standardized technical oversight mechanisms. Furthermore, the lack of a universally accepted set of safety standards makes it difficult to establish a common certification process \citep{DBLP:journals/corr/abs-2310-17688, DBLP:journals/corr/abs-2407-12929}.

The future of technical oversight in AI regulation will likely involve a combination of national and international efforts to create a more unified regulatory framework. There is a growing need for research and development in AI explainability, particularly in making complex models more interpretable without sacrificing their performance. Additionally, governments and organizations must invest in developing the technical expertise necessary to conduct effective audits and assessments of AI systems \citep{bommasani2024considerations}.

Moreover, AI governance must evolve to address the emerging risks associated with advanced AI capabilities, such as autonomous decision-making systems and AI-driven cyberattacks. These risks will require more stringent oversight mechanisms, including licensing regimes for high-risk AI applications and real-time monitoring of AI systems in sensitive sectors like healthcare, finance, and national security \citep{DBLP:journals/corr/abs-2310-17688}.

\subsubsection{Ethics and Compliance Proposals}

The rapid evolution of AI technologies has sparked widespread interest in ensuring these systems are developed and deployed in an ethically responsible manner. Various regulatory proposals, such as the EU AI Act \citep{DBLP:journals/corr/abs-2310-04072}, aim to establish guidelines that mitigate the risks associated with AI, such as bias, privacy violations, and unintended societal harm. However, these regulations must also balance innovation and ethical accountability. 

The dual imperatives of ethical AI and regulatory compliance require robust frameworks that can adapt to the growing complexities of AI. This survey seeks to analyze current ethics and compliance proposals, with a focus on human-centered, responsible AI (HCR-AI), the need for sustainability, and the implementation of AI governance models.

\paragraph{Ethical Frameworks for AI Regulation} Ethical frameworks are essential to AI regulation as they guide the development of AI systems in a way that respects human rights, promotes fairness, and ensures accountability. Many frameworks, such as the one proposed by \cite{mantymaki2022defining}, highlight the need to embed ethics into the organizational processes of AI development. Their ``Hourglass Model'' of governance emphasizes the integration of ethical principles throughout the AI system's lifecycle, from design to deployment.

Another study underscores the importance of ethics-based audits for LLMs \citep{DBLP:journals/corr/abs-2402-01651}, emphasizing the need for these audits to assess AI systems' reasoning abilities in ethically sensitive scenarios. Such audits help identify the ethical values embedded in the AI's decision-making processes and ensure they align with societal expectations.

\paragraph{Compliance Challenges in AI Regulation} AI regulation faces significant compliance challenges, particularly in enforcing ethical standards across jurisdictions. For instance, the EU AI Act represents a pioneering effort in regulating AI, but it also raises concerns about the Act's complexity and the potential hindrance to innovation \citep{DBLP:journals/corr/abs-2310-04072}. Compliance requires organizations to meet stringent requirements concerning transparency, accountability, and bias mitigation, but inconsistent regulations across different regions can complicate these efforts. To address these challenges, a structured approach to AI governance is necessary. The governance model proposed by \cite{mantymaki2022defining} and the ethical auditing frameworks explored by \cite{DBLP:journals/corr/abs-2402-01651} provide practical tools for organizations to ensure compliance without stifling innovation.

\paragraph{Sustainability in AI Regulation} A growing body of research emphasizes the role of sustainability in AI regulation. As AI systems increasingly demand significant computational resources, their environmental impact, particularly energy consumption and carbon footprint, has become a critical concern \citep{DBLP:conf/acl/StrubellGM19}. Integrating sustainability into AI governance ensures that AI systems not only meet ethical standards but also contribute to long-term societal goals, such as those outlined in the United Nations Sustainable Development Goals (SDGs) \citep{DBLP:journals/csur/RolnickDKKLSRMJ23}.

\paragraph{Human-Centered Responsible AI (HCR-AI)} The concept of Human-Centered Responsible AI (HCR-AI) is gaining traction in regulatory discussions. \cite{DBLP:conf/chi/TahaeiCQKMSLBAH23} outline a framework that prioritizes human oversight and accountability in AI decision-making. This approach aligns with ethical AI principles by ensuring that AI systems serve humanity's broader interests while maintaining transparency and interpretability.

Ethics and compliance in AI regulation are increasingly becoming focal points of AI governance discussions. The implementation of ethical frameworks, such as the EU AI Act and HCR-AI models, provides a foundation for ensuring that AI technologies are used responsibly. However, challenges remain, particularly regarding compliance across different jurisdictions and the need for sustainable AI practices. As AI continues to evolve, regulators must strike a balance between fostering innovation and safeguarding ethical standards.

\subsection{Policies}

We evaluate and compare a wide variety of AI governance policies across nations as well as discuss future directions along this dimension.

\subsubsection{Current Policy Evaluation}
Governments, industries, and academic institutions are striving to make policies that balance AI innovation with ethical safeguards. We explore current policies for AI ethics and compliance, evaluating both technical and regulatory solutions.

Ethical concerns in AI range from algorithmic bias to the lack of transparency in decision-making processes. These issues are particularly problematic in sensitive domains like healthcare, finance, and criminal justice. While many technical solutions have been proposed to address these challenges, such as bias mitigation algorithms, there is a consensus that ethical AI development requires a combination of technical innovation and regulatory oversight.

\paragraph{Fairness and Bias} Fairness in AI remains one of the most discussed ethical concerns. Bias in AI systems often results from biased training data, leading to discriminatory outcomes, particularly against minority groups. Recent studies show that even well-intentioned AI models can propagate existing societal biases if not carefully managed \citep{DBLP:journals/corr/abs-2407-07300}. For instance, facial recognition technologies have been shown to have higher error rates for individuals with darker skin tones, raising concerns about their use in law enforcement and surveillance \citep{DBLP:journals/corr/abs-2304-04914}. Several frameworks have been developed to mitigate bias, including fairness-aware algorithms and bias detection tools. However, these technical solutions alone are not sufficient. Regulatory frameworks must also address how AI systems are trained, tested, and deployed to ensure fairness and equity in outcomes \citep{DBLP:journals/corr/abs-2307-12218}.

\paragraph{Transparency and Explainability} Transparency is essential to ensure that AI systems can be trusted. The ``black box'' nature of many AI models, particularly deep learning algorithms, makes it difficult to understand how decisions are made. This opacity poses challenges for accountability and trust, especially when AI systems are used in high-stakes decisions such as medical diagnoses or financial loan approvals \citep{balasubramaniam2022transparency}. Explainability is a proposed solution, where AI systems are designed to provide understandable outputs that can be audited by humans. While progress has been made in this area, achieving full transparency remains a challenge due to the complexity of AI models. Regulatory proposals, such as those in the European Union's AI Act, are focusing on mandating transparency in high-risk AI applications \citep{DBLP:journals/corr/abs-2407-07300}.

\paragraph{Accountability} Accountability in AI involves ensuring that there is a clear chain of responsibility for AI-driven decisions. As AI systems become more autonomous, the question of who is responsible for decisions becomes more complex. Various proposals have been put forward, including AI governance frameworks within organizations that would oversee ethical AI deployment \citep{DBLP:journals/corr/abs-2307-12218}. Additionally, policymakers are considering the creation of external regulatory bodies to enforce accountability in AI applications.

Moreover, compliance with AI ethics and regulatory standards requires robust governance mechanisms. These mechanisms can include both internal governance frameworks within organizations and external oversight by regulatory bodies.

\begin{itemize}
    \item \textbf{Internal Governance}: Many organizations have established AI ethics boards or committees tasked with overseeing the ethical development and deployment of AI systems. These internal bodies are responsible for ensuring that AI projects comply with ethical standards and regulatory requirements. However, the effectiveness of such bodies is often questioned, particularly when they lack enforcement power or are primarily driven by corporate interests \citep{fjeld2020principled}.
    \item \textbf{External Oversight}: External regulatory bodies play a crucial role in ensuring that AI systems comply with ethical and legal standards. These bodies, such as the proposed national authorities in the EU's AI Act, are responsible for auditing AI systems, investigating complaints, and imposing penalties for non-compliance. The creation of independent oversight bodies is seen as a critical step in ensuring that AI systems are held accountable for their actions \citep{fjeld2020principled}.
\end{itemize}

While significant progress has been made in developing ethical and regulatory frameworks for AI, challenges remain. One of the primary challenges is the rapid pace of AI development, which often outstrips the ability of regulators to respond effectively. Additionally, there is an ongoing tension between encouraging innovation and enforcing ethical standards.

Future efforts should focus on creating adaptive regulatory frameworks that can evolve alongside technological advancements. Moreover, international cooperation will be essential in creating globally consistent standards that prevent regulatory arbitrage \citep{DBLP:journals/see/FloridiCKT20}.

\subsubsection{Policy Comparison}
The increasing adoption of AI across various sectors has prompted governments worldwide to implement regulatory frameworks that address concerns related to safety, ethics, and governance. As AI technologies advance, so do the risks associated with their deployment, necessitating a comparison of the different approaches that major regulatory bodies have undertaken. We hence explore the policies adopted in the United States, European Union, and China, drawing comparisons based on risk management, legal enforceability, flexibility, and monitoring requirements \citep{engler2023eu, dlapiperComparingExecutive, ceimia}.

The European Union’s AI Act is a comprehensive regulatory framework aimed at establishing strict governance around AI technologies. The AI Act adopts a risk-based approach, classifying AI applications into several categories based on their potential risks: minimal risk, high risk, and unacceptable risk \citep{ceimia, engler2023eu}. High-risk systems, such as those used in healthcare or biometric identification, are subject to stringent requirements, including human oversight, risk management, and post-market monitoring \citep{dlapiperComparingExecutive}. One of the key aspects of the EU's framework is its alignment with the General Data Protection Regulation (GDPR), ensuring strong privacy protections \citep{engler2023eu}.

In contrast, the United States follows a more decentralized and flexible approach, largely driven by the U.S. Executive Order (EO) on AI. The EO focuses on standards and guidelines rather than binding legislation, with an emphasis on promoting innovation and maintaining competitiveness in AI development \citep{engler2023eu, ceimia}. Key aspects include the use of ``red-teaming'' for high-risk systems, pre- and post-market assessments, and a focus on cybersecurity. The U.S. policy is highly sector-specific, with varying degrees of regulation across different industries such as healthcare and transportation \citep{comunale2024economic}. This flexibility is both a strength and a limitation, as it can adapt to new technological advances but may leave gaps in enforcement.

China has implemented a governance framework for AI, which emphasizes safety control and ethical use, encouraging the integration of AI into state governance systems. The framework focuses on national security and the social implications of AI, with stringent oversight over companies developing AI technologies \citep{ceimia}.

Specifically, key differences and similarities among policies taken by USA, EU and China are summarized as follows:

\begin{itemize}
    \item \textbf{Risk-Based vs. Flexible Approaches}: The EU's AI Act represents a structured, risk-based approach, where high-risk AI applications are tightly regulated to prevent harmful consequences. The U.S., on the other hand, adopts a more flexible and industry-specific regulatory model, focusing on guidelines rather than strict rules \citep{dlapiperComparingExecutive, engler2023eu}. China's approach centers on using AI for social good, with regulatory controls designed to ensure that AI serves social morality, business ethics, individual privacy protection, etc.
    \item \textbf{Enforcement and Legal Binding}: The AI Act enforces legally binding obligations, with penalties reaching up to 6\% of a company's global turnover for non-compliance \citep{engler2023eu}. In contrast, the U.S. policy leans towards voluntary compliance, with penalties arising more from sectoral laws than from a unified AI regulatory framework. China’s regulatory measures are centralized, with state oversight mechanisms and severe penalties for non-compliance \citep{ceimia}.
    \item \textbf{Implementation Challenges}: One of the most significant challenges for the EU is the complexity of implementing AI standards across member states and industries, particularly for ``high-risk'' AI systems \citep{engler2023eu}. In the U.S., the lack of a cohesive federal law on AI governance may create inconsistencies between states \citep{DBLP:journals/ethicsit/AlmeidaSF21}.
\end{itemize}

As AI continues to evolve, the regulatory landscape will need to adapt. The EU’s risk-based, legally binding framework offers strong protections but may struggle with flexibility, while the U.S.’s guideline-based approach promotes innovation but risks leaving gaps in oversight. Policymakers must work towards harmonizing these approaches to ensure safe, fair, and effective global AI governance.

\subsubsection{Future Policy Directions}
In practice, policymakers face the challenge of creating a balanced regulatory framework that encourages AI innovation while addressing ethical, legal, and societal concerns. In this section, we explore the future policy directions in AI regulation, emphasizing the need for flexible, transparent, and inclusive frameworks that adapt to the evolving AI landscape.

\paragraph{Risk-Based Regulatory Approach} A prominent direction for future AI regulation is the adoption of a risk-based regulatory framework. Under this approach, AI applications are classified based on their potential risks to individuals, society, and national security. High-risk applications, such as those used in critical sectors like healthcare, autonomous vehicles, or criminal justice, would be subject to stringent regulatory oversight, including mandatory transparency, regular audits, and accountability mechanisms. Lower-risk applications, like AI-powered customer service tools, might face more lenient requirements, promoting innovation without unnecessary regulatory burdens. A risk-based approach ensures that regulation is proportional to the level of threat posed by an AI system, allowing for both innovation and protection. Policymakers must also consider sector-specific standards and harmonize regulations across international borders to avoid fragmented approaches that could stifle global cooperation.

\paragraph{Promoting Ethical AI Development} Ethical AI development is another crucial pillar of future AI policy. Policymakers need to prioritize the establishment of ethical guidelines that ensure AI systems are designed and deployed in ways that respect human rights, fairness, and non-discrimination. This includes implementing mechanisms to eliminate bias in AI algorithms, enhancing transparency in decision-making processes, and ensuring that AI systems are accountable for their outcomes. Moreover, public participation and multi-stakeholder collaboration, including input from ethicists, civil society, and industry experts, should be a core component of regulatory development. This inclusive approach will allow for the consideration of diverse perspectives, ensuring that AI regulation reflects societal values and priorities.

\paragraph{International Cooperation and Standards Harmonization} As AI is a global technology, international cooperation is essential to creating effective and consistent regulatory frameworks. Countries should work together to develop shared standards and principles that can guide the responsible development and deployment of AI. Establishing global norms will help prevent regulatory arbitrage, where companies seek the least restrictive environments, and ensure that AI systems adhere to ethical standards no matter where they are developed or used. International organizations, such as the United Nations, and the Organization for Economic Co-operation and Development (OECD), are already taking steps to foster global dialogue on AI regulation. Future policy directions should build on these efforts, encouraging the creation of international treaties or agreements that promote ethical AI while balancing innovation with accountability.

AI regulation is a rapidly evolving field that requires adaptable, forward-looking policies. A risk-based regulatory approach, ethical guidelines, and international cooperation are key to ensuring that AI technologies contribute positively to society while mitigating their potential risks. Future policymakers must work collaboratively with stakeholders across sectors and borders to create a regulatory environment that fosters innovation, protects human rights, and ensures that AI serves the common good.

\subsection{Visions}

Finally, we discuss both the long-term vision for AI governance and vision of the integration of AI and society, as well as risks and opportunities in realizing these visions.

\subsubsection{Long-term Vision}
The long-term vision of AI governance emphasizes the need for a framework that encourages innovation while ensuring safety and ethical compliance. Leading scholars and policymakers often discuss the ideal outcome as one where AI systems enhance human flourishing, improve social welfare, and strengthen global economic capabilities \citep{wef2021, allen2024roadmap}. This vision is rooted in the idea of responsible AI development, which entails creating AI systems that are transparent, fair, and accountable. For instance, the European Union’s AI Act strives to establish a comprehensive legal framework to ensure these systems are secure and aligned with ethical standards \citep{itic2024}.

A key component of the long-term vision is the democratization of AI technologies, ensuring that access and benefits are equitably distributed across different regions and populations. Moreover, this vision underscores the importance of global cooperation in shaping regulatory frameworks that prevent harmful use of AI, while fostering the development of technologies that benefit humanity as a whole \citep{futureoflifeTurningVision}.

\subsubsection{Vision of Technological and Social Integration}
The integration of AI into society is with both technical and ethical challenges. The vision here focuses on the harmonious blending of AI capabilities with societal needs and values. AI technologies, if developed and deployed responsibly, have the potential to address key societal challenges—ranging from healthcare and education to environmental sustainability. However, achieving this requires that AI systems be designed with the intention of respecting human rights, privacy, and fairness \citep{upennArtificialIntelligence, csisPathTrustworthy}.

In particular, there is a strong push towards aligning AI with social values through collaborative governance, wherein multiple stakeholders—including industry leaders, civil societies, and governments—participate in shaping AI policies \citep{kai2024}. This ensures that AI technologies are not only technically robust but also socially inclusive and beneficial. For instance, initiatives like the U.S. AI Bill of Rights outline key ethical principles such as data protection and bias prevention, which are critical to ensuring that AI systems align with human values and societal norms \citep{whitehouseBlueprintBill}.

\subsubsection{Risks and Opportunities in Realizing Visions}
While the visions for AI governance are ambitious, the path toward their realization is fraught with both risks and opportunities. On the risk side, the rapid advancement of AI raises concerns over data privacy, security vulnerabilities, and algorithmic bias. One of the most significant challenges is ensuring that AI systems do not exacerbate existing social inequalities through biased decision-making processes \citep{upennArtificialIntelligence}. Furthermore, issues like data poisoning and model extraction attacks present serious security risks that could undermine trust in AI technologies \citep{upennArtificialIntelligence}.

Conversely, the opportunities presented by AI are immense. When governed effectively, AI can revolutionize industries, drive economic growth, and provide solutions to global challenges such as climate change and public health crises. For example, generative AI has the potential to transform creative industries, while machine learning can enhance predictive models in fields like finance and healthcare \citep{ai2024artificial}. Additionally, AI has the capacity to improve governance systems themselves by optimizing decision-making processes and increasing governmental transparency and accountability \citep{margetts2022rethinking}.

Balancing these risks and opportunities requires a nuanced approach to governance, one that is agile enough to adapt to the rapidly evolving technological landscape while remaining grounded in ethical principles. Policymakers must consider both the immediate risks posed by AI technologies and the long-term benefits they offer when formulating governance frameworks \citep{Covino2024}.

In conclusion, the visions for AI governance reflect a broad consensus on the importance of creating a regulatory environment that fosters innovation while mitigating risks. Long-term goals emphasize the ethical deployment of AI for social good, while the vision of technological and social integration focuses on aligning AI with societal values. The successful realization of these visions will require proactive governance that balances the inherent risks and opportunities of AI, ensuring that this transformative technology can be harnessed for the benefit of all.

%% file: sections/Challenges_and_Future_Directions.tex
\section{Challenges and Future Directions}

To conclude this survey, we discuss potential avenues for future research on the safety of LLMs, as well as present our perspectives on key topics that should be prioritized in upcoming research efforts. By identifying these critical areas, we aim to contribute to the ongoing discourse surrounding the responsible development and application of LLMs, ensuring their safe integration into various societal and technological contexts.

\subsection{Exploring Safe Architectures}

Beyond the development of safety mechanisms and modules, rethinking the foundational architecture of LLMs is crucial for long-term safety improvements. Current LLM architectures, primarily based on transformer models, exhibit significant vulnerabilities, including susceptibility to adversarial attacks, data leakage, and unintended memorization of sensitive information. These issues often stem from the inherent trade-offs between model size, performance, and safety. As LLMs increase in scale to handle more complex tasks, they also become more prone to leaking confidential or harmful data and are harder to control effectively.

Exploring new architectures with safety as a core design principle is an emerging research area that could revolutionize the field \citep{Brendon_Wong_2023, Sutton_2024}. For instance, models that inherently compartmentalize knowledge or have built-in redundancy and recovery mechanisms could mitigate the risks posed by attacks or unintentional information leaks. Moreover, it is also important to explore architectures that not only enhance the immediate safety of models but also provide a more robust foundation for addressing future, unforeseen challenges. In this context, the development of architectures that prioritize both performance and safety should become a focal point for future research, helping to ensure that LLMs remain secure, reliable, and effective as they continue to evolve.

\subsection{Safety Control Modules}

To improve the safety of LLMs, introducing a dedicated safety control module presents a promising avenue for exploration. Current LLM architectures often lack integrated, real-time safety checks that can monitor and mitigate potential threats or vulnerabilities as they emerge. A safety control module could function as an intermediary between the LLM and its output, intercepting and scrutinizing responses before they are delivered to users. Such a module would be designed to detect harmful content, privacy breaches, or other forms of unsafe behavior dynamically. For instance, \citet{DBLP:journals/corr/abs-2312-06674} propose an LLM-based input-output safeguard to enhance AI safety and content moderation by classifying human prompts and model outputs. This exploration lays a foundation for future research that could further refine safety control modules.

It is worth noting that the design and implementation of such a module come with significant challenges. The module must be highly adaptable, capable of identifying a wide range of safety issues without compromising the efficiency or accuracy of LLMs. Additionally, it must be scalable, as LLMs continue to grow in size and complexity. There is also a risk of the module over-filtering outputs, potentially leading to unnecessary censorship or inhibiting the LLM's ability to generate creative or novel responses. Therefore, developing a balance between rigorous safety checks and maintaining the LLM's core functionality will be essential. Future research should focus on refining these mechanisms, exploring how such modules can operate autonomously while allowing for human oversight when necessary.

\subsection{Toward Effective and Unified Safety Mechanisms}

Current approaches to enhancing the safety of LLMs exhibit notable limitations and inefficiencies, underscoring the urgent need for more robust and effective solutions. For instance, a study investigating the use of DPO to reduce toxicity \citep{DBLP:conf/icml/LeeBPWKM24} has found that the alignment algorithm does not completely eliminate toxic content but instead circumvent it by bypassing sensitive neural regions. Additionally, model editing is inefficient either, as it requires manual, instance-by-instance modifications that only affect specific pieces of information without altering related content \citep{DBLP:journals/tacl/CohenBYGG24,DBLP:conf/emnlp/ZhongWMPC23,DBLP:conf/emnlp/QinZHYLJ24}. Similarly, machine unlearning—designed to enable models to forget harmful information—has demonstrated instability. Experimental evidence suggests that even after harmful data is ostensibly erased, a machine-unlearned LLM may still recall private data, and with few-shot fine-tuning, it can revert to its previous state, recovering the erased information \citep{lucki2024adversarial,DBLP:journals/corr/abs-2402-16835}. Incomplete forgetting exacerbates the issue, as models can still leak private information when prompts are altered. Therefore, more effective methods are necessary. To meet the increasing safety demands of LLMs, a universal, unified mechanism capable of permanently eliminating unsafe information is imperative.

\subsection{Improving Safety Evaluations for LLMs}

First, most existing evaluation metrics are tailored to specific benchmarks or tasks, providing a fragmented and limited view on LLMs (either capability or safety). This specificity highlights the pressing need for a unified evaluation metric/framework capable of comprehensively assessing LLMs across a wide range of scenarios. Such a metric would ensure these models are well-equipped to meet the demands of various tasks and contexts. Such a framework must account for differences in architectures, training data, and intended use cases among LLMs, offering a balance between consistency in evaluation and flexibility to accommodate different model designs.

Second, the rapid evolution of LLMs has exposed significant gaps in current evaluation methodologies, precipitating what some researchers describe as an ``evaluation crisis''. This triggers emerging interests in the development of science of evaluations underscoring the need for universal evaluation theories and methodologies that can address complex, real-world scenarios \citep{DBLP:journals/corr/abs-2404-00021,zhan2024short}.

\subsection{Toward Multivalent International Cooperation and Interdisciplinary Community Building}

AI safety governance must evolve to address the growing complexity and global integration of AI technologies. Future directions emphasize the need for multilateral regulatory frameworks that harmonize standards across jurisdictions, ensuring interoperability and joint enforcement mechanisms. Such frameworks should account for diverse ethical and cultural values, integrating cross-sector collaborations that bring together technical, legal, and ethical expertise. Interdisciplinary approaches are essential, particularly in developing governance tools that incorporate both technical and ethical metrics, and conducting human-AI interaction studies to ensure AI systems function equitably across different socio-economic contexts. Moreover, AI governance must focus on multivalent value systems, where ethical imperatives—such as inclusivity and sustainability—shape regulatory practices.

A key priority is the establishment of global AI safety communities that foster continuous dialogue and collaboration. These communities should consist of transdisciplinary research consortia, participatory governance platforms, and educational initiatives, ensuring diverse voices contribute to shaping AI safety standards. Finally, AI safety governance must address geopolitical and ethical considerations, such as AI’s role in international relations and its impact on global inequalities, which require coordinated strategies to mitigate risks and ensure safe AI deployment globally. This holistic, inclusive approach will help AI governance frameworks evolve in tandem with the challenges posed by advanced AI systems.

%% file: sections/Conclusion.tex
\section{Conclusion}

The rapid advancement of LLMs has ushered in a new era of AI capabilities, with the potential to revolutionize various sectors and aspects of society. However, these technological leaps forward have also introduced a myriad of safety concerns and ethical challenges, underscoring the need for a deeper understanding of their associated risks and the development of effective mitigation strategies. This necessitates a comprehensive survey to guide the safe deployment and responsible use of LLMs.

This survey has delved into the multifaceted issues surrounding LLM safety, spanning from value misalignment, robustness against adversarial attacks, misuse, and the emerging issues surrounding autonomous AI systems. Each of these aspects carries unique implications for the responsible use of LLMs and requires tailored strategies for risk management.

The study also highlights the importance of interpretability in understanding LLM behavior, which is critical for improving safety measures and fostering user trust. Furthermore, the insights into safety measures implemented by leading AI organizations offer valuable guidance for industry stakeholders aiming to enhance the safety of their models. Additionally, our review of agent safety and governance underlines the need for a well-coordinated regulatory framework that can adapt to the rapid evolution of LLM technologies. This involves both national and international efforts to ensure that AI deployment is aligned with ethical standards and societal values.

As LLMs become increasingly integrated into critical decision-making processes, ensuring their safe and ethical deployment is not only a technical challenge but a societal imperative. This survey aims to serve as a foundational reference for researchers, policymakers, and industry practitioners, facilitating a more nuanced understanding of LLM safety and fostering collaboration across technical and regulatory domains. Ultimately, we hope that this work will contribute to the safe and beneficial development of LLMs, ensuring their alignment with the broader goals of societal well-being and human flourishing.